\documentclass[twocolumn,twoside]{IEEEtran}                

\usepackage{amsfonts}
\usepackage{flushend}
\usepackage[affil-it]{authblk}
\usepackage[linktocpage=true]{hyperref}
\usepackage{amssymb}
\usepackage{amsbsy}
\usepackage{amsmath}
\usepackage{cite}
\usepackage{subfigure}
\usepackage{multirow}
\usepackage{cite}
\usepackage{url}
\usepackage[pdftex]{graphicx}
\usepackage{color}
\usepackage{wrapfig}
\usepackage{algorithmic}
\usepackage{algorithm}
\usepackage[all]{xy}

\newcommand{\Real}{\mathbb{R}}

\begin{document}

\title{Sequential Principal Curves Analysis}

\author{Valero Laparra and Jes\'us Malo}
\affil{Image Processing Laboratory (IPL). Universitat de Val{\`{e}}ncia. Spain. \\ \{valape,jmalo\}@uv.es, http://isp.uv.es
\vspace{-0.8cm}}

\markboth{IPL-Universitat de Valencia Technical Report 2015}{Laparra and Malo. Sequential Principal Curves Analysis}

\maketitle

\begin{abstract}

This report includes all the technical details of the
Sequential Principal Curves Analysis (SPCA) in a single document.
SPCA is an unsupervised nonlinear and invertible feature extraction technique.
The identified curvilinear features can be interpreted as a set of nonlinear sensors:
the response of each sensor is the projection onto the corresponding feature.
Moreover, it can be easily tuned for different optimization criteria
--e.g. \emph{infomax}, \emph{error minimization}, \emph{decorrelation}--
by choosing the right way to measure distances along each curvilinear feature.

Even though proposed in \cite{Laparra12} and shown to work in multiple
modalities in \cite{Laparra15b}, the SPCA framework has its original roots in the
nonlinear ICA algorithm in \cite{Malo06b}. Later on, the SPCA philosophy for nonlinear
generalization of PCA originated substantially faster alternatives at the cost of introducing different
constraints in the model. Namely, the Principal Polynomial Analysis (PPA) \cite{Laparra14a},
and the Dimensionality Reduction via Regression (DRR) \cite{Laparra15a}.
This report illustrates the reasons why we developed such family and is the appropriate
technical companion for the missing details in \cite{Laparra12,Laparra15b}.
See also the data, code and examples in the dedicated sites
http://isp.uv.es/spca.html and http://isp.uv.es/after\_effects.html

\end{abstract}


\vspace{-0.4cm}

\tableofcontents

\section{Introduction}

\IEEEPARstart{C}{lassical} unsupervised learning such as Principal Components Analysis (PCA)
and Independent Component Analysis (ICA) is useful to design artificial sensory systems and
to understand the organization of natural sensory systems.
On the artificial side, examples include representations/transforms for image coding \cite{Clarke85,Gersho92,Hancock92,Taubman01} and
image categorization \cite{Lowe04,Wright10}. On the natural side, examples include the
analysis of visual cortex \cite{Field96,Bell97,Hoyer00,Simoncelli01,Doi03}.
PCA and ICA obtain basis of the space according to different optimization criteria.
These basis functions can be interpreted as linear sensors: the projection of data
onto these basis represents the response of the set of sensors.
PCA defines a sensor hierarchy: for example, an image sensory system made out of
principal directions with highest eigenvalues \emph{minimizes the image reconstruction error} \cite{Clarke85,Gersho92}.
In ICA, the basis is intended to provide responses as independent as possible, which is equivalent
to design a sensory system that maximizes the transmitted information (\emph{infomax}) \cite{Laughlin83,Bell95}.

Even though linear unsupervised techniques have been successful to (1) \emph{design sets of
artificial sensors}, and (2) to \emph{explain the receptive fields of biological sensors}, they have fundamental problems:
\begin{itemize}
\item \textbf{Sensors are linear.} Linear transforms are not efficient when the data do not come from a linear manifold or when
the natural sensory system at hand is nonlinear.
For example, in vision, PCA or ICA models are too simple for natural images \cite{Simoncelli97,Bucigrossi99,Hyvarinen03,Malo06ieee,Malo06b,Gutierrez06,Camps08,Hyvarinen09,Malo10},
and linear feature extraction cannot explain the nonlinear responses of spatial frequency analyzers in V1 \cite{Foley94,Watson97,Carandini94,Cavaugnagh02,Carandini12}.
\item \textbf{Features are treated independently.} Projecting the data in each basis function independently
is an advantage to interpret each dimension of the transform as a separate sensor. However, this simplistic model
ignores eventual relations between linear responses. For instance, in vision,
contrast masking experiments \cite{Foley94,Watson97} and neuron recordings \cite{Carandini94,Cavaugnagh02,Carandini12}
indicate inhibitory interactions between linear mechanisms. Moreover, coefficients of linear ICA-like image
transforms are not independent \cite{Simoncelli97,Bucigrossi99,Hyvarinen03,Gutierrez06,Malo06b,Camps08,Hyvarinen09,Malo10,Malo06ieee}.
\item \textbf{The metric is global.} Linear transforms imply a constant Jacobian and hence the
induced metric is global, i.e. equal in each point \cite{Epifanio03,Malo06ieee,Laparra10}.
This is an important restriction since artificial and natural sensory systems have limited resolution and
more resources have to be allocated in more populated regions of the space.
This means a point-dependent resolution/metric/sensitivity, and not just a different weight per
dimension \cite{Watson97,Laparra10}.
\end{itemize}

A number of nonlinear manifold learning techniques have been developed that could be used to address the obvious problems of linear techniques.
Unfortunately, more sophisticated methods are harder to interpret as a set of sensors.
In order to be suitable for the interpretation or design of sensory systems, a nonlinear
technique should have the following features:
\begin{itemize}
\item \textbf{Explicit form of the sensors.}
An explicit representation of the sensors (analogous to the principal directions in PCA or the independent directions in linear ICA)
is desirable. Implicit transforms may be flexible enough to capture data nonlinearities but one has no separate
access to the properties of each sensor, which is mandatory if the properties of specific biological sensors
have to be reproduced. Note that an explicit set of nonlinear sensors in the input space is helpful since it defines a
curvilinear coordinate system. In this situation the response of such sensory system is just an intuitive change of coordinates.
\item \textbf{Out-of-sample problem.} It is important that the method is applicable to data which is not in the training set.
\item \textbf{Invertibility.} Invertibility ensures that changes in the response domain may be analyzed back in the original space.
This is useful to assess the representation error when considering a reduced set of sensors (dimensionality reduction) or when they have limited resolution (transform quantization).
Invertibility is also useful to obtain the meaning of the coordinates of the transform domain back in the input space. 
\item \textbf{Tunable metric.} As stated above, sensitivity should be focused on specific regions of the input space according to the Probability Density Function (PDF). The method has to include different principles guiding
this point-dependent metric. A clear definition of the metric according to different organization criteria allows to explore the principles underlying the behavior of biological systems.
\end{itemize}

Conventional nonlinear manifold learning methods do not fulfill all the above requirements at the same time.

For instance, though efficient in many tasks, {\em spectral} methods~\cite{Roweis00,Belkin02,Weinberger04} and {\em kernel}
methods~\cite{Scholkopf98} do not generally yield intuitive mappings between the original and the intrinsic curvilinear
coordinates of the low dimensional manifold.
In addition, even though a metric can be derived from particular kernel functions~\cite{Burges99}, the interpretation
of the transformation is hidden behind implicit mappings, and out-of-sample extensions are typically difficult, if
not impossible.

An alternative family of manifold learning methods describe complicated manifolds as a mixture of local
models~\cite{Kambhatla97} that may be merged into a single global
representation~\cite{Roweis02,Verbeek02,Teh03,Brand03}. The explicit direct and inverse transforms can be
derived from the obtained mixture model. However, explicit description of the sensors is not straightforward,
and moreover, the effect of the local coordination
in the (eventually point-dependent) metric was not explicitly analyzed.
In the context of nonlinear ICA, an alternative way of merging locally disconnected representations
was proposed in~\cite{Malo06b}. In that case, the global representation was based on the fact that the
Jacobian of the global nonlinear ICA may be differentially approximated by local linear ICA~\cite{Lin99}.
Accordingly, the global representation was obtained by integrating the Jacobian.
However, obtaining the global representation through integration
in arbitrary paths requires special symmetries in the manifold (e.g. Stokes condition).
Moreover, the invertibility of the transform was not addressed therein~\cite{Malo06b}.

Self-Organizing Maps (SOM)~\cite{Kohonen82} and variants~\cite{Bishop98} are based on tuning a predefined
topology in such a way that the curved sensors and the {\em complete} lattice of discrete responses
are obtained simultaneously. Therefore, training is computationally demanding in highly dimensional
scenarios as is the case for visual stimuli.
Moreover, the appropriate choice of the SOM lattice resolution (number of discrete perceptions per dimension)
requires a priori knowledge of the sensory system one wants to model.
Another candidate to describe visual sensors is isometric feature mapping (\emph{Isomap})~\cite{Tenenbaum2000}.
It computes geodesic distances in the manifold obtaining an unfolded representation of the data.
Nevertheless, it is not invertible and hence the explicit description of the sensors in the input
space is not easy to obtain. Moreover, including different design criteria in the \emph{Isomap} metric
is not straightforward.

Here we present the general formulation of an alternative learning technique: the Sequential Principal Curves Analysis (SPCA). The basic SPCA idea is generalizing the explicit set of sensors of linear
unsupervised learning by using a set of Principal Curves (PCs).
SPCA assumes a smooth manifold made of local clusters, as the local coordination
literature \cite{Roweis02,Verbeek02,Teh03,Brand03}. However, unlike \cite{Roweis02,Verbeek02,Teh03,Brand03},
no mixture of models is computed here. On the contrary, as in \cite{Malo06b}, we integrate a
suitable Jacobian intended for component independence. The distinctive features of SPCA are:
(i) new PDF-based Jacobian (or metric) that can be tuned to either the \emph{infomax} or the \emph{error minimization} principles,
(ii) using the concept of secondary PCs \cite{Delicado01}, we obtain a set of non-linear sensors made
of a specific sequence of secondary curves,
and (iii) straightforward out-of-sample extension and invertibility.

\begin{figure*}[t!]
\begin{center}
\begin{tabular}{c}
\hspace{-0.7cm}
\includegraphics[width=15cm]{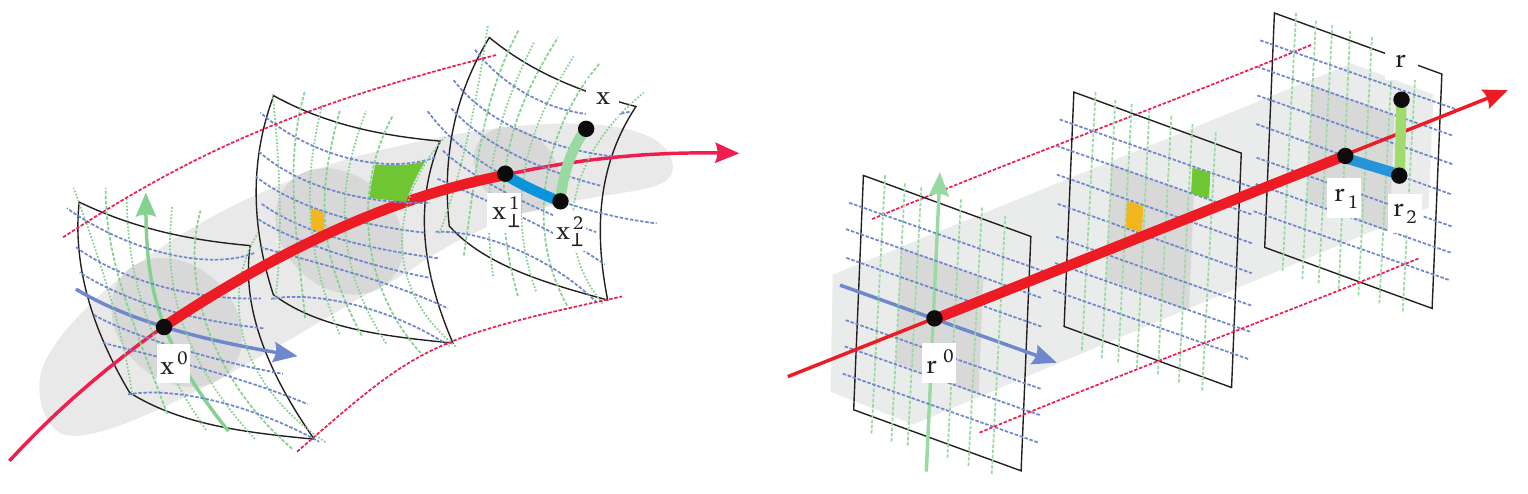}	
\end{tabular}
\end{center}
\vspace{-0.3cm}\caption{\small The SPCA leading idea is removing redundancies by unfolding the dataset along the first and secondary principal curves (PCs) while performing local equalization along the way. Left plot represents the input domain ${\bf x}$ and right plot represents the response domain ${\bf r}$. Gray regions represent the underlying PDFs. As in SOM, a curvilinear lattice is assumed (thin red, blue, and green curves). However, unlike in SOM, the computation of the whole lattice is not needed: to transform certain ${\bf x}$ just the path in bold style, made of segments of PCs, is required. Moreover, the specific resolution per dimension emerges from data: the proposed Jacobian with the embedded metric, Eq. \ref{generalSPCA_jacobian}, implies that highly populated regions in the input (e.g. orange area) are expanded in the response while lower density regions (e.g. green area) are shrunk. Given an origin, ${\bf x}^o$, in the first PC (red line) and some point of interest, ${\bf x}$, the response for the point of interest is given by the lengths (the integrals in Eq.~\ref{global_response2}) along this path: the first (or standard) PC in red, the second PC (in blue) in the orthogonal subspace at ${{\bf x}_{\bot}^1}$, which is the (geodesic) orthogonal projection of ${\bf x}$ on the first PC, and the third PC (in green) in the orthogonal subspace at ${{\bf x}_{\bot}^2}$.}
\label{digrama2}
\end{figure*}

Fig. \ref{digrama2} illustrates the SPCA concept.
SPCA instrumentally requires an algorithm to draw first~\cite{Hastie89} and secondary~\cite{Delicado01} PCs
\emph{from specific points} (i.e. a local-to-global algorithm). Suitable choices include those in \cite{Delicado01,Einbeck05,Einbeck10}
or the one used here (see the Appendix below). However, note that the choice to draw individual PCs is not the core of SPCA,
but the PDF-related metric and the specific sequential path for the Jacobian integration leading to the curvilinear
coordinate system (the nonlinear sensors).

Related work includes extensions of PCA generalizing principal components from straight
lines to curves, based on: (1) non-analytic principal curves~\cite{OzertemTesis,Ozertem11},
(2) fitting analytic curves~\cite{Jolliffe02,Donnell94,Besse95,Laparra12ppa}, and (3) implicit
methods using neural networks and autoencoders~\cite{Kramer91,Hinton06,Scholz07}.

The distinctive property of SPCA is that it identifies an explicit system of separate sensors
with tunable resolution. In \cite{OzertemTesis,Ozertem11} the authors suggest that their projection
method could be used for signal representation \emph{if applied
sequentially}. However, they acknowledge that their current setting lacks the required accuracy \cite{Ozertem11}.
The nonlinear features identified by neural networks \cite{Kramer91,Hinton06,Scholz07} are not explicit in the
formulation, complicating their use in the interpretation of nonlinear sensors (as for instance to understand
image texture analyzers in V1). Finally, the consideration of biologically plausible organization principles
(such as \emph{infomax} or \emph{error minimization}) and their metric effects were not addressed in~\cite{OzertemTesis,Ozertem11,Jolliffe02,Donnell94,Besse95,Laparra12ppa,Kramer91,Hinton06,Scholz07}.

The work is organized as follows. Section \ref{strategies} reviews the \emph{infomax} and the \emph{error minimization} principles and their geometric effects. Section \ref{motivation} motivates the proposed Jacobian
and integration path by making observations on smooth manifolds. Section \ref{spca} describes
our proposal: the Sequential Principal Curves Analysis with tunable metric.
The experimental Section \ref{experiments} empirically checks the SPCA assumptions and abilities through a series of examples:
(1) sensible geodesic projections,
(2) convergence of transform and inverse,
(3) nonlinear ICA and optimal transform coding as a function of the PDF-based metric.
(4) dimensionality reduction,
(5) domain adaptation,
(6) enhanced classification through generalization of the Mahalanobis metric.
Finally the Appendix describes the instrumental algorithm used here to draw a single Principal Curve.

\section{Infomax, error minimization and local metric} \label{strategies}

Processing input samples ${\bf x} \in \mathbb{R}^d$ requires the design of an {\em appropriate} set of $d$ sensors that
transform observations into ${\bf r}=R({\bf x}) \in \mathbb{R}^d$.
Limited sensor resolution or internal noise in the responses are modeled as some sort of quantization Q,
\begin{equation}
\xymatrix{
{\bf x} \ar@/^/[r]^{R} \ar@/_1pc/@{<-}[rr]_{R^{-1}}
 & {\bf r} \ar@/^/[r]^{Q} & {\bf r^\star}}
\label{generic_transform}
\end{equation}
The \emph{infomax}~\cite{Laughlin83,Bell95} and \emph{error minimization}~\cite{MacLeod03} design principles
imply different links between the sensitivity of the system (its Jacobian $\nabla R({\bf x})$),
the non-Euclidean metric induced by the system~\cite{Epifanio03,Malo06ieee,Laparra10}, and the signal PDF, $p({\bf x})$.
However, these links restrict but do not determine the response
function \cite{Hyvarinen99,Laparra10rbig}. In this section we review the links and in Section \ref{motivation}
we make additional considerations that lead to a particular design for $R$.

On the one hand, {\em infomax} looks for transforms $R$ such that ${\bf r}$ has independent components or maximum entropy,
which leads to~\cite{Laughlin83,Bell95}:
\begin{equation}
|\nabla R({\bf x})| \propto p({\bf x})
\label{noise_free_jacobian}
\end{equation}
On the other hand, minimization of $|{\bf x} - R^{-1}({\bf r^\star})|_2$ leads to a different constraint. The optimal MSE sensitivity fulfils~\cite{MacLeod03},
\begin{equation}
|\nabla R({\bf x})| \propto p({\bf x})^{1/3}
\label{noisy_jacobian}
\end{equation}
which is consistent with the classical optimal MSE distribution of discrete perceptions in Vector Quantization~\cite{Gersho92}.
The exponent accompanying the PDF in the Jacobian will be hereafter referred to as $\gamma$.

Non-linear responses have geometrical effects. Intuitively, a sensory system with non-trivial response
induces a non-Euclidean metric in the input space: if the sensitivity (slope of the response) is bigger
at some region of the input space,
distortions in that region
will be more relevant to the system.
In particular, assuming that
the internal representation, ${\bf r}$, is Euclidean,
the following {\em perceptual} metric matrix is induced by the system at the input domain~\cite{Epifanio03,Malo06ieee,Laparra10}:
$M({\bf x}) = \nabla R({\bf x})^\top \cdot \nabla R({\bf x})$.

Accordingly, relations between the sensitivity of the system and the PDF of the input space
(as required by \emph{infomax} or \emph{error minimization})
will also give rise to relations between the induced metric and the PDF:
\begin{equation}
 |M({\bf x})| \propto p({\bf x})^{2\gamma}
 \label{metrica}
\end{equation}

The above restrictions on $|\nabla R|$ and $|M|$ do not fully determine $R$.
In fact, an infinite family of transforms is suitable for \emph{infomax} \cite{Hyvarinen99,Laparra10rbig}.
Next section makes additional considerations that inspire our proposal for $\nabla R$.

\section{Motivation in the infomax context} \label{motivation}

The following properties of curved smooth manifolds motivate the proposed Jacobian and integration path:
\begin{enumerate}
\item 
      Global linear transforms, ${\bf v} = W \cdot {\bf x}$, are not appropriate for component independence
      in curved manifolds since the conditional mean with regard to one linear component (or direction),
      say $v_1$, is not constant in the range of $v_1$.
      This means that the conditional PDF, $p(v_2,\ldots,v_d|v_1)$, depends on $v_1$ and hence, the orthogonal
      subspace $(v_2,\ldots,v_d)$ statistically depends on $v_1$.
\item Unfolding along one PC is a sensible step towards independence since it makes equal the
      first moment (the mean) of the conditional PDFs along the PC. However, complete independence may require additional processing.
\item Additional processing after unfolding should make equal the higher order moments.
      For instance, if manifolds are approximated as mixtures of Gaussian clusters, the second
      moment (the covariance) along the PC can be made equal by local expansions (local metric changes) in
      the unfolded domain. In general, by choosing a metric proportional to the probability
   	  (Eq. \ref{noise_free_jacobian}), the conditional PDFs are uniformized. In this way, all the moments are
      constant along the PC.
\end{enumerate}
\subsection[Unfolding along a Principal Curve]{Unfolding along the PC: the cumulants perspective}

Unfolding along PCs (or alignment of the clusters in the manifold) implies a step in the right (independence)
direction but it is not enough since a metric change is still needed. This is easy to see by looking at the
cumulant expansion of the conditional PDFs.

Unfolding along a PC with parameter, $u_1$, implies independence with regard to orthogonal
subspaces {\em iff} it gives rise to:
\begin{equation}
      p(u_2,\ldots,u_d|u_1) = p(u_2,\ldots,u_d)
\end{equation}
Therefore, the cumulant generating functions of both sides of the above equation should be equal:
\begin{equation}
1 - j \boldsymbol{\omega}^{\top} {\bf m}_1 + \frac{1}{2} \boldsymbol{\omega}^{\top} \boldsymbol{m}_2 \boldsymbol{\omega} - \ldots =
1 - j \boldsymbol{\omega}^{\top} {\bf m}'_1 + \frac{1}{2} \boldsymbol{\omega}^{\top} \boldsymbol{m}'_2 \boldsymbol{\omega} - \ldots
\label{cumulants}
\end{equation}
where ${\bf m}_i$ and ${\bf m}'_i$ are the $i$th-order moments of each PDF, and $\boldsymbol{\omega}$ is the
parameter of the characteristic functions. Independence holds if ${\bf m}_i = {\bf m}'_i$, $\forall i$.

One PC satisfies the parametric equation~\cite{Hastie89}:
\begin{equation}
      f(u_1) = {\mathbb E}[ {\bf x} | \lambda({\bf x}) = u_1 ]
\end{equation}
where $\lambda({\bf x})$ is the orthogonal projection of ${\bf x}$ on the curve, so
$\{{\bf x} | \lambda({\bf x})=u_1\}$ is the orthogonal subspace at the curve
point $u_1$. According to this, the curve passes through the average of the orthogonal subspace
(the origin of the subspace in the unfolded representation):
\begin{equation}
      {\mathbb E}[ u_2,\ldots,u_d | u_1] = {\bf 0}, \forall u_1 \Rightarrow {\mathbb E}[u_2,\ldots,u_d] = {\bf 0}
\end{equation}
which means that {\em unfolding along a PC} makes the averages equal: ${\bf m}_1={\bf m'}_1={\bf 0}$.
However higher order moments (e.g. variance) may not be equal along the curve.
Therefore, unfolding is good in independence terms (i.e. it helps to fulfill Eq. \ref{cumulants}), but
additional processing is needed to ensure the equality of all higher order moments.
Next subsection shows an example based on Gaussian clusters where the additional processing after unfolding
reduces to making the covariances equal (i.e. equalization).

\subsection{Equalization after unfolding}

\begin{figure}[t!]
\begin{tabular}{cc}
    \begin{minipage}[c]{0.5\textwidth}
          \begin{tabular}{c}
              \hspace{-0.5cm}\includegraphics[width=3.9cm]{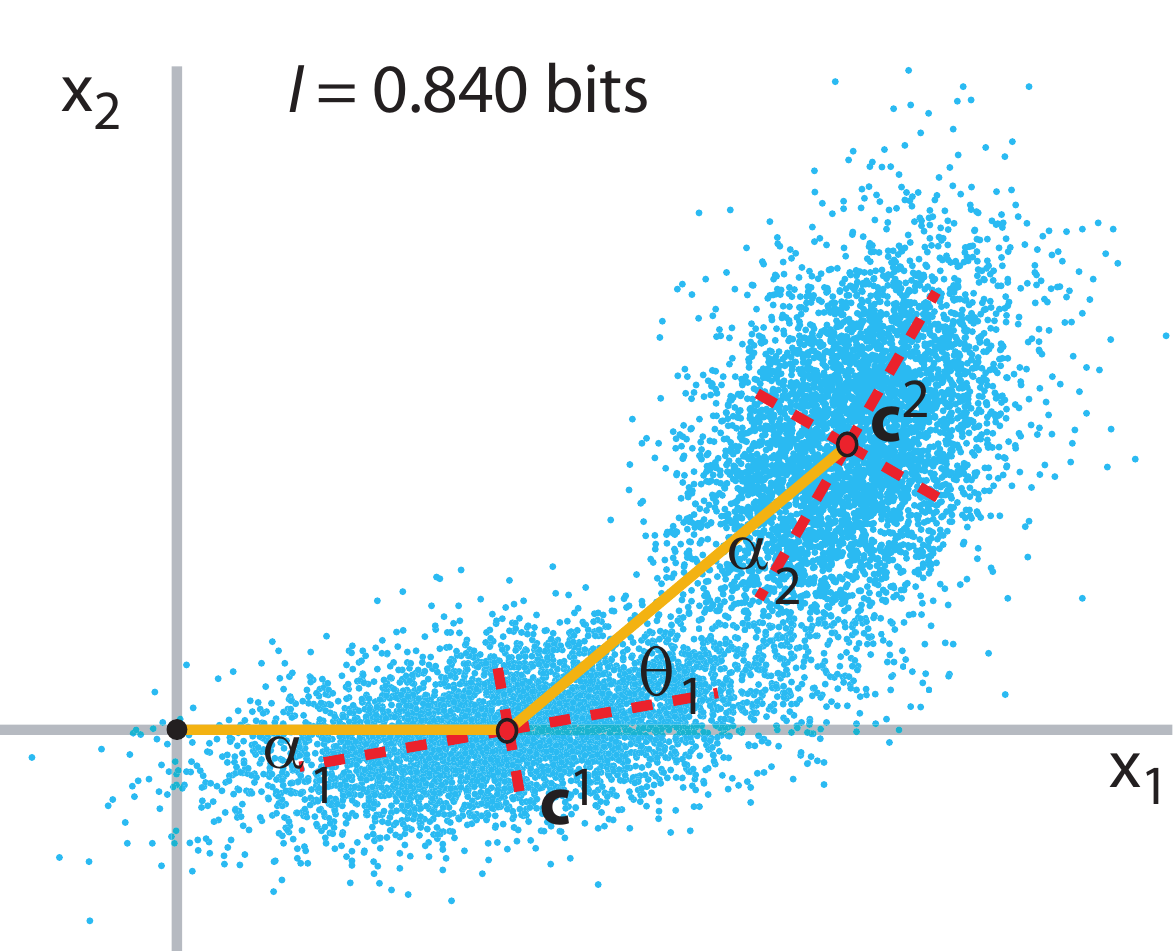}  \\
              \hspace{-0.5cm}\includegraphics[width=3.9cm]{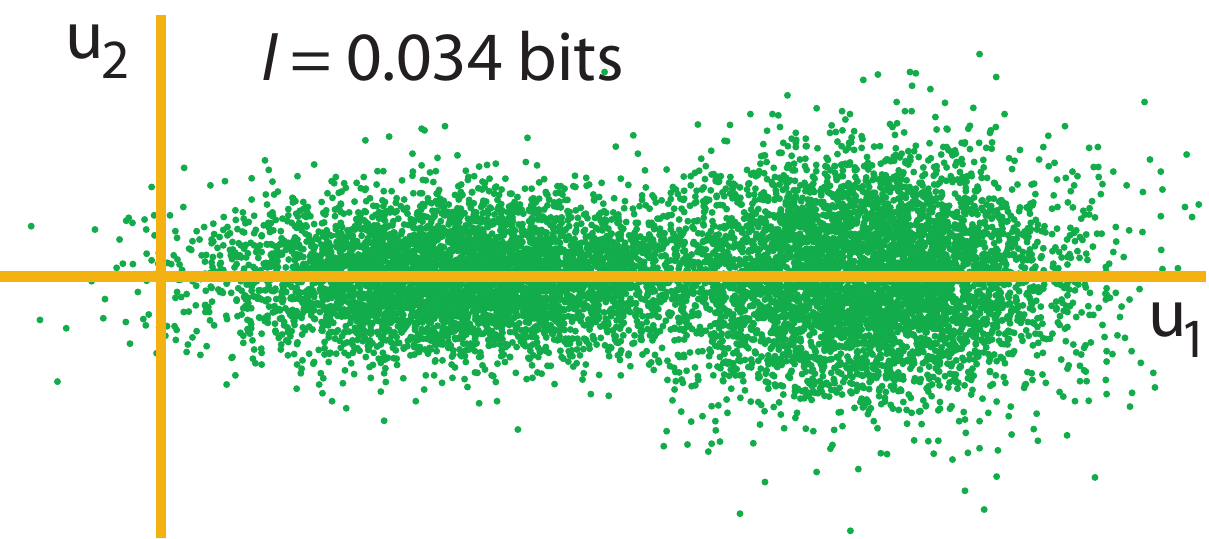}  \\
              \hspace{-0.5cm}\includegraphics[width=3.9cm]{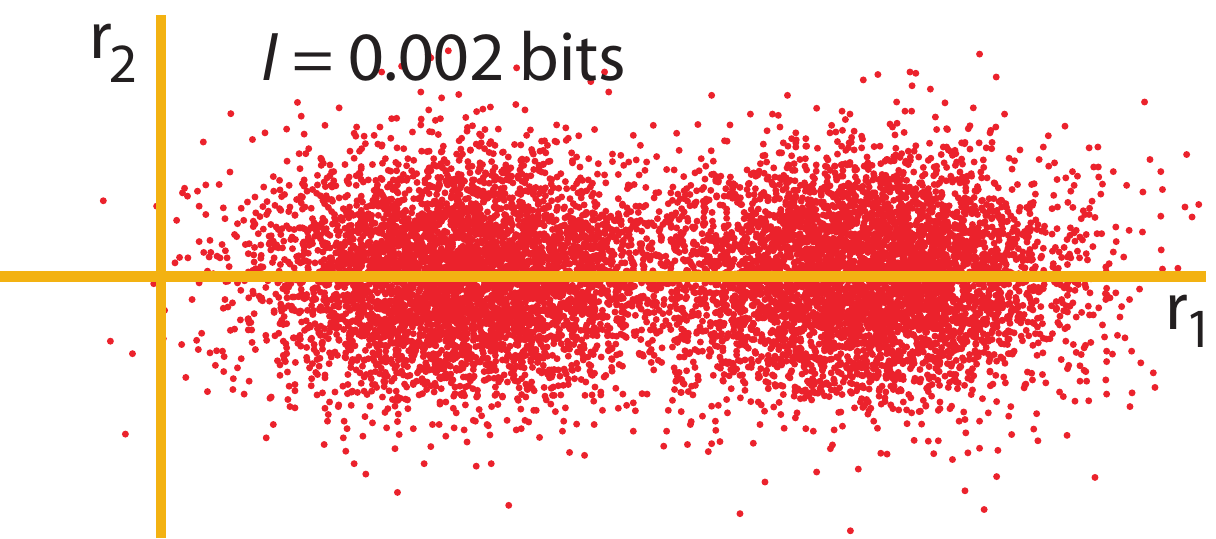}
          \end{tabular}
    \end{minipage}
    &\hspace{-6cm}
    \begin{minipage}[c]{0.5\textwidth}
          \begin{tabular}{c}
               \includegraphics[width=5.5cm]{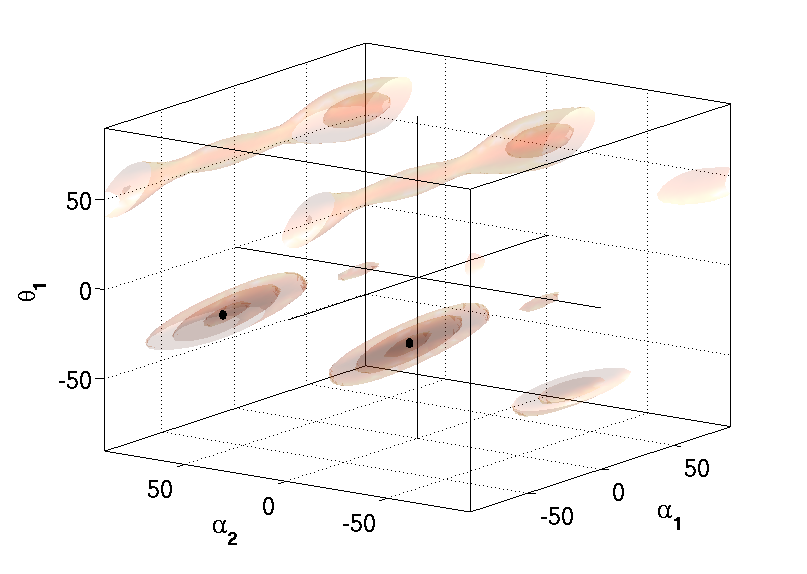} \\
            \hspace{-0.3cm}\includegraphics[width=4.1cm]{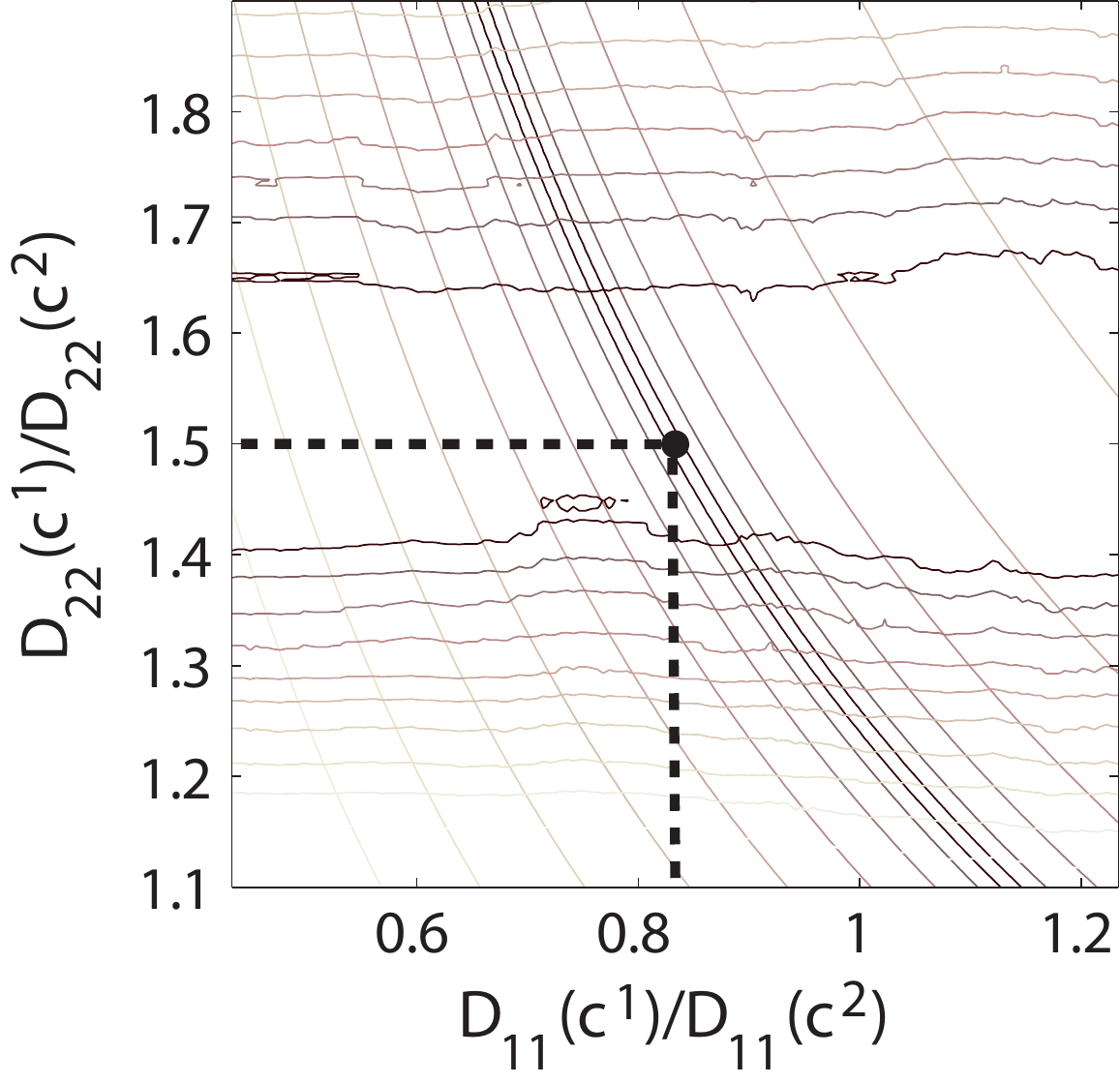}
          \end{tabular}
    \end{minipage}
\end{tabular}
    \caption{ {\em Unfolding} and {\em local equalization} are optimal in multi-information and
    in achieving uniform population {\em per} unit-volume cluster.
    Top left plot shows a simple curved manifold, which was generated from particular misalignments
    $\alpha_1=10^o$, $\alpha_2=20^o$, and $\theta_1=30^o$, and particular ratios among the standard
    deviations in the local principal directions, namely $0.83$ and $1.5$. The effect of varying the parameters
    of the unfolding transform $U$ in reducing mutual information, $\Delta I$, is shown in the right top plot
    (darker means higher $\Delta I$).
    Middle left plot shows the same data under the identified optimal $U(\mathbf{x})$, and
    bottom left plot shows the data that jointly minimize $I$ while achieve uniform density clusters.
    The bottom right plot displays $I$ and iso-density contours (see text for details).
    The latent values are correctly identified by minimizing $I$ while enforcing uniform PDF.
    Conversely, unfolding along a PC, local model alignment and local equalization minimize $I$.
    }
    \label{unfolding_effect}
\end{figure}

Here we explore the effect of unfolding and local equalization in multi information terms.
To this end we consider an elementary curved manifold made of two different local clusters
(Fig.~\ref{unfolding_effect}), but the conclusions may be extended to more complicated manifolds
made of local clusters.

By definition, the PC will go through the averages of the local clusters, i.e. the points
$\mathbf{c}^1$ and $\mathbf{c}^2$ in our elementary example. A set of concatenated rotations with appropriate angles around the points along the PC, $\mathbf{u}=U(\mathbf{x})$, will eventually unfold the PDF by aligning the averages of each local cluster as well as the axes of the local models.

The change in multi-information, $\Delta I = I(\mathbf{x})-I(\mathbf{u})$, using this family of concatenated rotations, $U$, is easy to compute~\cite{CoverThomas06}:
\begin{equation}
     \Delta I = \sum_{x_i} h(x_i) -  \sum_{u_i} h(u_i) + {\mathbb E}[ \log |\nabla U(\mathbf{x})| ].
     \label{variacion_I}
\end{equation}
Note that this reduces to the change in the sum of marginal entropies since $|\nabla U(\mathbf{x})|=1, \, \forall \mathbf{x}$
as the local behavior is just a rotation.

In this illustrative example, the family of concatenated local rotations depends on three parameters: the angles $\alpha_1$, $\alpha_2$ and $\theta_1$, and we may exhaustively look for the transform $U$ that optimizes the multi-information reduction.
We computed the sum of marginal entropies, or $\Delta I$, for the whole family of transforms $U$:
see the isosurfaces of $\Delta I$ in the parameter spaces at the top right. This example clearly indicates
that the maximum reduction in $I$ (parameters identified by black dots in the figure) also leads
to unfolding and model alignment (center left figure).

However, the variance of each local cluster may still depend on the location along the unfolded PC. This residual dependence can be removed by locally expanding or contracting the unfolded domain through an appropriate point dependent metric: $\mathbf{r} = D(\mathbf{x}) \cdot U(\mathbf{x})$. Horizontal contour lines at the bottom right plot show the $I$ values for different combinations of metric changes applied at each cluster.
Multi-information by itself does not completely constrain the line element along the unfolded dimension.
However, if one additionally requires constant probability in the transformed domain (clusters of uniform volume), the solution is unique
as shown by the combination of multi-information contours and the curved contours indicating uniform volume clusters.
The minimum $I$ condition together with the equal volume condition (black dot) is obtained when the first cluster is vertically expanded to have the same height as the second cluster.
As intuitively expected, the optimal local expansion or contraction in $I$ terms and in achieving constant volume regions is the one that makes the clusters equal (bottom left figure).

In a more general $d$-dimensional scenario, the required additional processing after unfolding along a Principal Curve
could be setting the line element for local equalization in \emph{every direction} of
the orthogonal subspaces. This would achieve a constant (uniform) PDF along the curve thus
ensuring the equality of all higher order moments. In 2-$d$ settings (as the one above) orthogonal subspaces are 1-$d$
and the additional equalization required after unfolding can be easily solved. Unfortunately, in general,
orthogonal subspaces have dimension $(d-1)$, and the solution of multivariate equalization is not
unique \cite{Hyvarinen99,Laparra10rbig}.
A suboptimal alternative is performing the required equalization sequentially: following one
secondary PC at a time.
Therefore, unfolding and point-dependent equalization can be extended dimension-wise by sequentially drawing additional locally orthogonal PCs.

Even though a sequential (dimension-wise) procedure does not guarantee complete independence, equalization
along secondary PCs is more sensible than using arbitrary linear directions since, according
to their definition \cite{Delicado01}, they capture the main structure in the orthogonal subspaces.
Moreover, the benefits of unfolding and point-dependent metric suggested so far, trivially generalize to more complicated manifolds that can be locally decomposed into clusters, as commonly assumed in the literature~\cite{Roweis02,Verbeek02,Teh03,Brand03,Malo06b}.

 Accordingly, {\em unfolding along PCs} and the use of {\em local metrics} will be the basic processing elements of the proposed technique.

\section{Sequential Principal Curves Analysis} \label{spca}

The proposed transform consists of (1) certain differential behavior, and
(2) certain integration path for this Jacobian.

\subsection[Jacobian]{Proposed differential behavior (Jacobian).}
Local unfolding and equalization along PCs,
can be expressed as:
\begin{equation}
\nabla R({\bf x}) = D({\bf x}) \cdot \nabla U({\bf x})
\label{generalSPCA_jacobian}
\end{equation}
where ${\bf u}=U({\bf x})$ is the unfolding transform that consists of concatenated local rotations
along the proposed path made of a sequence of PCs, and the diagonal matrix $D({\bf x})$ represents the local length of the line element along this path (change of metric). Note that $\nabla U({\bf x})$ is orthonormal for all ${\bf x}$ since the unfolding $U$ can be formulated as a set of concatenated local rotations.
In fact, in agreement with~\cite{Delicado01,Einbeck05,Einbeck10}, the method we (instrumentally) use here to draw local-to-global PCs relies on
local PCA to estimate the tangent to the curve (see Appendix). Note that using local-PCA to compute PCs is consistent with the assumption
of local Gaussian clusters, which is usual in the literature~\cite{Roweis02,Verbeek02,Teh03,Brand03,Malo06b}.

In order to adapt the metric to the density, we set the elements of $D$ using the marginal PDF on the unfolded coordinates and an appropriate exponent $\gamma\geq 0$:
\begin{equation}
     D({\bf u})_{ii} \propto p_{u_i}(u_i)^\gamma
     \label{generalSPCA_scaling}
\end{equation}
where each marginal PDF is estimated following $k$-neighbors rule.
The induced metric is: $M({\bf x}) =  \nabla U({\bf x})^\top \cdot D({\bf x})^2 \cdot \nabla U({\bf x})$.
Assuming that local clusters can be factorized by the local rotations (e.g. $\nabla U({\bf x})$ are local PCAs),
we have:
\begin{eqnarray}
\begin{array}{ll}
|M({\bf x})| & = |D({\bf x})|^2 \propto \prod_{i=1}^{n} p_{u_i}(u_i)^{2\gamma} = p({\bf u})^{2\gamma}
\\[3mm]
& = p({\bf x})^{2\gamma} |\nabla U({\bf x})|^{-2\gamma} = p({\bf x})^{2\gamma},
     \label{determinant_SPCA_metric}
\end{array}
\end{eqnarray}
which is the behavior required in Eq.~\eqref{metrica} and encompasses both \emph{infomax} and \emph{error minimization} depending
on $\gamma$.


\subsection[Integration path]{Proposed integration path.}
Given an arbitrary origin
on the first PC, ${\bf x}^o$, assumed to give zero response,
${\bf r}^o = {\bf 0}$, and some point of interest, ${\bf x}$; the transform $R(\bf{x})$ will be given by the integration of the
proposed Jacobian along certain integration path from ${\bf x}^o$ to ${\bf x}$.
As illustrated in Fig. \ref{digrama2}, the proposed path is made of segments of the \emph{first} and \emph{secondary} PCs.
The \emph{first} PC is just the standard PC of the set~\cite{Hastie89}, while \emph{secondary} PCs were introduced in~\cite{Delicado01}.
In the proposed path, the $i$-th PC is followed up to the geodesic projection of the point ${\bf x}$ on this PC, ${{\bf x}_{\bot}^i}$.
Here geodesic projections are understood as \emph{projections that follow the local structure of the manifold}.
Geodesic projections are obtained from orthogonal projections according to the procedure described later.

%
%

\begin{figure*}[t!]
\begin{center}
\includegraphics[width=16.5cm]{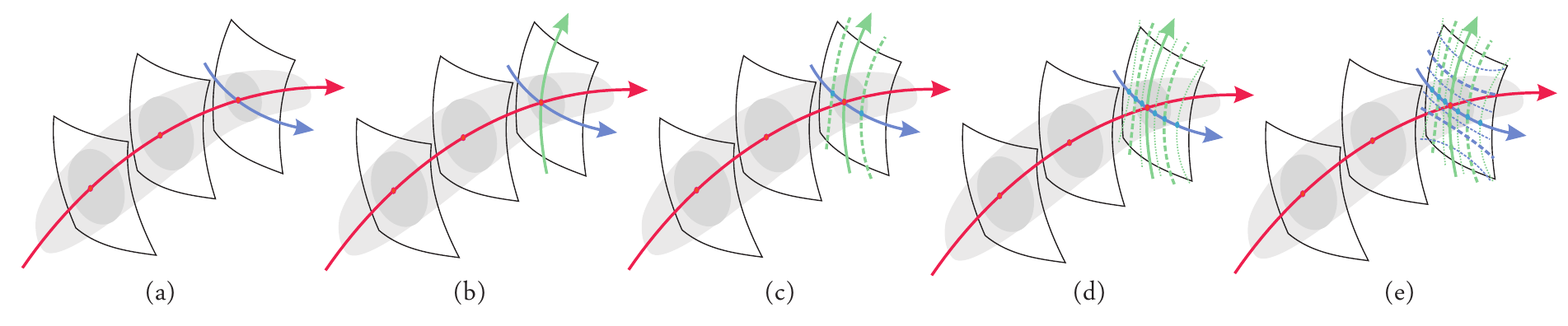}
\end{center}
\vspace{-0.3cm}\caption{ Curved locally orthogonal subspaces from secondary PCs and the appropriate PDF-dependent metric. Delicado's secondary PCs (e.g. second PC in blue and third PC in green in \emph{a} and \emph{b}) capture the structure in the orthogonal subspace with regard to the Hastie's PC (in red). A locally orthogonal curved subspace can be defined by considering additional 3rd PC curves at \emph{certain} points on the 2nd PC (dashed and dotted green curves in \emph{c} and \emph{e}).
Taking these points to be separated by constant cumulated density, as would be given by constant steps in
$r_2$ (or integrated metric in 2nd dimension), the 3rd PCs divide the subspace in regions of constant population.
If the same procedure is applied on the remaining dimensions (blue PCs in \emph{e}), one obtains the lattice with constant
population in each non-uniform cell, as in Fig. \ref{digrama2}.}
\label{diagrama3}
\end{figure*}

%

\subsection{Direct Transform.}
The SPCA transform of $\bf{x}$ is given by integrating the proposed $\nabla R$ along the proposed path:
\begin{equation}
{\bf r} = R({\bf x}) = C \cdot \int_{{\bf x}^o}^{{\bf x}} \nabla R({\bf x'}) \cdot d{\bf x'} = C \cdot \int_{{\bf x}^o}^{{\bf x}} D({\bf x'}) \cdot \nabla U({\bf x'}) \cdot d{\bf x'} \\
\label{global_response1}
\end{equation}
where $C$ is just a constant diagonal matrix that independently scales each component of the response.
The selected path implies displacements in one PC at a time. According to this, the vector $d{\bf u'} =\nabla U({\bf x'}) d{\bf x'}$ has only one non-zero component: the one corresponding to the considered PC at the considered segment. Therefore, the response of each sensor to the point ${\bf x}$ is just the length on each PC in the path,
measured according to the metric related to the local PDF with the selected $\gamma$,
\begin{equation}
r_i = C_{ii} \cdot \int_{{\bf x}_{\bot}^{i-1}}^{{\bf x}_{\bot}^i} D({\bf x'}) \cdot \nabla U({\bf x'}) \cdot d{\bf x'} =  C_{ii} \int_{0}^{u_{i \bot}^i} p_{u_i}(u'_i)^\gamma \, du'_i \label{global_response2}\\
\end{equation}
The scaling constants, $C_{ii}$, are a global response ranking.

SPCA is initialized by setting (i) the origin of the coordinate system, and (ii) the scale of the different dimensions and the order in which they will be visited by the sequential algorithm. Sensible choices for the origin are those suggested in other local-to-global PC algorithms \cite{Delicado01,Einbeck05,Einbeck10}: the most dense point of the distribution (if known) or the mean of the data. Then a set of $d$ locally orthogonal PCs is drawn at the selected origin, which will be used to set the order and the relative scale of the dimensions. In our case, we set the scaling constants $C_{ii}$ according to an information distribution criterion: we use the number of quantization bins {\em per} dimension given by classical bit allocation results in transform coding~\cite{Gersho92}, i.e higher marginal entropy first.
Other criteria could be used, as for instance the total standard deviation of the projected data (as in global PCA) or the total Euclidean length of the curvilinear axes.

\subsection{Inverse Transform.}
Given a set of training samples from the source, the origin in the input space, ${\bf x}^o$, and
the initialization (dimension order and scale), the computation of the inverse, ${\bf x}=R^{-1}({\bf r})$, is simple.
It just involves drawing the first PC through the origin and taking the length $r_1$ on this curve, measured according
to $p_{u_1}(u_1)^\gamma$. Displacement on the first curve by the length $r_1$ leads to the first projection ${{\bf x}_{\bot}^1}$.
Then, the second locally orthogonal curve is drawn from ${{\bf x}_{\bot}^1}$, and one takes a second displacement $r_2$ on
this second PC leading to the second projection, ${{\bf x}_{\bot}^2}$. This process is repeated sequentially in every dimension
until the point ${\bf x}$ is found by taking the displacement $r_d$ from ${\bf x}^{d-1}_{\bot}$ on the {\em d}-th PC.
\vspace{0.15cm}

\emph{Out-of-sample.} Note that the transform can be easily applied to new samples (eqs. \ref{global_response1} and \ref{global_response2}) as long as they are not too far from the training samples, i.e. as long as they come from the same source.
\vspace{0.15cm}

The reminder of the section is devoted to show how the concept of \emph{secondary PCs} in~\cite{Delicado01}
together with the local metric proposed here can be used to define (1) locally orthogonal subspaces, and (2) projections following
the local structure of the manifold, which we called geodesic projections above.
These are the ingredients behind the
proposed SPCA path. 

\subsection{Locally orthogonal subspaces.} Hastie's {\em first} PC summarizes the whole dataset \cite{Hastie89}. Then, one may consider hyperplanes orthogonal to this first PC and local basis at those hyperplanes to describe the residuals. In principle, any set of ($d-1$) linearly independent vectors living in the corresponding hyperplane would suffice. However, as noted by Delicado~\cite{Delicado01}, the residual at those hyperplanes may also have a curved structure. Therefore, it makes sense to draw a secondary PC at the hyperplane to capture this
structure (e.g. blue curve in Fig. \ref{diagrama3}.a).
Fig. \ref{diagrama3} shows how the secondary PCs idea together with the proposed PDF-dependent metric can
be used to define the lattice assumed in Fig. \ref{digrama2}.

Even though SPCA does not require its explicit computation, such underlying lattice following the structure
of the PDF illustrates the equalization properties of the SPCA representation.
Delicado lists a set of conditions for these secondary PCs to exist \cite{Delicado01}.
A rigorous application of SPCA should start by
checking that Delicado's conditions hold for the dataset at hand.
In Section \ref{experiments} we take an empirical approach
and show that in practice SPCA does induce the kind of PDF-dependent
curvilinear lattices assumed in this section.

\subsection{Geodesic projections.}
\begin{figure}[t!]
\begin{center}
\begin{tabular}{c}
\hspace{-0.25cm}
\includegraphics[width=9cm]{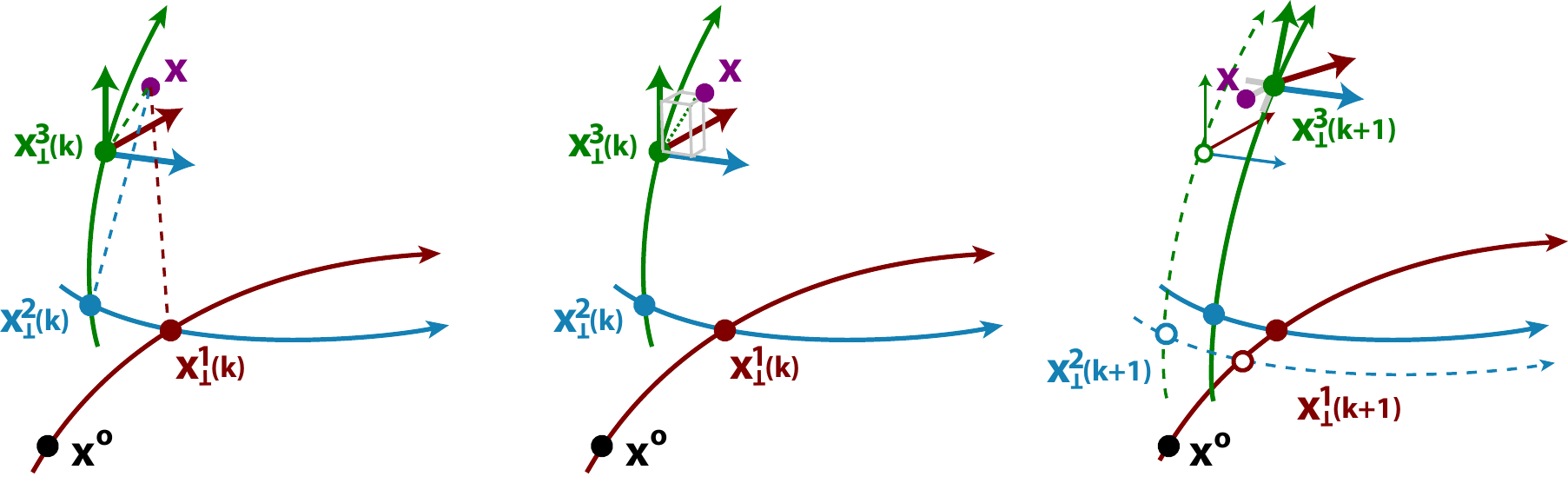}
\end{tabular}
\end{center}
\vspace{-0.3cm}\caption{ Iterative procedure to find the geodesic projections of the point to be transformed ${\bf x}$. In the first iteration
the orthogonal projection is used as starting point for each secondary PC of the sequence (left plot). However, in general, the target point is not reached by drawing secondary PCs from the orthogonal projections. The residual from attempt $k$ to reach the point, ${\bf x}_\bot^d(k)-{\bf x}$, can be expressed in the local linear coordinates at the approximated point (center plot). The components of the residual in the local representation can be used to correct the position of the estimated projection on the curves, Eq. \ref{iterative}.
At the new \emph{corrected} projections, new local secondary PCs are drawn. The resulting path gets closer to the target (right plot).}
\label{diagrama4}
\end{figure}

In order to reach the target $\bf{x}$ following the structure of the manifolds (e.g. through geodesics)
Delicado's \emph{secondary} PCs are drawn from certain projections on the 1st, 2nd, ... PCs.
If orthogonal projections are used,
in general the target is not reached by the PCs.
Fig. \ref{diagrama4} describes the proposed iterative procedure to reach the target.
The idea is modifying the current projections (or lengths $r_i$) to reduce the approximation error.
At iteration $k$ we have:
\begin{eqnarray}
\begin{array}{ll}
{\bf r}(k+1) = {\bf r}(k) + \alpha \, \boldsymbol{\delta}(k)
\\[1mm]
\boldsymbol{\delta}(k) = C \cdot D({\bf x}_\bot^d(k)) \cdot \nabla U({\bf x}_\bot^d(k)) \cdot ({\bf x}_\bot^d(k)-{\bf x})
     \label{iterative}
\end{array}
\end{eqnarray}
Results in synthetic and real manifolds in section \ref{experiments} confirm the practical convergence of the geodesic projection thus leading to accurate transforms and inverses.

\section{Experiments and Examples}\label{experiments}

In this section we focus on the intrinsic properties of SPCA which are independent
of the algorithm to draw individual PCs\footnote{The particular
algorithm chosen to draw individual PCs will certainly have an impact in SPCA results,
but its relevance is merely instrumental. Details on our particular choice to draw individual PCs,
the effect of its parameters, and the procedure to set them, can be found at the
Appendix. However, note that other local-to-global algorithms to draw individual PCs (e.g.
PCs in \cite{Delicado01,Einbeck05,Einbeck10}) would also be suitable for SPCA.}.
According to this, in each experiment we assume that that the algorithm chosen to draw individual
PCs is tuned to the manifold at hand. In our case, this reduces to the minimization
of the projection error (see Appendix).
An implementation of SPCA with worked examples reproducing all the results of the work is available at http://isp.uv.es/spca.html.

The foundations of SPCA and its applications are analyzed in four sets of
experiments:
\begin{itemize}
  \item \textbf{Underlying assumptions in SPCA}. These experiments are intended to
  illustrate that the statements in Section \ref{spca} actually hold in practice:
        (1) the concept of secondary PCs of Delicado can be used to define locally orthogonal subspaces and
        geodesic projections, and (2) the iterative procedure to get the geodesic projections converges giving
        rise to a transform with accurate inverse. We check this in synthetic data and in two kinds of real data:
        color manifolds and image (spatial texture) manifolds.
  \item \textbf{Unfolding, Nonlinear ICA and Transform Coding}.
        These experiments show that tuning the SPCA metric leads either to (1) plain unfolding,
        (2) identification of nonlinear independent components,
        or (3) optimal transform coding. We check this both in synthetic data and in image texture data.
  \item \textbf{Dimensionality reduction}. The performance of SPCA is compared to standard manifold
        learning techniques in a noisy swiss roll.
  \item \textbf{Domain Adaptation}. The performance of SPCA to find a meaningful canonical representation for manifold
        alignment is compared to PCA + whitening in a color constancy application.
  \item \textbf{Classification: generalization of Mahalanobis distance}.
        The manifold dependent metric associated to SPCA can be used in classification problems (e.g. k-nearest neighbors) as alternative to Mahalanobis distance.
\end{itemize}

\subsection{Underlying assumptions in SPCA}

\begin{figure*}[t!]
\begin{center}

\setlength{\tabcolsep}{1pt}
\begin{tabular}{cccc}
\includegraphics[width=3.2cm]{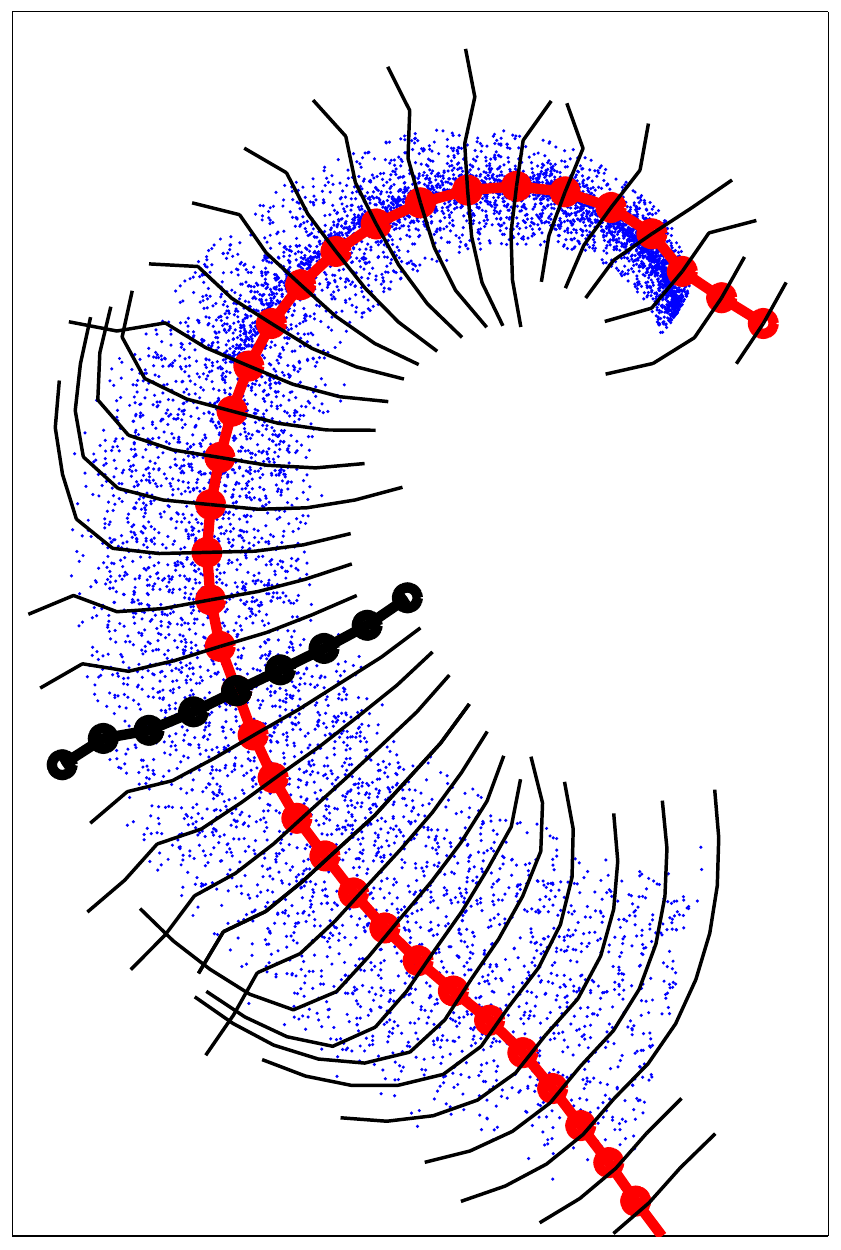} &
\includegraphics[width=3.2cm]{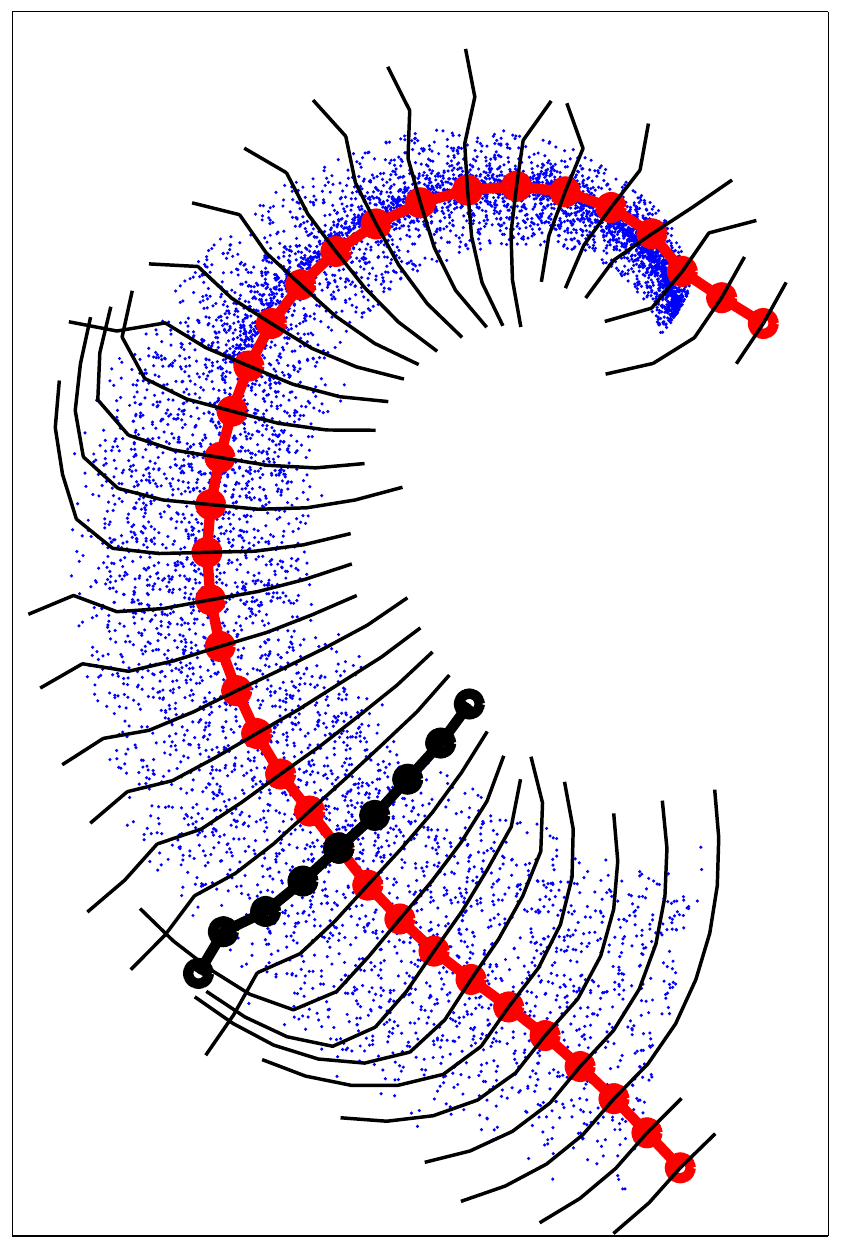} &
\includegraphics[width=3.2cm]{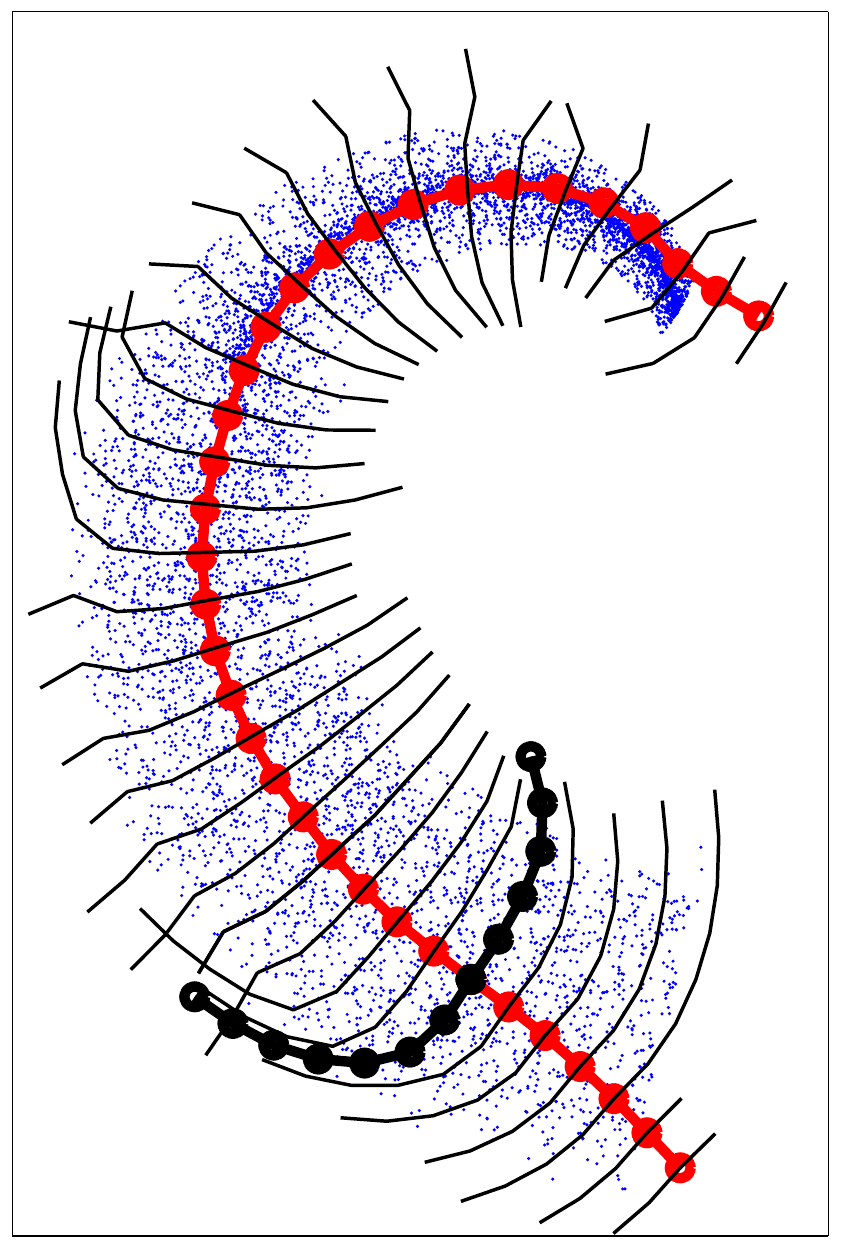} &
\includegraphics[width=3.2cm]{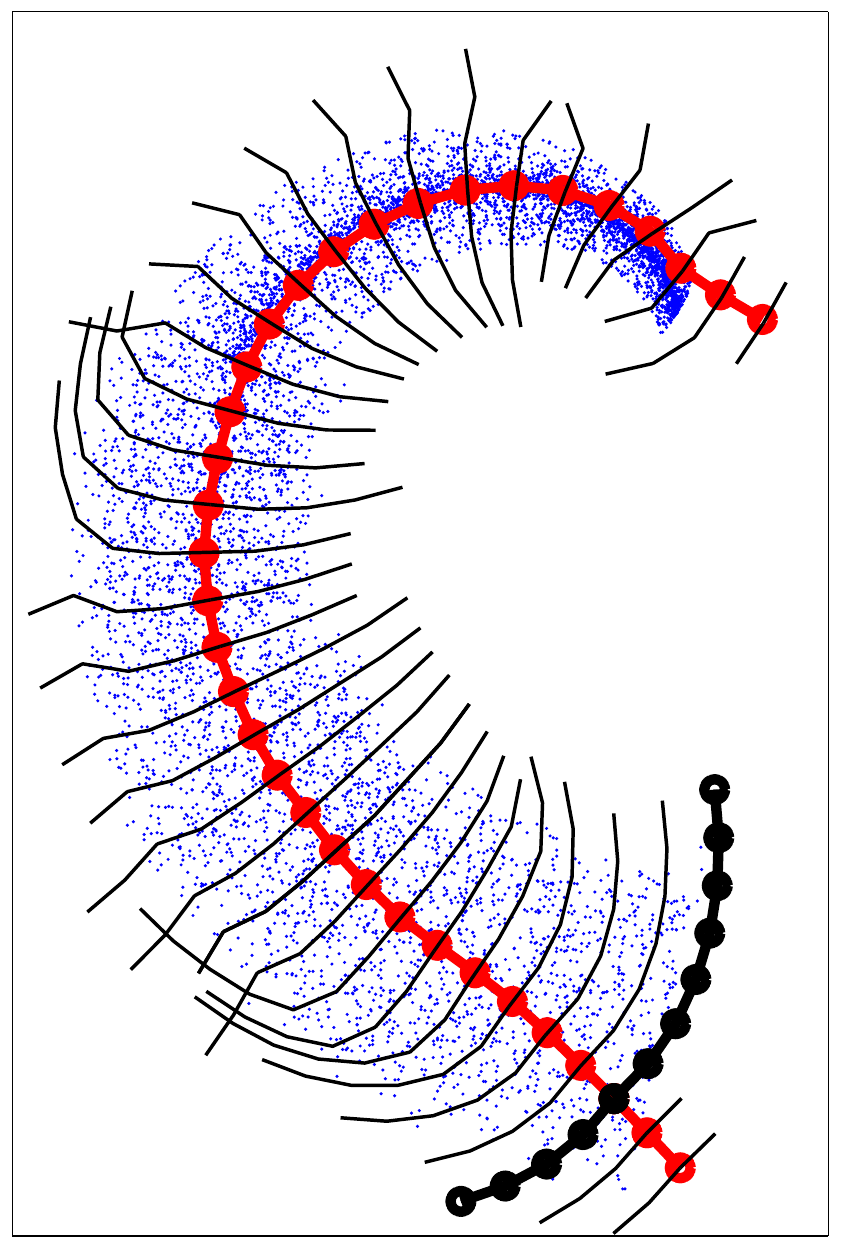} \\
\end{tabular}
\end{center}
\vspace{-0.25cm}
\caption{ Locally orthogonal subspaces (in black) and geodesic projections (e.g. along black curves) are independent of the selected origin of coordinates (highlighted dots on the red curve).
}
\label{subspaces}
\end{figure*}

\begin{figure*}[t!]
\begin{center}

\setlength{\tabcolsep}{1pt}
\begin{tabular}{ccc}
\includegraphics[width=5.8cm]{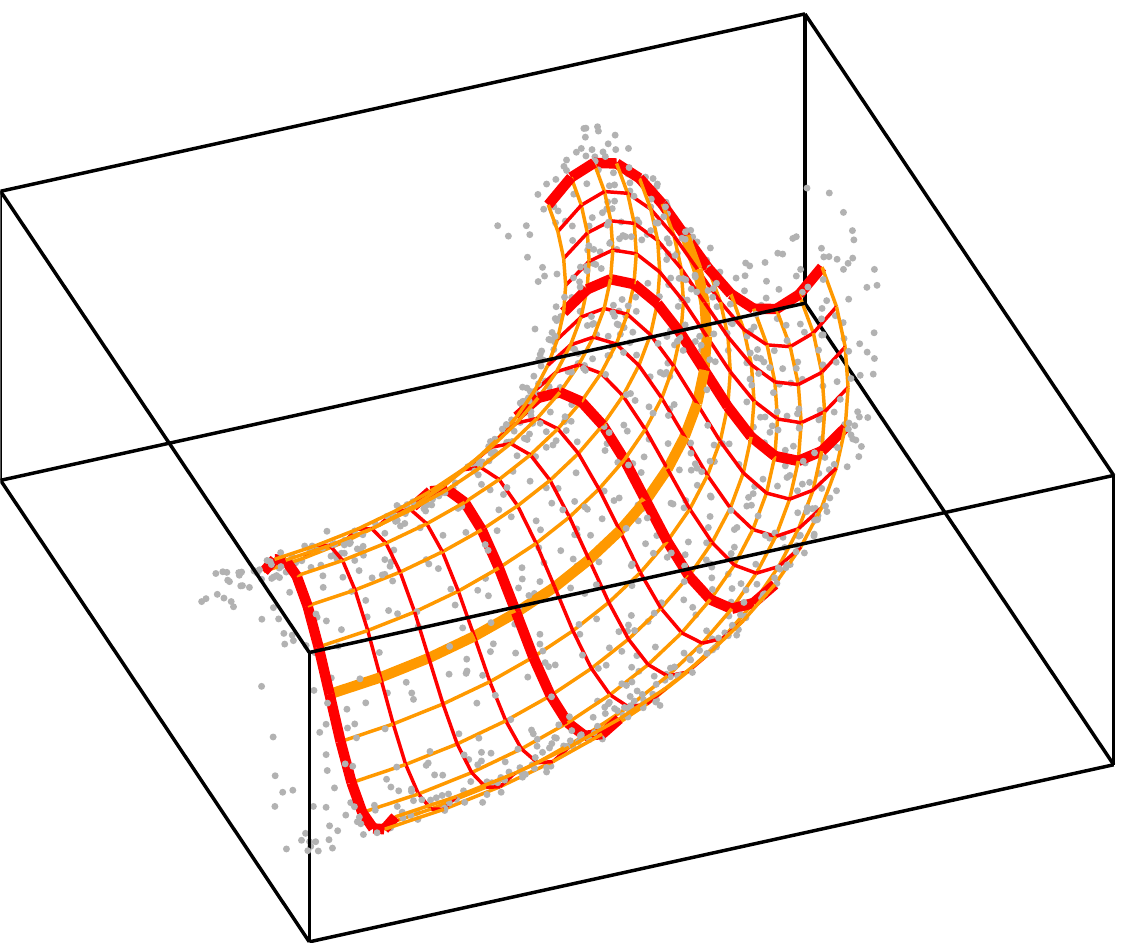} &
\includegraphics[width=5.8cm]{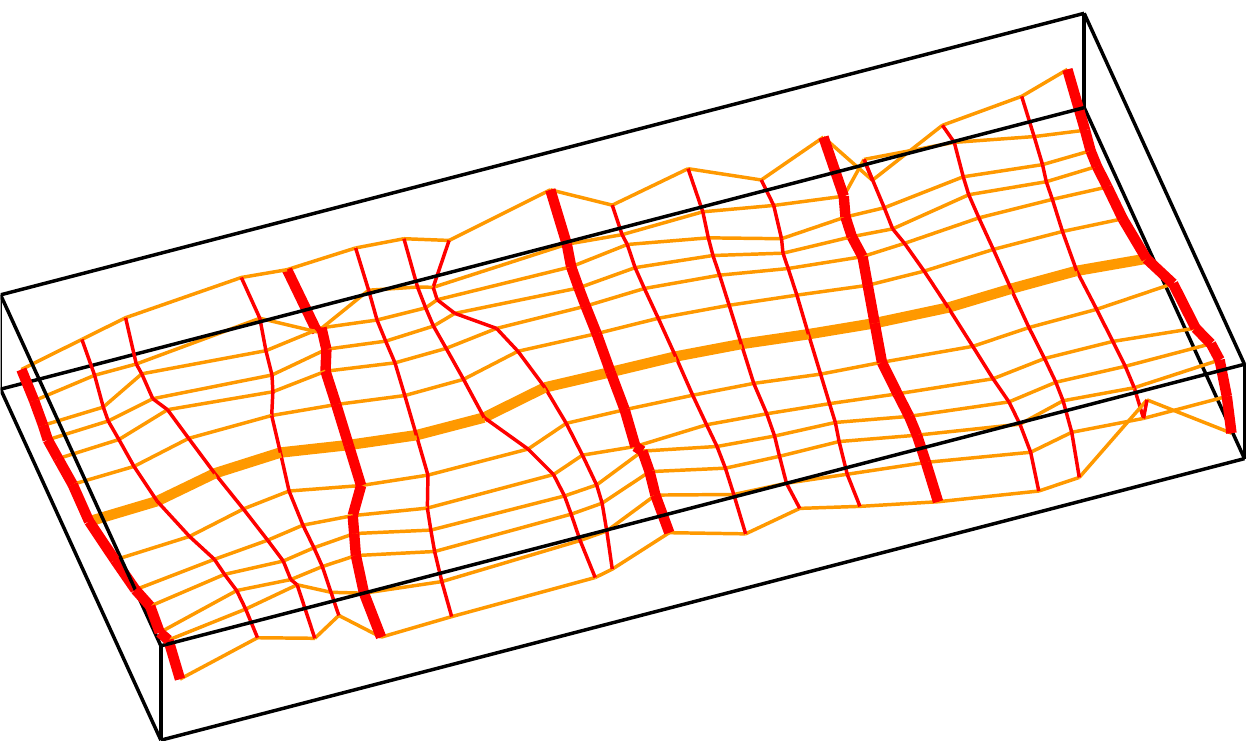} &
\includegraphics[width=5.8cm]{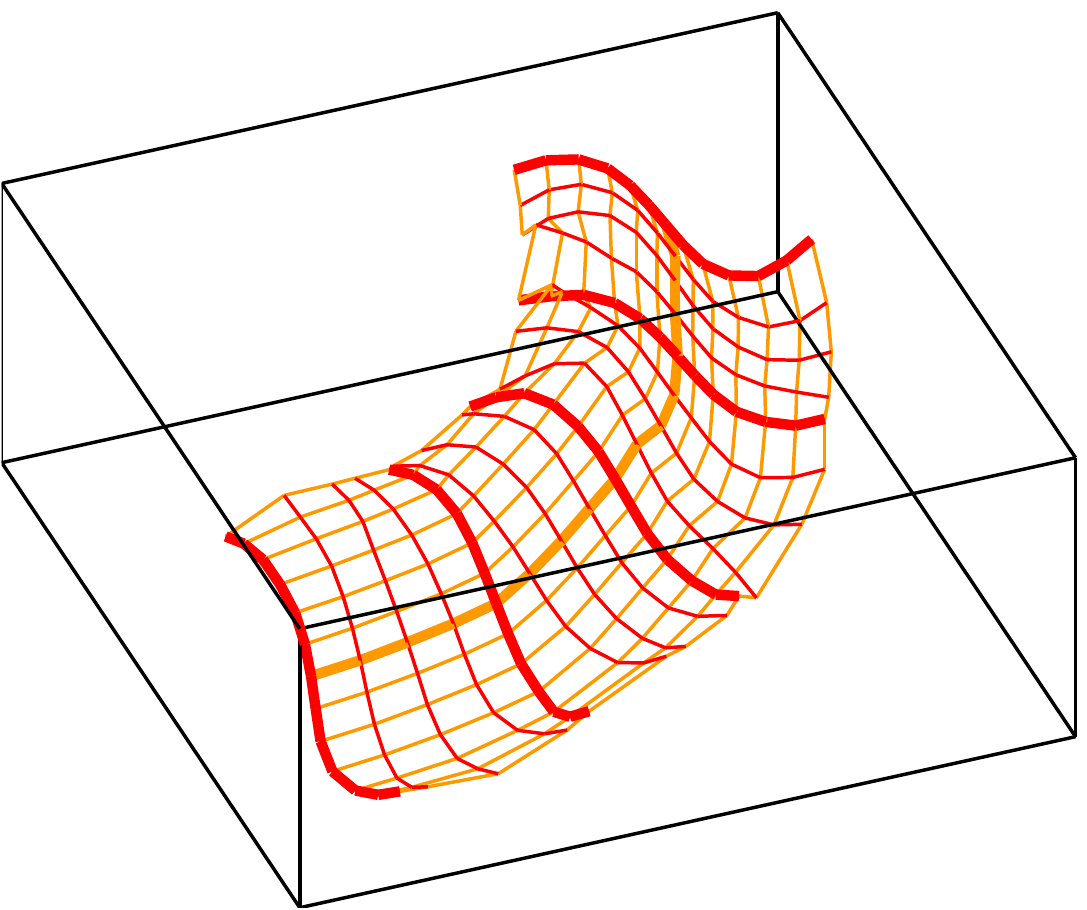} \\
\includegraphics[width=5.8cm]{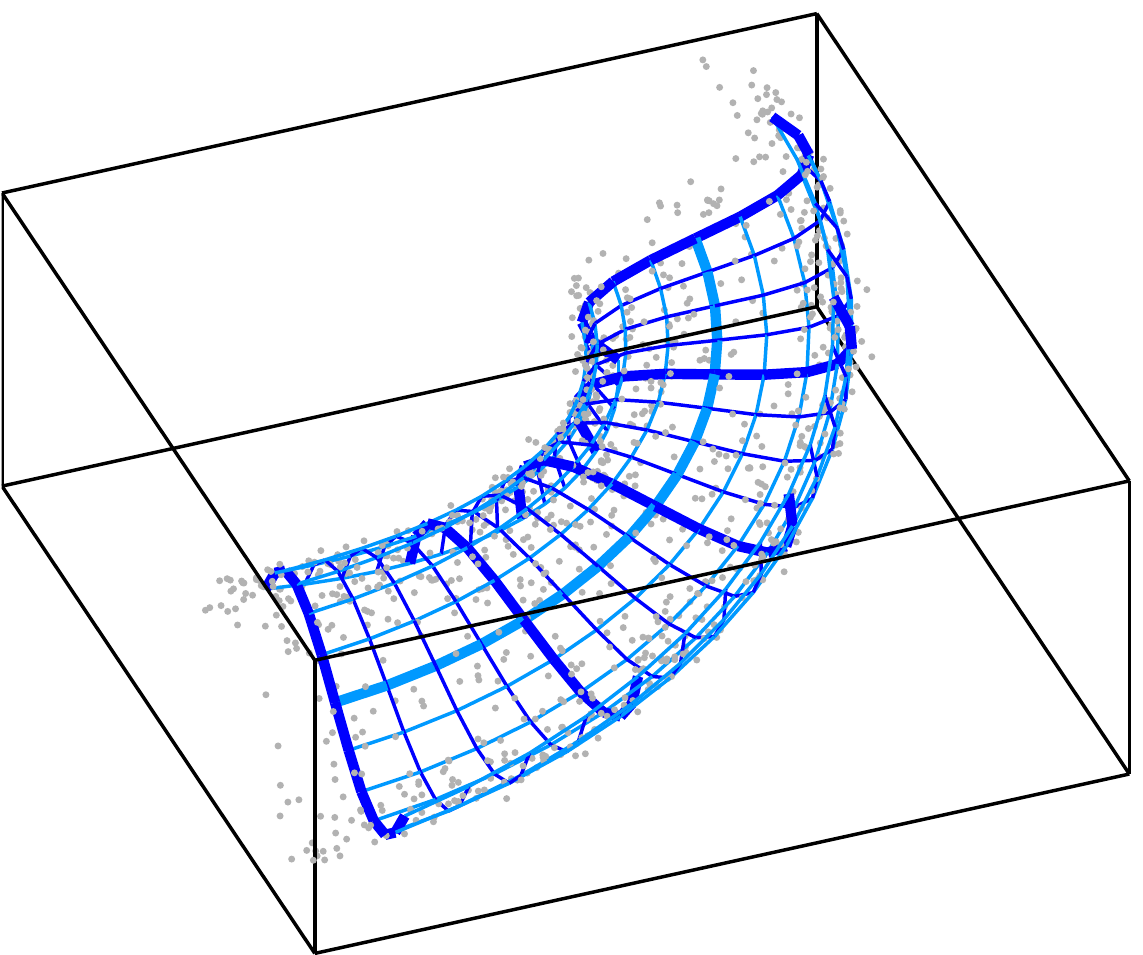} &
\includegraphics[width=5.8cm]{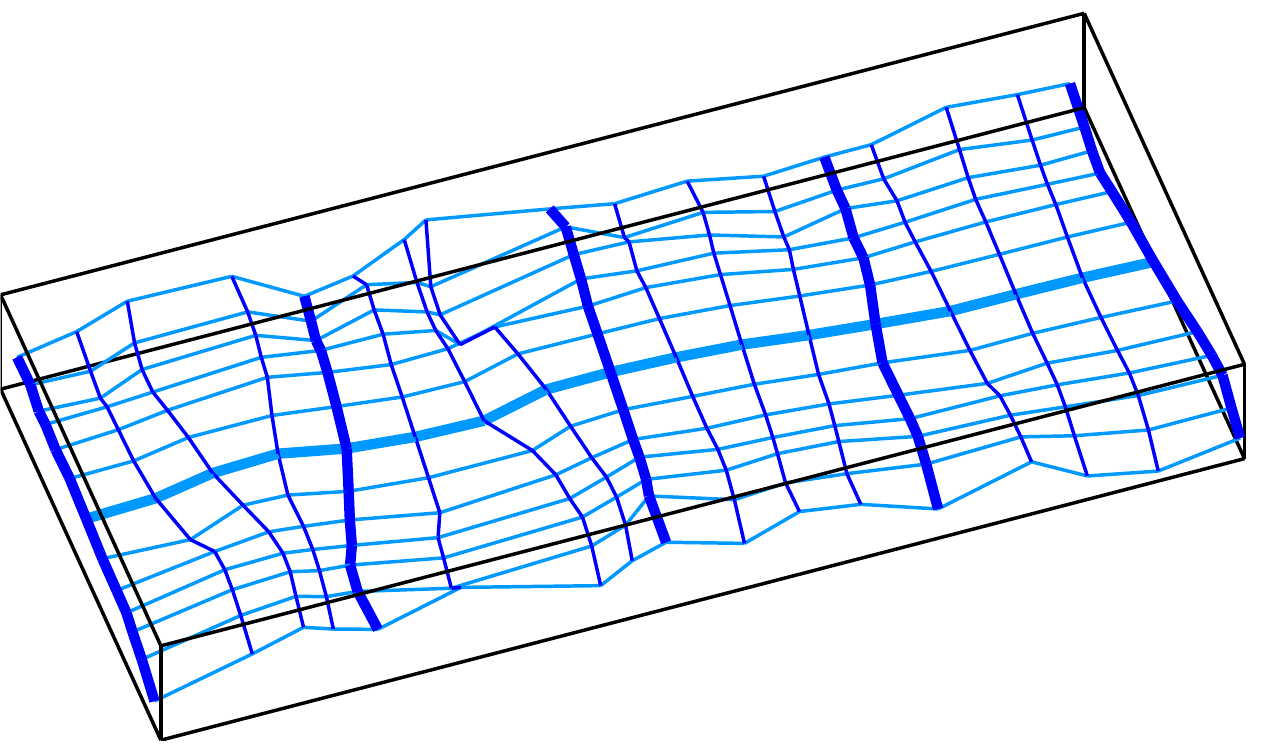} &
\includegraphics[width=5.8cm]{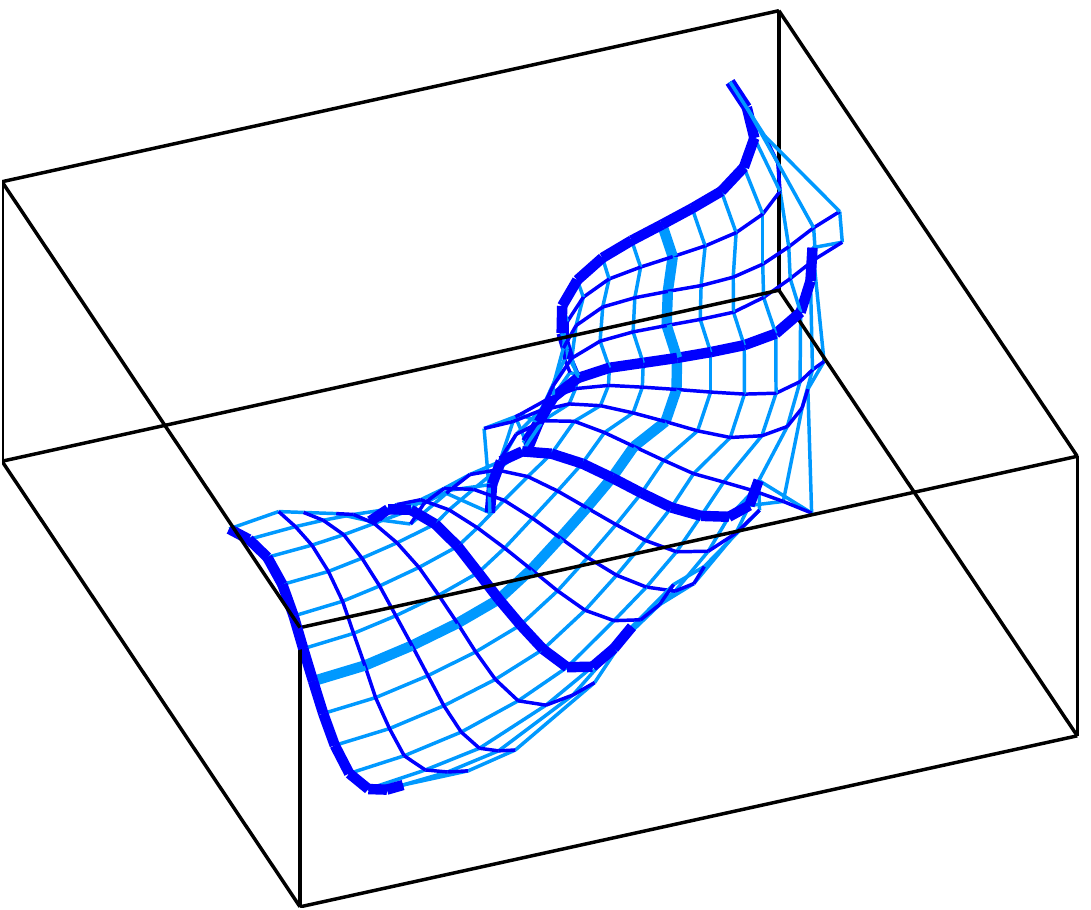} \\

\end{tabular}
\end{center}
\vspace{-0.25cm}
\caption{ Locally orthogonal subspaces (curves in dark color) along the first Principal Curve (bold curve in light color) in Delicado's manifolds \cite{Delicado01}: the one in red with fixed locally orthogonal subspaces and the one in blue with twisted orthogonal subspaces.
\emph{Left:} Theoretical model and noisy data.
\emph{Center:} Original features in the SPCA transform domain (unfolding).
\emph{Right:} SPCA features (cartesian reticle in the transform domain) back in the input domain.
}
\label{subspaces3}
\end{figure*}

\textbf{Locally orthogonal subspaces and geodesic projections.}
Figures \ref{subspaces} and  \ref{subspaces3} shows 2D and 3D examples of how the described
SPCA procedure identifies curved subspaces which are locally orthogonal to the first PC.

In Fig. \ref{subspaces} we synthesized samples using marginal
PDFs of increasing variance along a spiral. In a part of the spiral we used a Laplacian marginal,
giving rise to a clear ridge in the manifold, while in the other part we used a uniform marginal.
1000 samples of such manifold were used for training.
Here we show orthogonal subspaces computed at linearly spaced steps using SPCA with
Euclidean metric ($\gamma=0$) and different origin points $\mathbf{x_0}$ on the first PC (highlighted
dot on the red curve).
Results show that (1) Euclidean metric does imply uniform spacing along the first PC (independent of
the local PDF), (2) results are fairly independent of the chosen origin $\mathbf{x_0}$, and (3) identified
subspaces display meaningful curvature so that samples far from the first PC are projected onto
the first PC following the local PDF. Examples of Fig. \ref{gamba}, below, confirm that SPCA appropriately identifies subspaces and their distribution along the manifold for other metric choices.

Fig. \ref{subspaces3} shows examples of the identified subspaces with the non-stationary manifolds
used by Delicado to define the concept of secondary Principal Curves \cite{Delicado01}.

These results illustrate the fact that Delicado's conditions for the existence of secondary PCs
hold in practice for smooth manifolds (even displaying strong curvature)
and the fact that these secondary PCs can be used to define locally orthogonal subspaces and
geodesic projections as proposed in Section \ref{spca}.
\vspace{0.15cm}

\begin{figure*}[t!]
\begin{center}
\hspace{-1.1cm}
\begin{tabular}{c|cc|c}
Color Data & \multicolumn{2}{c|}{Spatial Texture Data} & Convergence \\
\includegraphics[width=4.5cm,height=3.4cm]{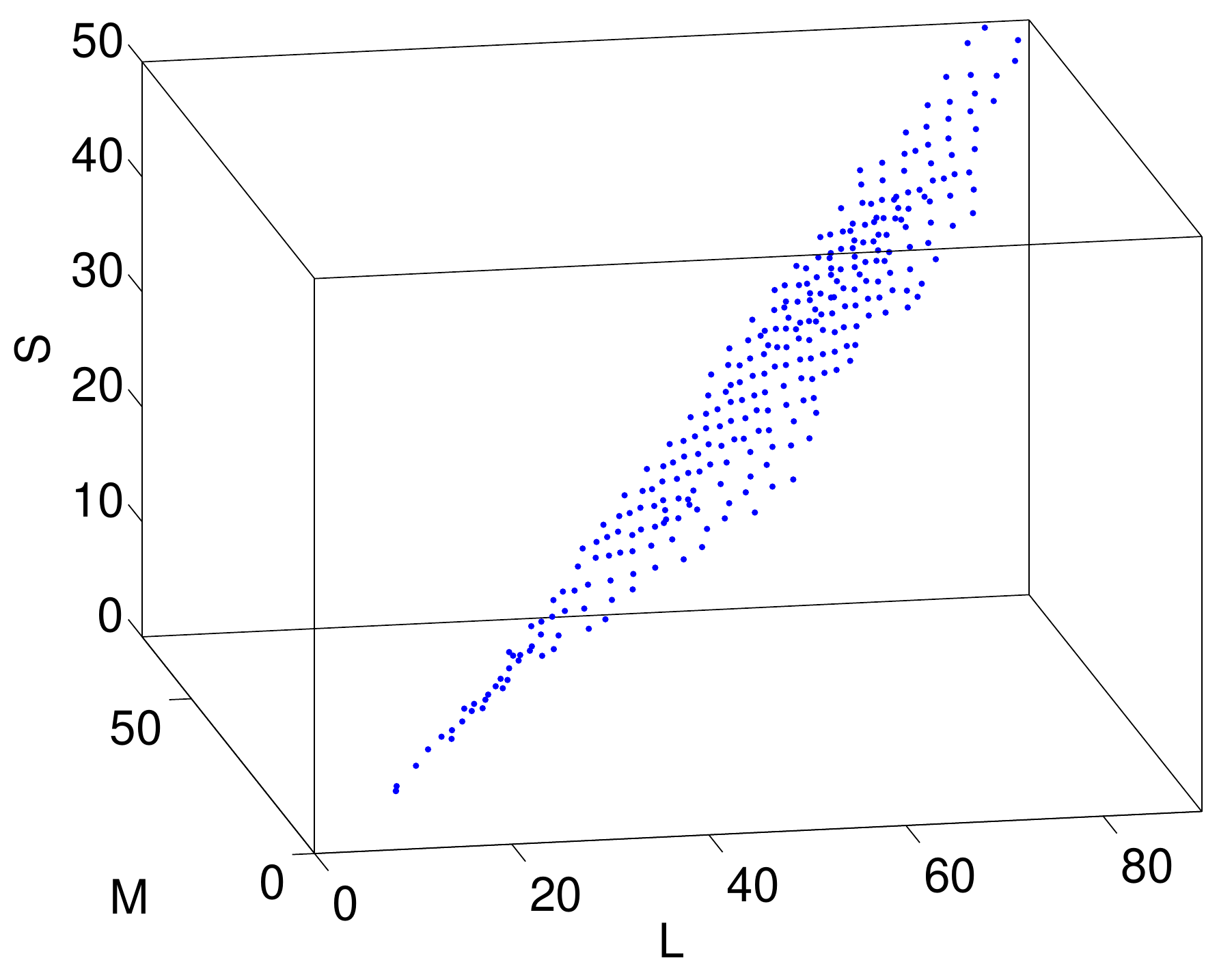} & \includegraphics[width=4.5cm,height=3.4cm]{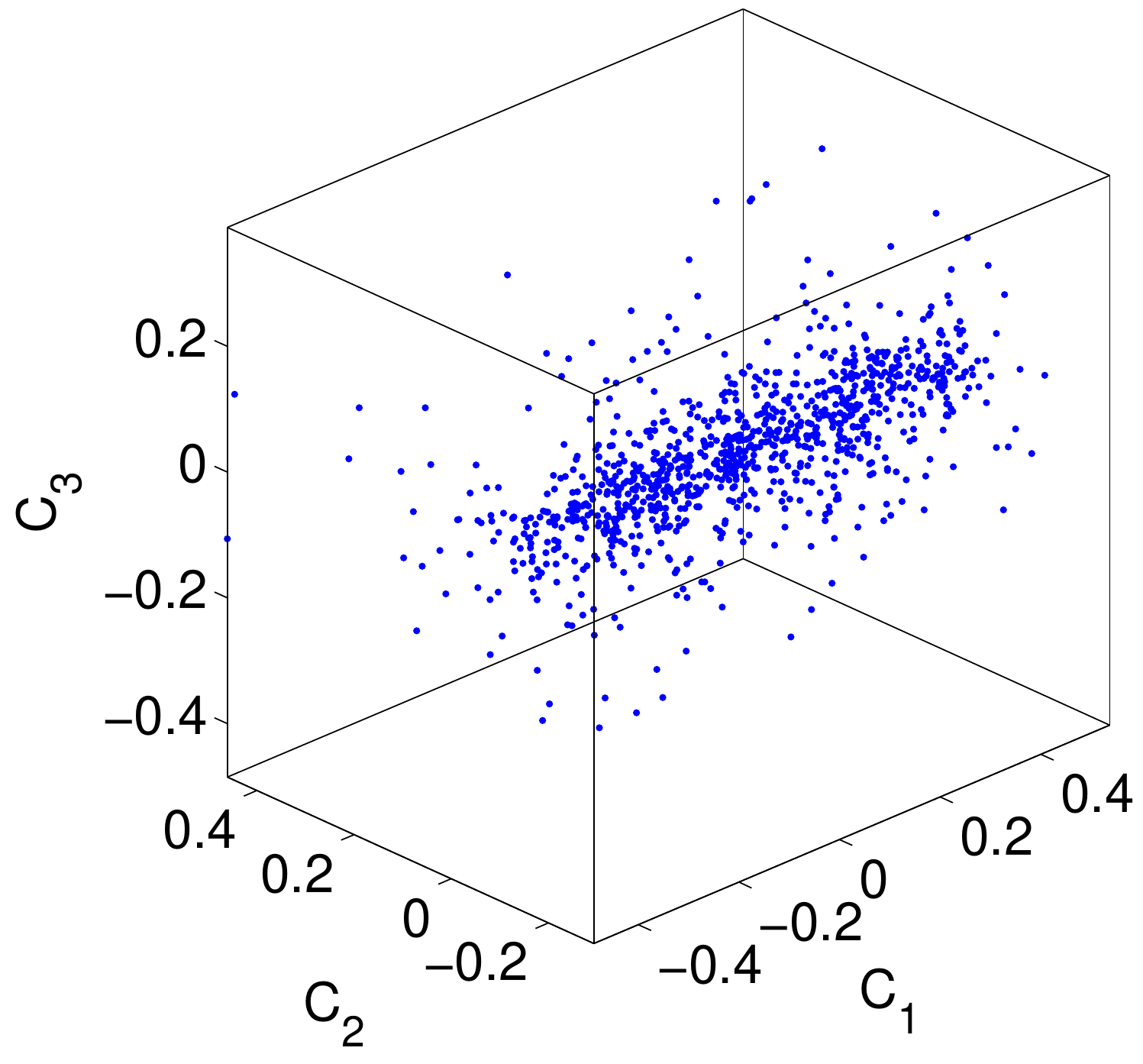} &
\includegraphics[width=3.4cm,height=3.4cm]{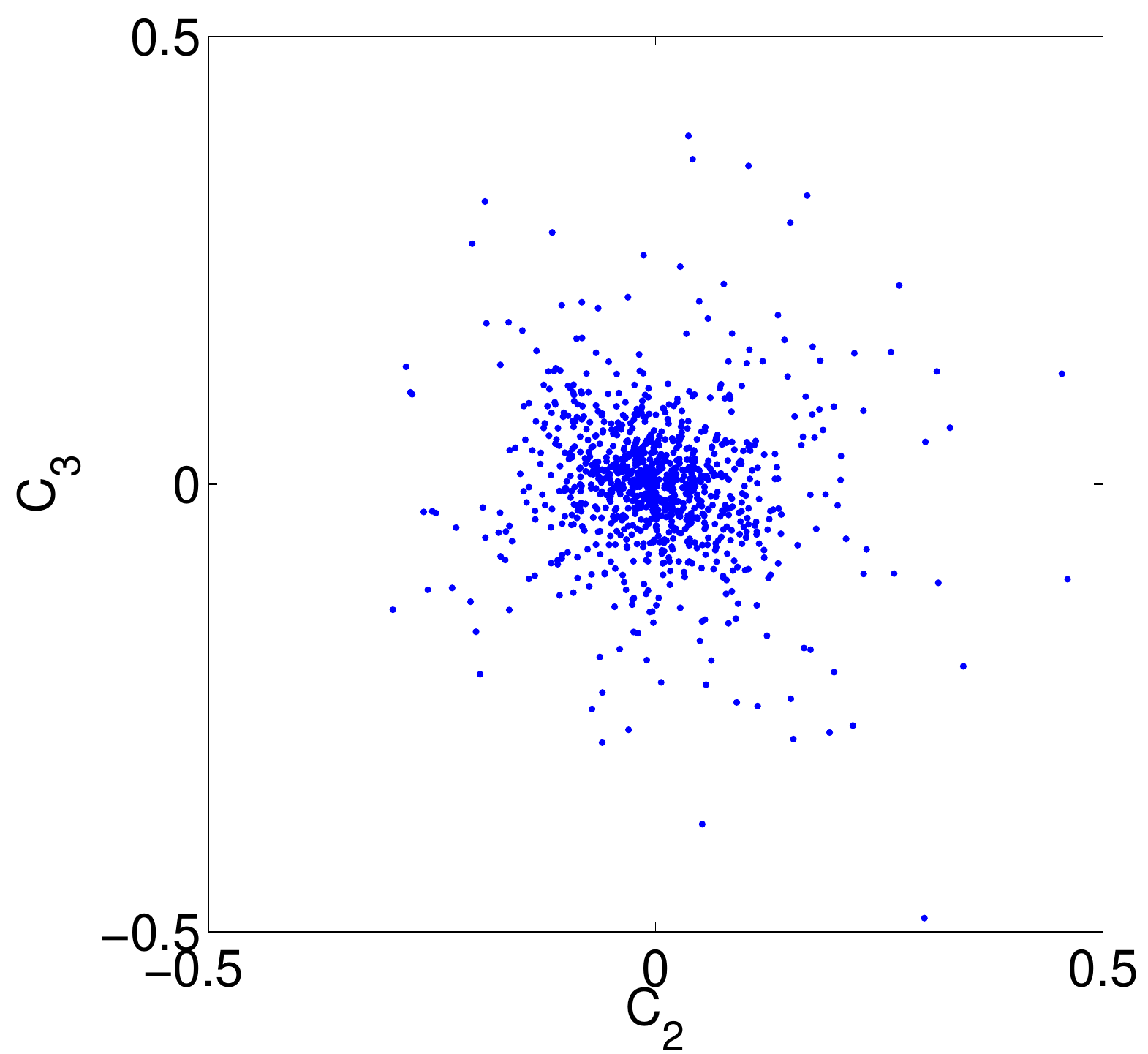} & \\
\includegraphics[width=4.5cm,height=3.4cm]{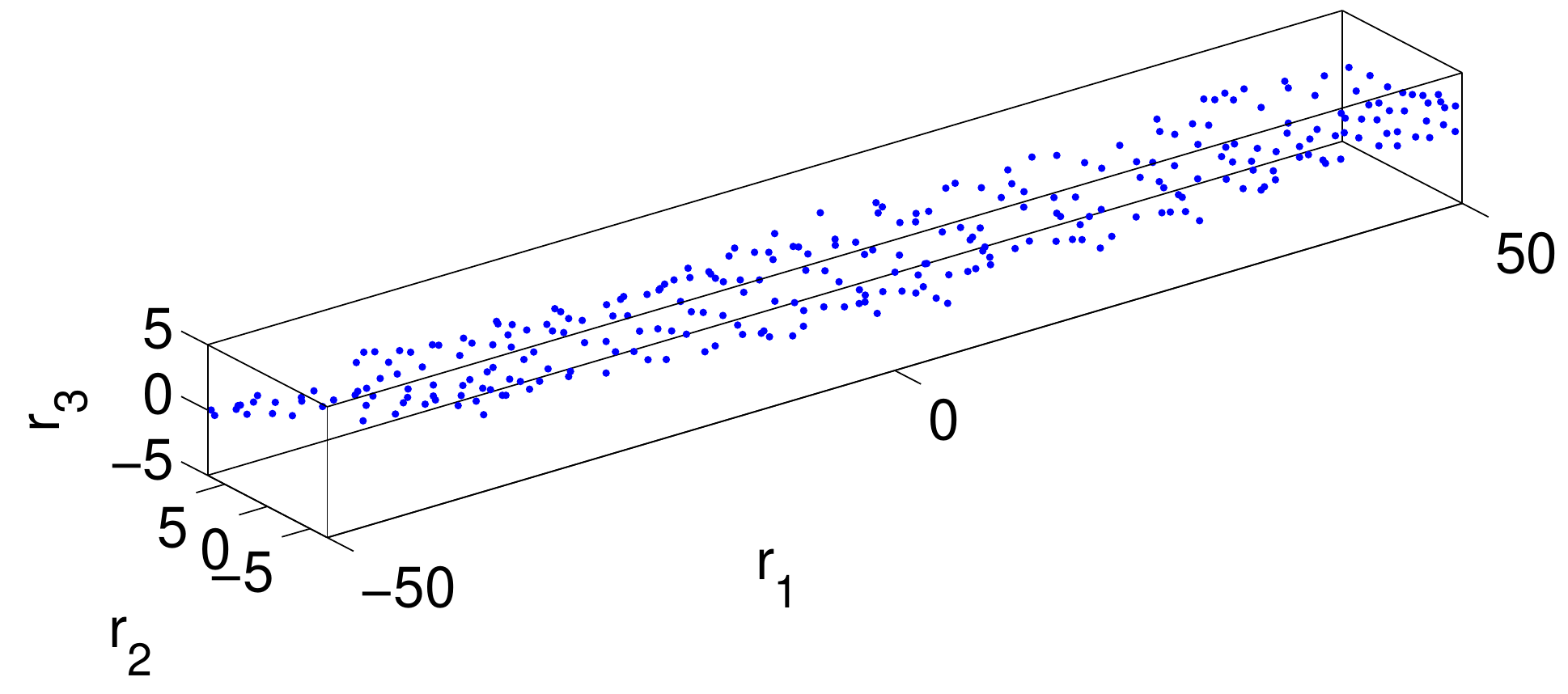} &
\includegraphics[width=4.5cm,height=3.4cm]{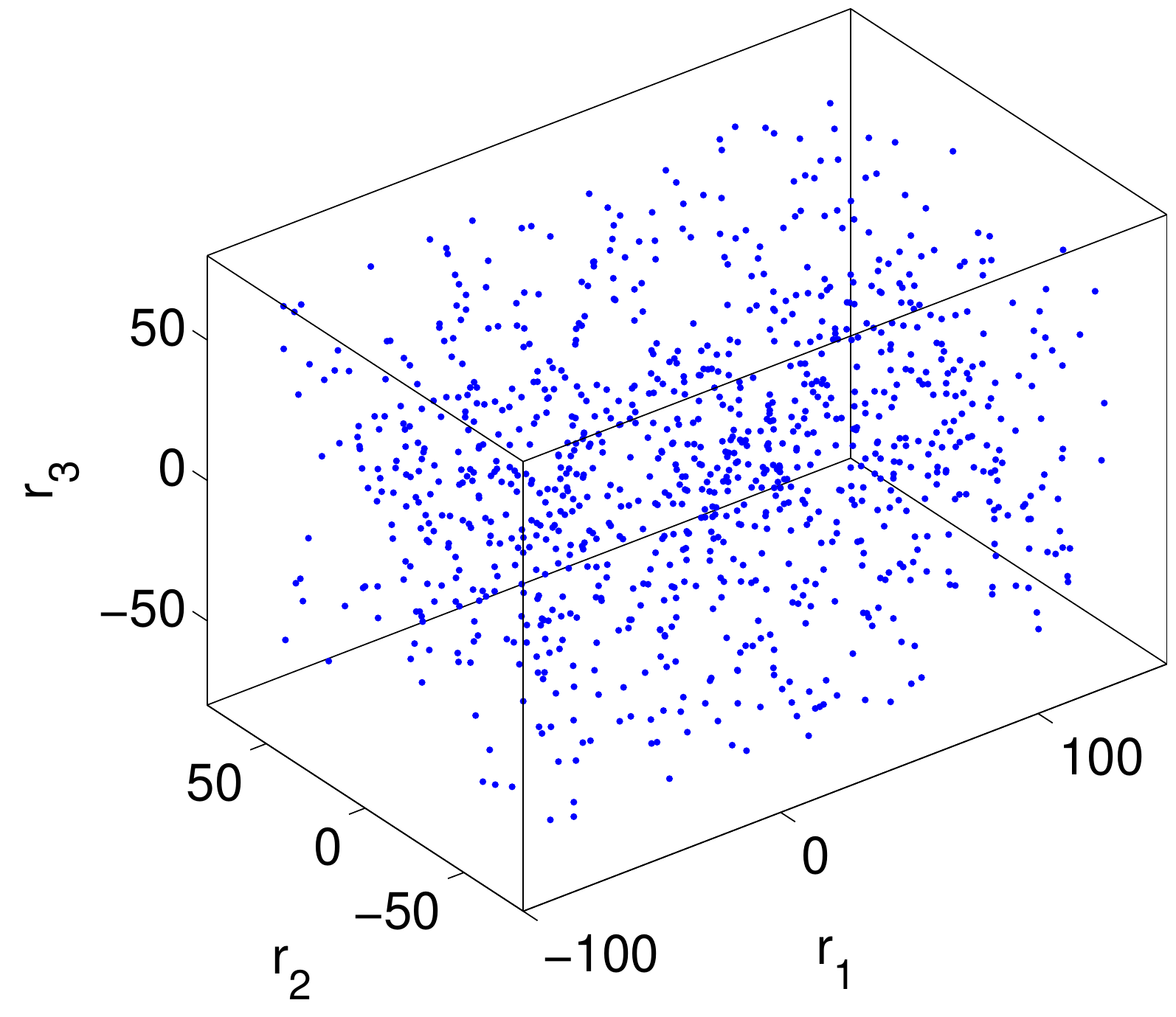} & \includegraphics[width=3.4cm,height=3.4cm]{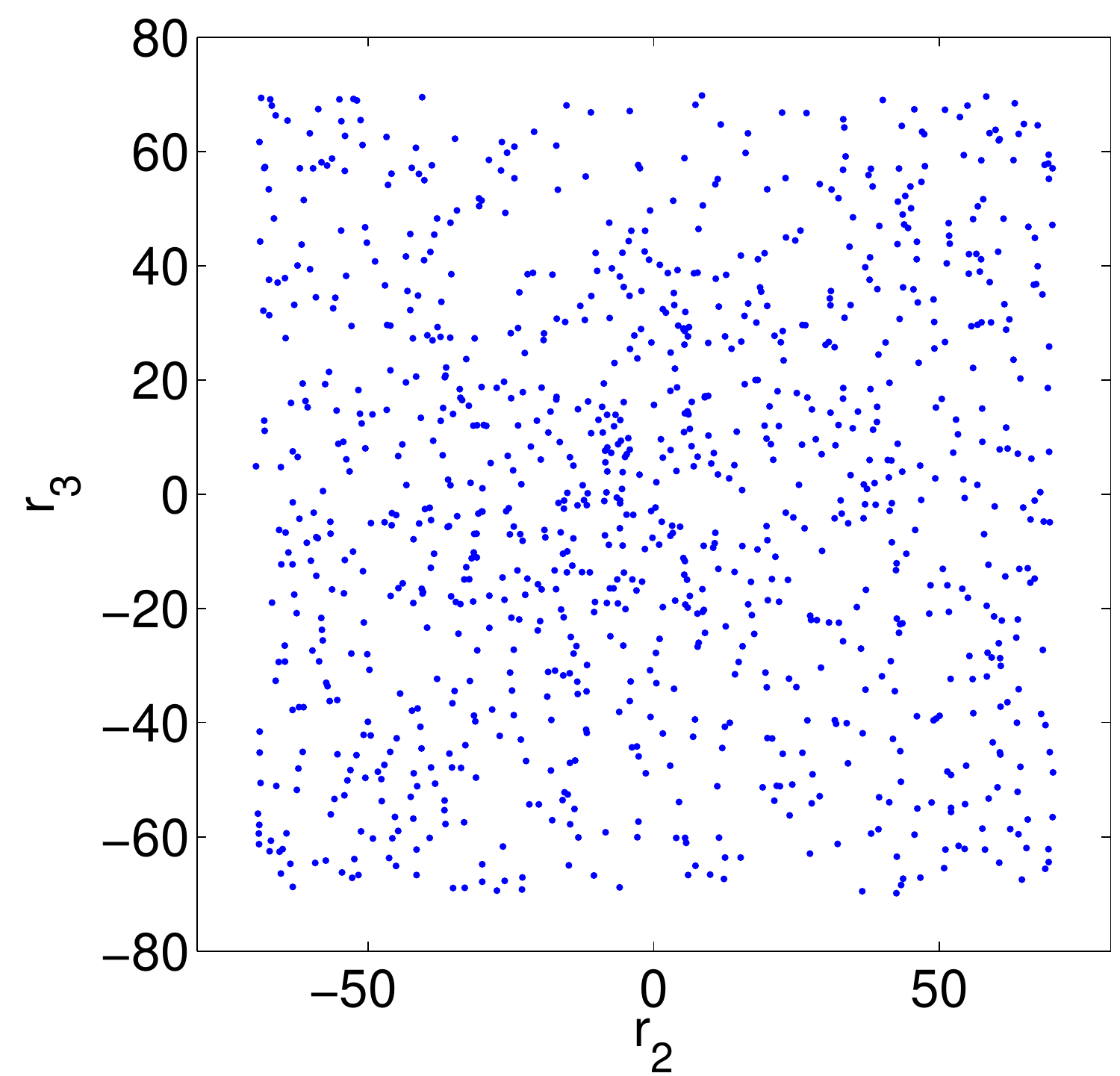} & \\
 & & & \multirow{2}{*}[6cm]{\includegraphics[width=5.0cm,height=4.5cm]{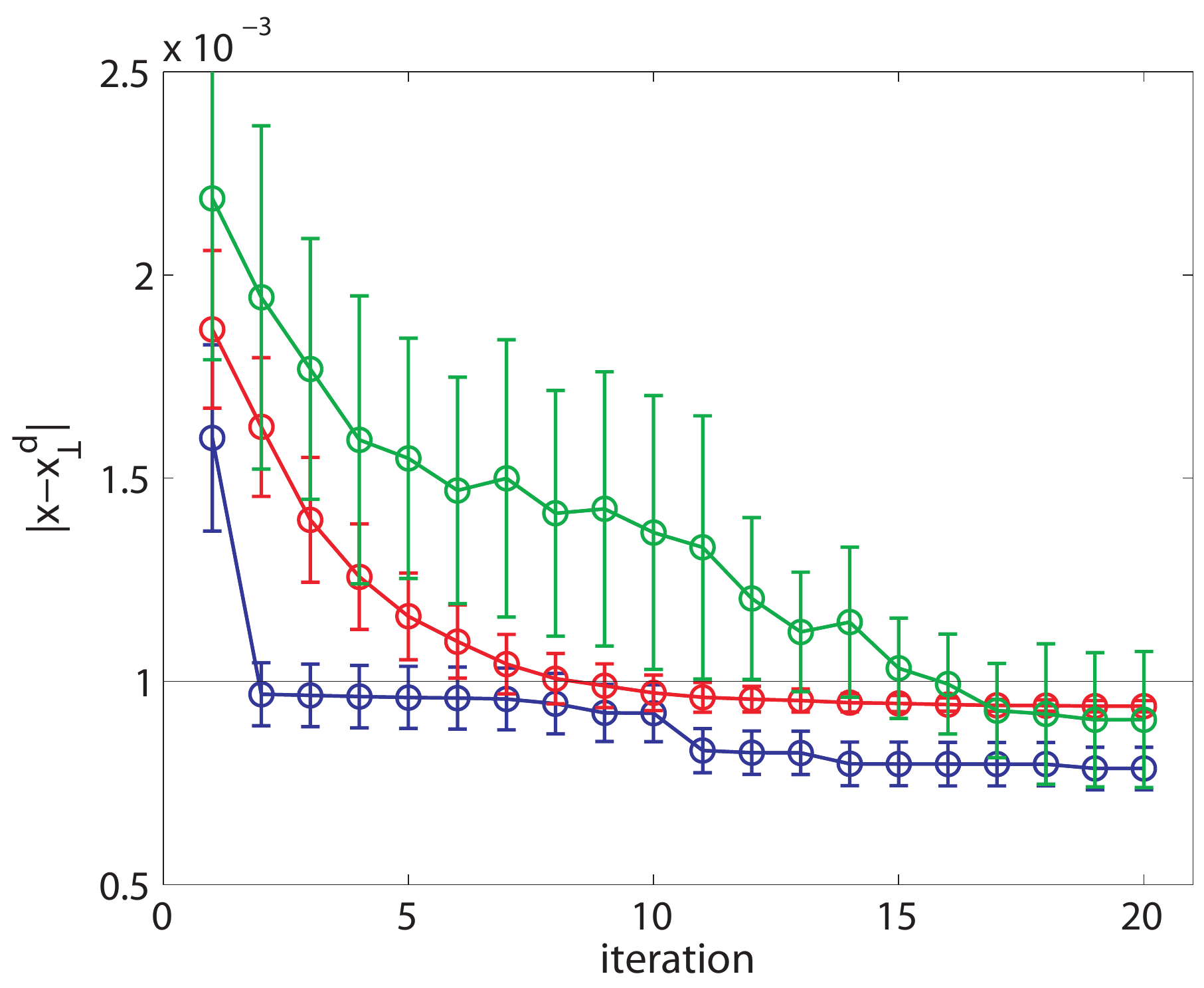}}\\[-0.0cm]
\end{tabular}
\end{center}
\caption{ Convergence of the transform (accuracy of the proposed path through the geodesic projections). Top, transformed test points from the synthetic 2d manifold (left) and the real 3d color set (right).
Euclidean metric $\gamma=0$ was used in the synthetic case and \emph{infomax} metric $\gamma=1$ was used in the color case. Points highlighted in red are those that require iterative search since the residual is bigger than some accuracy threshold (in these examples 0.001 times the maximum Euclidean distance). Bottom, residuals as a function of the iterations in the computation of the path.  Convergence saturates since iterations stopped once the accuracy threshold was reached.
}
\label{convergencia1}
\end{figure*}

\textbf{Convergence.}
The ability to reach any point in the manifold using the proposed path through the geodesic
projections (Eq. \ref{iterative} and Fig. \ref{diagrama4}) is the key to obtain meaningful transforms
with accurate inverse.
Here we check the convergence of such procedure in three cases: (1) the synthetic manifold
considered above, (2) tristimulus color data from a calibrated color image database \cite{Laparra12},
and (3) spatial texture -luminance only- data from another calibrated color image database \cite{Olmos04}.

Figure \ref{convergencia1} -top row in left and central panels- shows the shape of the manifolds coming
from real data. The color manifold in LMS space (long, medium and short wavelength) displays
the classical correlation between tristimulus components \cite{Simoncelli01,Koenderink10,Laparra12}.
Figure shows 300 test color samples, but 20000 samples were used for training.
The spatial texture data come from $15\times15$ luminance patches. PCA rotation and whitening was applied
to these patches (225-dimensional vectors). PCA was computed using $2.5 \cdot 10^6$ samples.
Then, 15000 particular image samples were selected for training falling close to a particular 3d subspace
(only 3 active PCA components and the rest close to zero). 1000 test samples are shown at the top
of the central panel. Axes are named as $C_i$ for \emph{contrast} of each PCA component.
The image texture samples show the classical elongation along the first principal component and
the sparse elliptical symmetry in the other dimensions \cite{Lyu09c,Sinz10}.

In all cases (synthetic data, color data and texture data), test points were transformed using SPCA
using different metrics ($\gamma=0$ for the synthetic and color data, and $\gamma=1$, i.e. equalization, for
the texture data). Original and transformed (unfolded) synthetic data are omitted in
Fig. \ref{convergencia1} since they are shown in Fig. \ref{gamba}. Real data transformed are shown in the
bottom row of the left and central panels in Fig. \ref{convergencia1}.
Note that $\gamma=1$ achieves component independence as predicted by the theory.

In these examples, the iteration described in Eq.~\ref{iterative} stopped when the reconstruction error $|\mathbf{x}-\mathbf{x}_\bot^d|$ was smaller than a
fraction of the maximum Euclidean distance in the manifold\footnote{In our case we used 0.001. Note that this amounts to
0.1 in LMS units for the color case, and about 0.15 $cd/m^2$ in luminance for the image case. In both real cases these
reconstruction errors are negligible since they are under the quantization error in conventional displays.}.

In the experiments, a large fraction of test points is reached within the accuracy threshold in a single iteration.
Figure \ref{convergencia1} shows the evolution of the reconstruction error for the points that were not reached in the
first iteration.
In our implementation, the convergence constant $\alpha$ in Eq.~\ref{iterative} was set to 0.01. However, faster
convergence results and improved accuracy could be obtained with adaptive $\alpha$ as in gradient descent procedures~\cite{Numerical92}. The aim of these convergence examples is just to illustrate that the transform works in practical situations: in the proposed examples, accurate direct and inverse transforms are obtained within less than 15 iterations of the algorithm. Note that convergence saturates since the iteration for each point stopped once the accuracy threshold was reached.

\subsection{Unfolding, Nonlinear ICA and Transform Coding}

\begin{figure}[t!]
\small
\begin{center}
\hspace{-0.275cm}
\begin{tabular}{ccc}
 $\gamma = 0$ & $\gamma = 1/3$ & $\gamma =  1$\\
\hspace{-0.0cm}\vspace{0.5cm}\includegraphics[width=2.83cm]{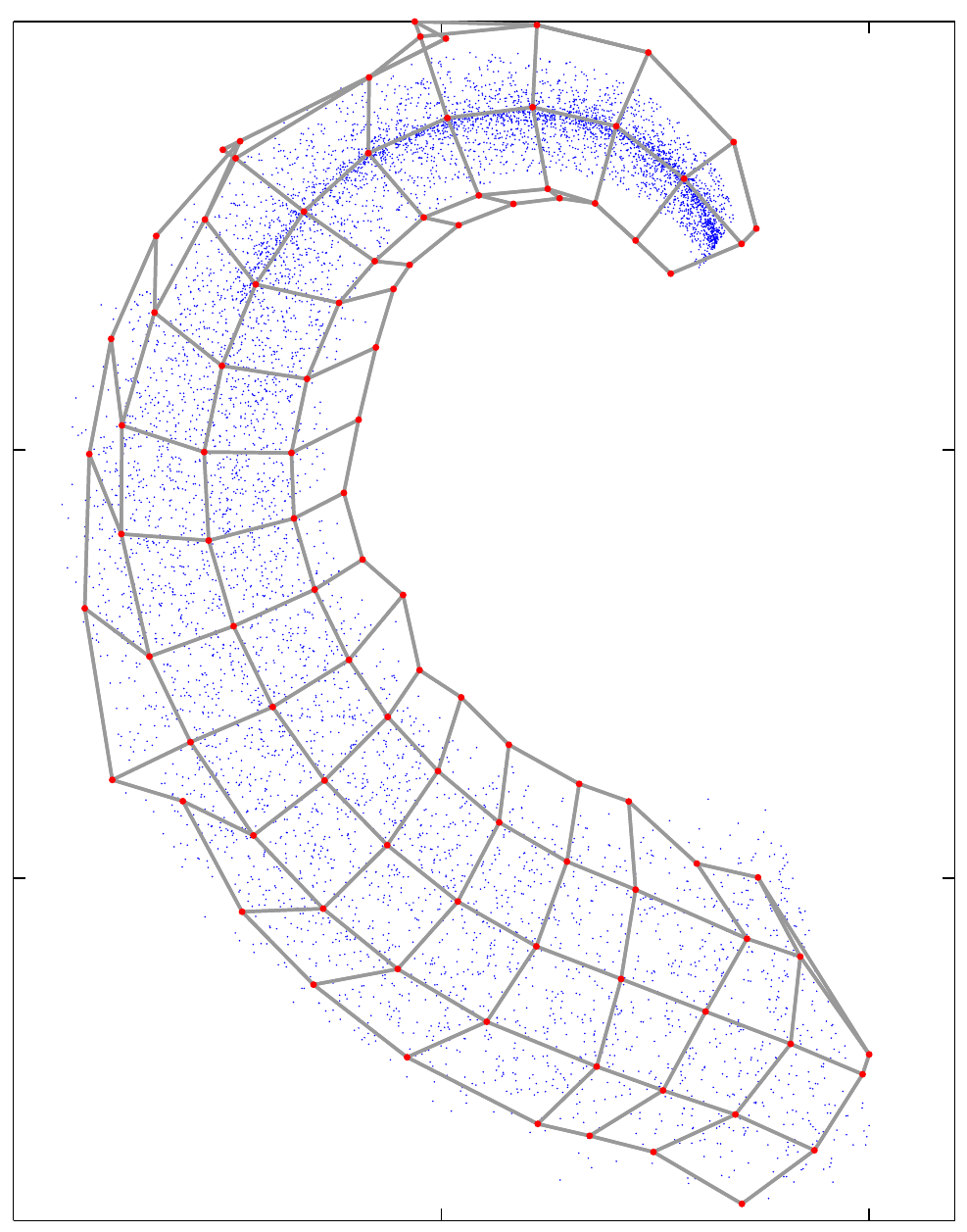} &
\hspace{-0.35cm}\includegraphics[width=2.83cm]{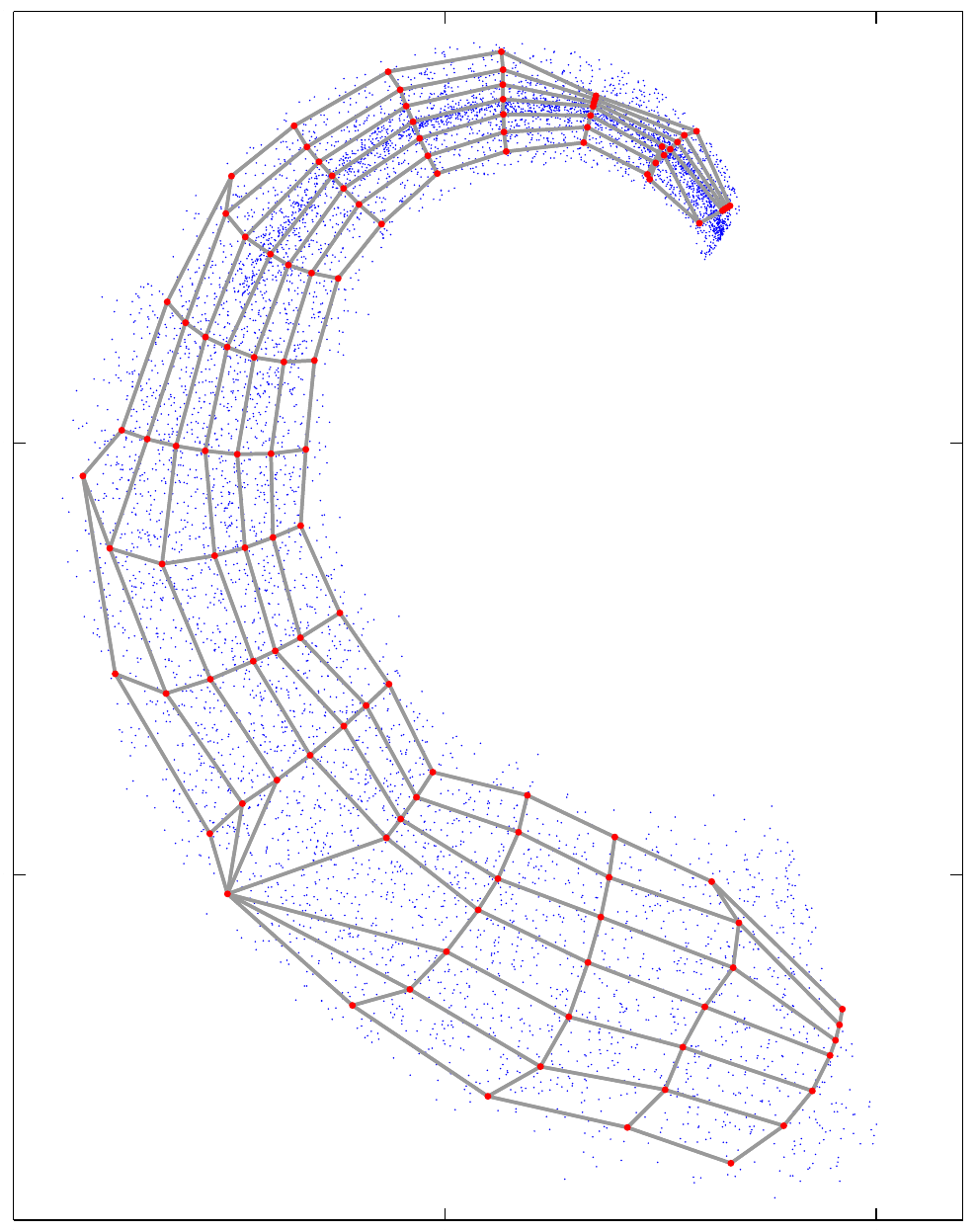} &
\hspace{-0.35cm}\includegraphics[width=2.83cm]{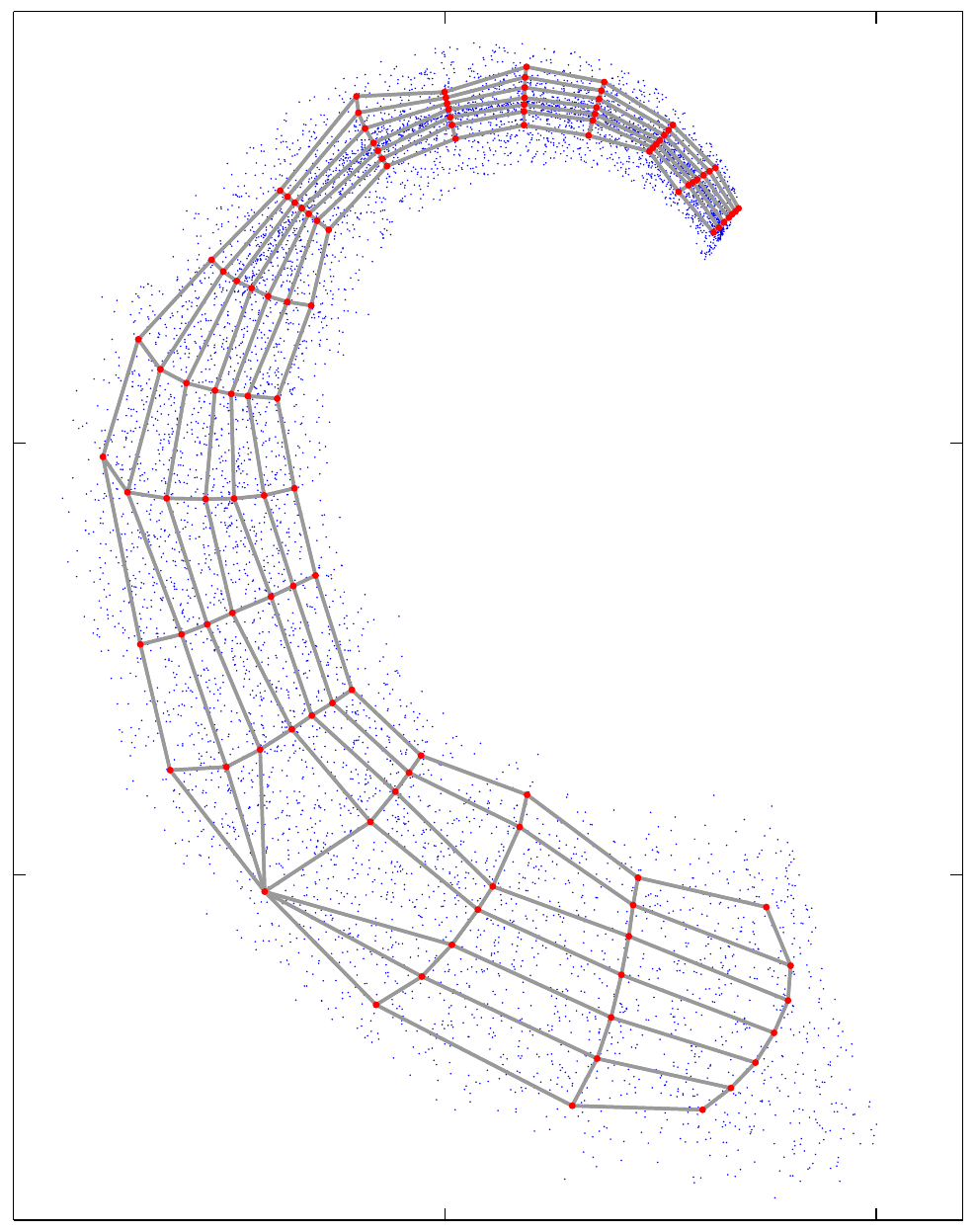} \\[-0.3cm]
 \hspace{-0.0cm} \vspace{-1.0cm}\includegraphics[width=2.83cm]{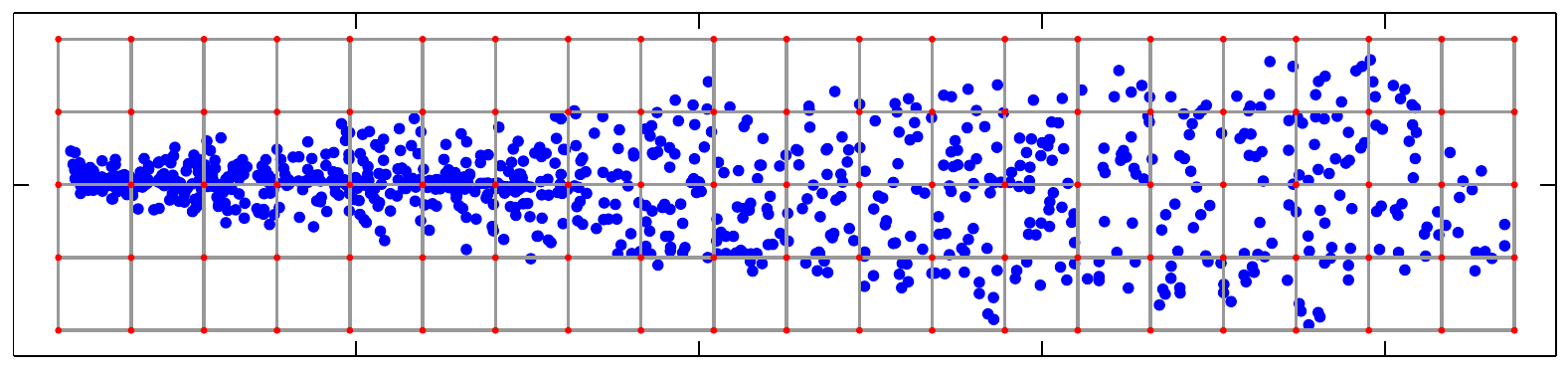} & & \\
 & \hspace{-0.35cm}\includegraphics[width=2.83cm,height=1.1cm]{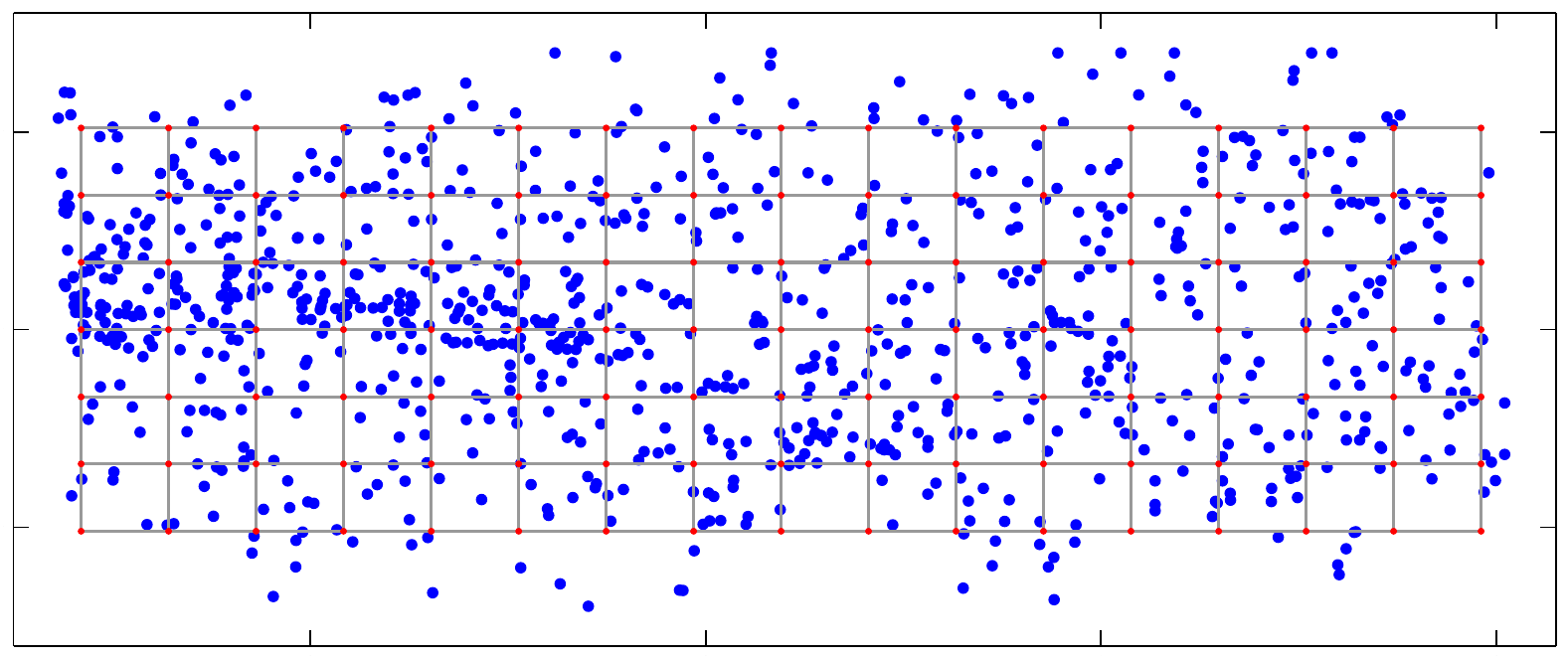} &
\hspace{-0.35cm}\includegraphics[width=2.83cm,height=1.1cm]{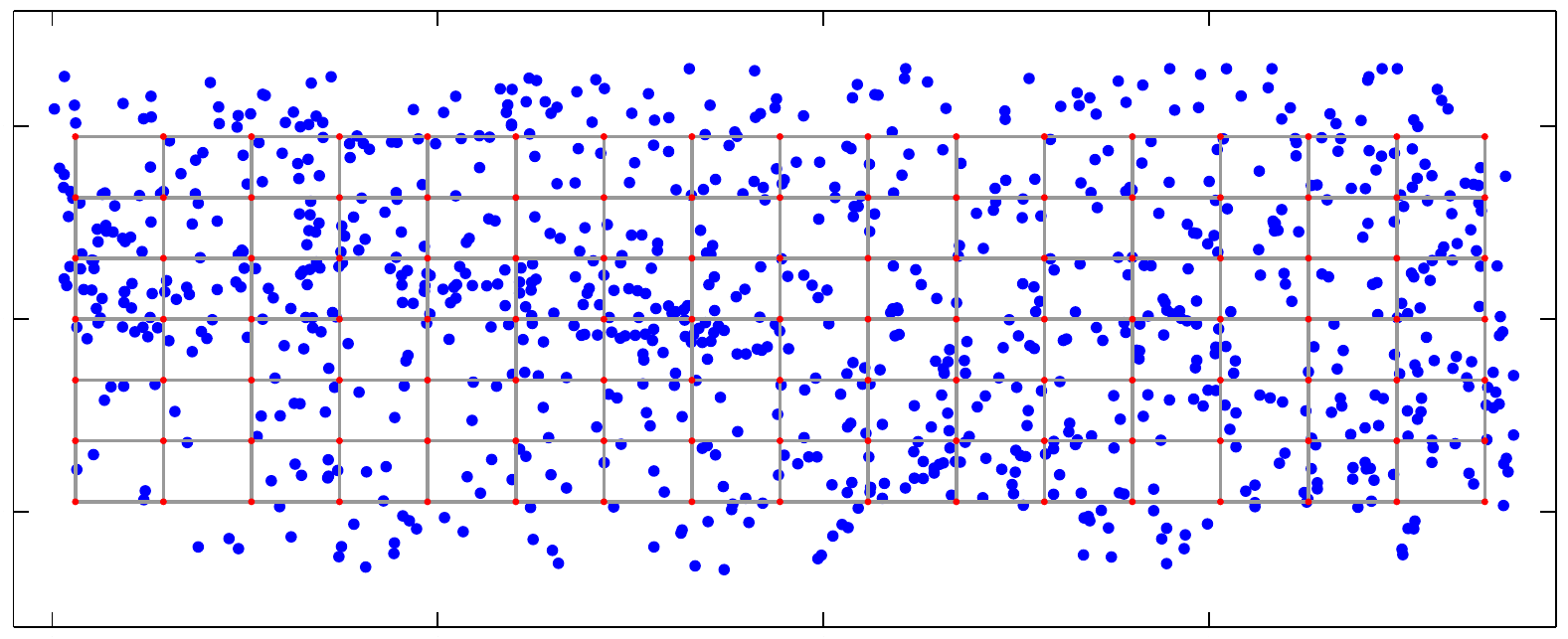} \\
\hline
$MI$ (bits)    0.27  &  0.06 & 0.05  \\
RMSE   0.55 & 0.53  & 0.56  \\
\hline
\end{tabular}
\end{center}
\vspace{-0.1cm}
\caption{ \emph{Infomax} or \emph{error minimization} through SPCA.
The sets in the first row were transformed using SPCA (second row) with different $\gamma$ values.
Additionally, Cartesian lattices in the response domain were inverted back into the input domain
leading to the curved lattices in the top row.
Results are analyzed in terms of independence (Mutual Information), and reconstruction error (RMSE).
In each case, MI was computed in the transform domain, while RMSE values refer to the quantization error
in the input domain using the corresponding lattices as codebook.
For reference, in the original domain results were $MI=0.75$ bits and RMSE=$0.63$ (using uniform scalar quantization).
Note how $\gamma = 1$ obtains better results in independence while $\gamma = \frac{1}{3}$ is better for RMSE minimization.
The Euclidean choice, $\gamma=0$, leads to constant-size cells independent of the PDF.}
\label{gamba}
\end{figure}

\textbf{Synthetic example.} The advantage of using PCs to design a set of sensors is that their flexibility makes them suitable to describe curved manifolds, as pointed out in Fig.~\ref{gamba}. No matter the metric used, an unwrapped representation of the data is obtained. When using $\gamma = 0$ the data is unfolded and the original local metric is preserved (e.g. the different distributions inside the manifold remain the same). When using $\gamma = 1$, we obtain a representation where the different distributions are almost uniformized, leading to a representation where the different dimensions are almost independent. Finally, when using  $\gamma = \frac{1}{3}$, the reconstruction error is minimized. In this latter case, redundancy is certainly reduced with regard to the input domain, however, the kurtotic structure of the Laplacian is more visible than in the second case. Note also the differences in the distribution of the inverted lattices: while in the $\gamma = 0$ case, lattice cells are approximately uniform no matter the local population (local metric  independent of the PDF), in the other cases, the size is related to the population, e.g. the $\gamma = 1$ case results in tighter slices around the peak of the Laplacian distribution. As anticipated in Section~\ref{motivation}, unfolding alone ($\gamma=0$) is not enough to remove redundancies in general, but incorporating local changes in the metric can achieve independent components.
\vspace{0.15cm}

\begin{figure}[b!]
\begin{center}
\small
\setlength{\tabcolsep}{2pt}
\begin{tabular}{cc}
\multicolumn{2}{c}{
\hspace{-0.0cm}\includegraphics[width=6.0cm,height=5.6cm]{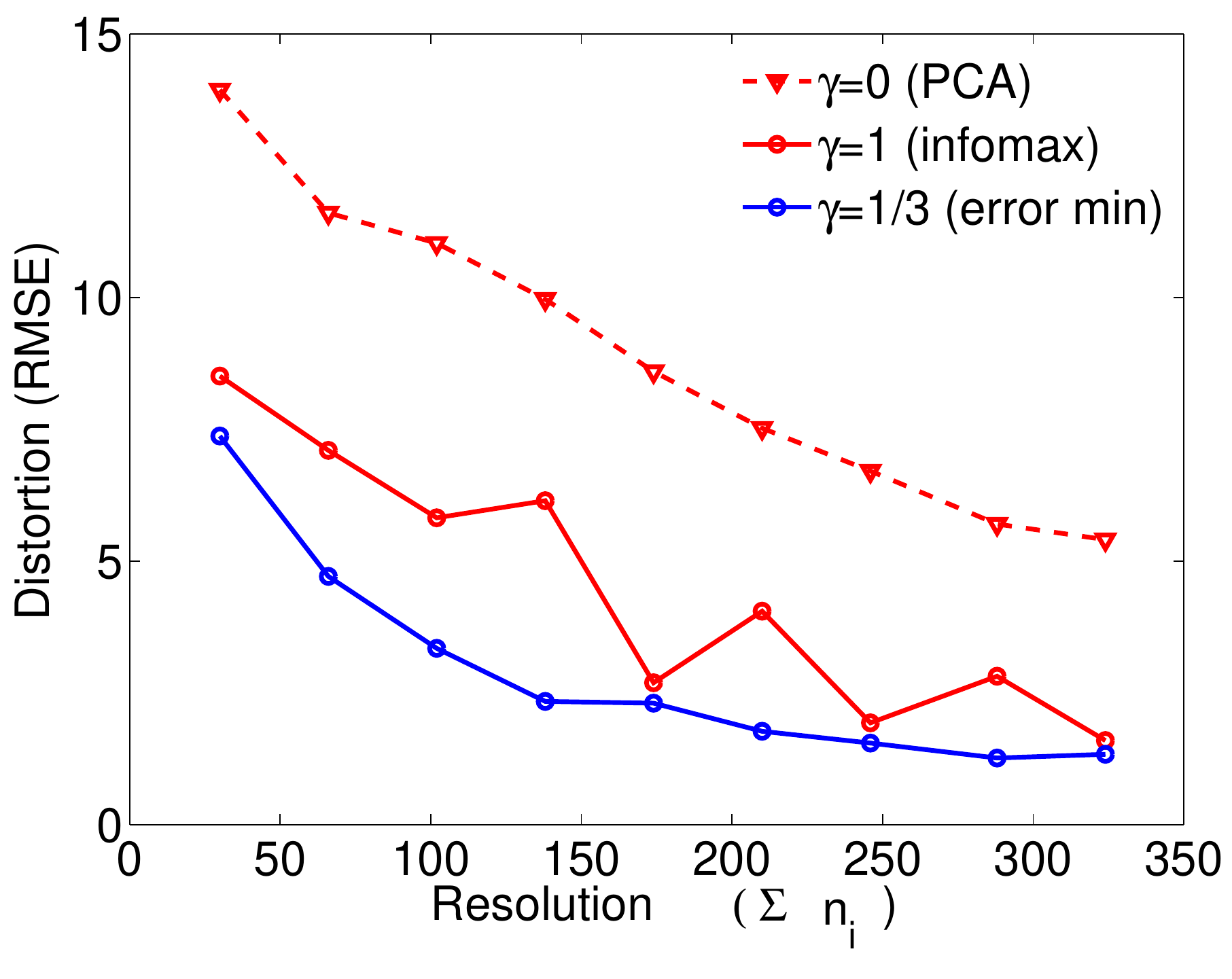}
} \vspace{-0.2cm}\\
\includegraphics[width=4.4cm,height=4.4cm]{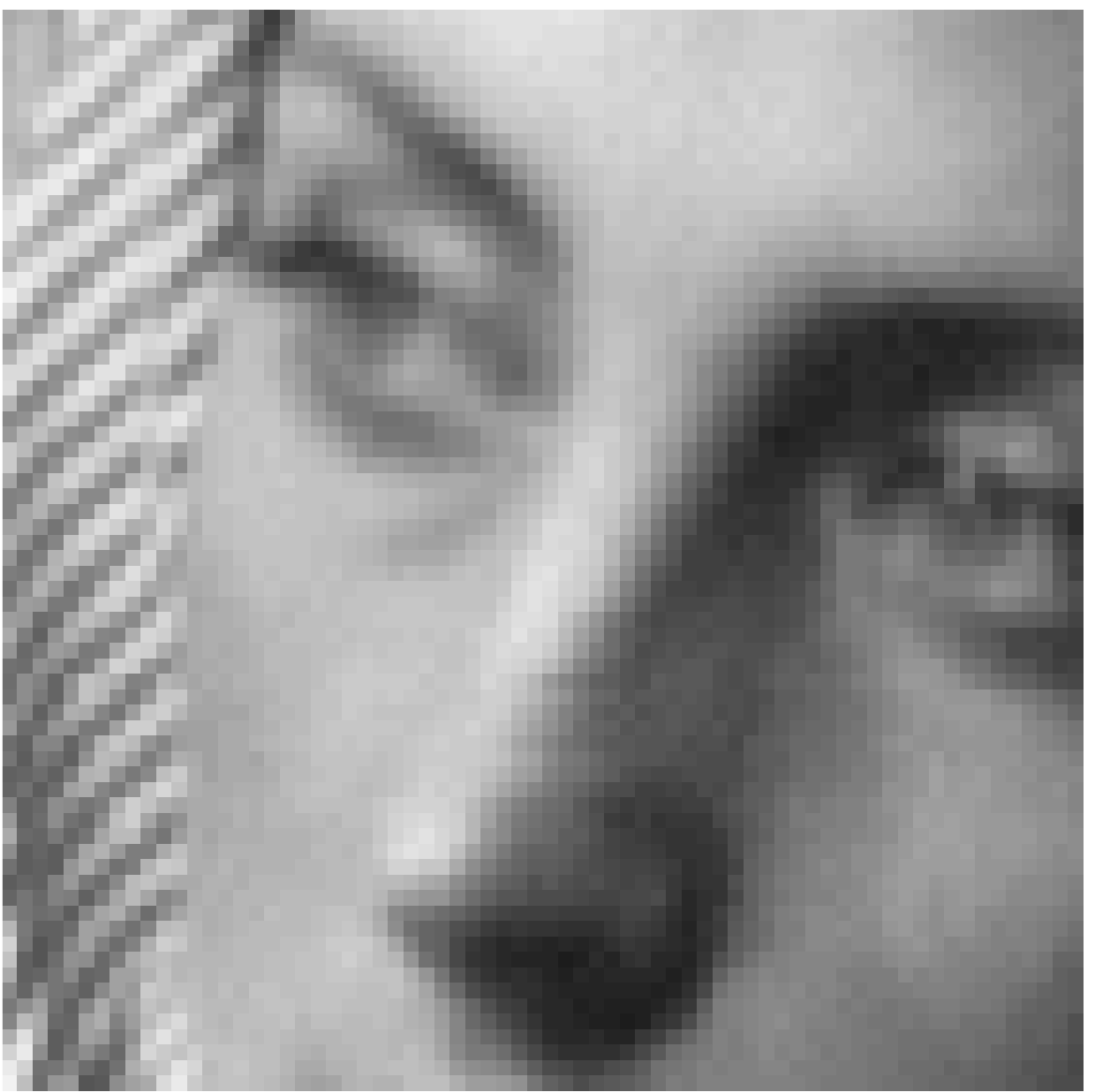} &
\includegraphics[width=4.4cm,height=4.4cm]{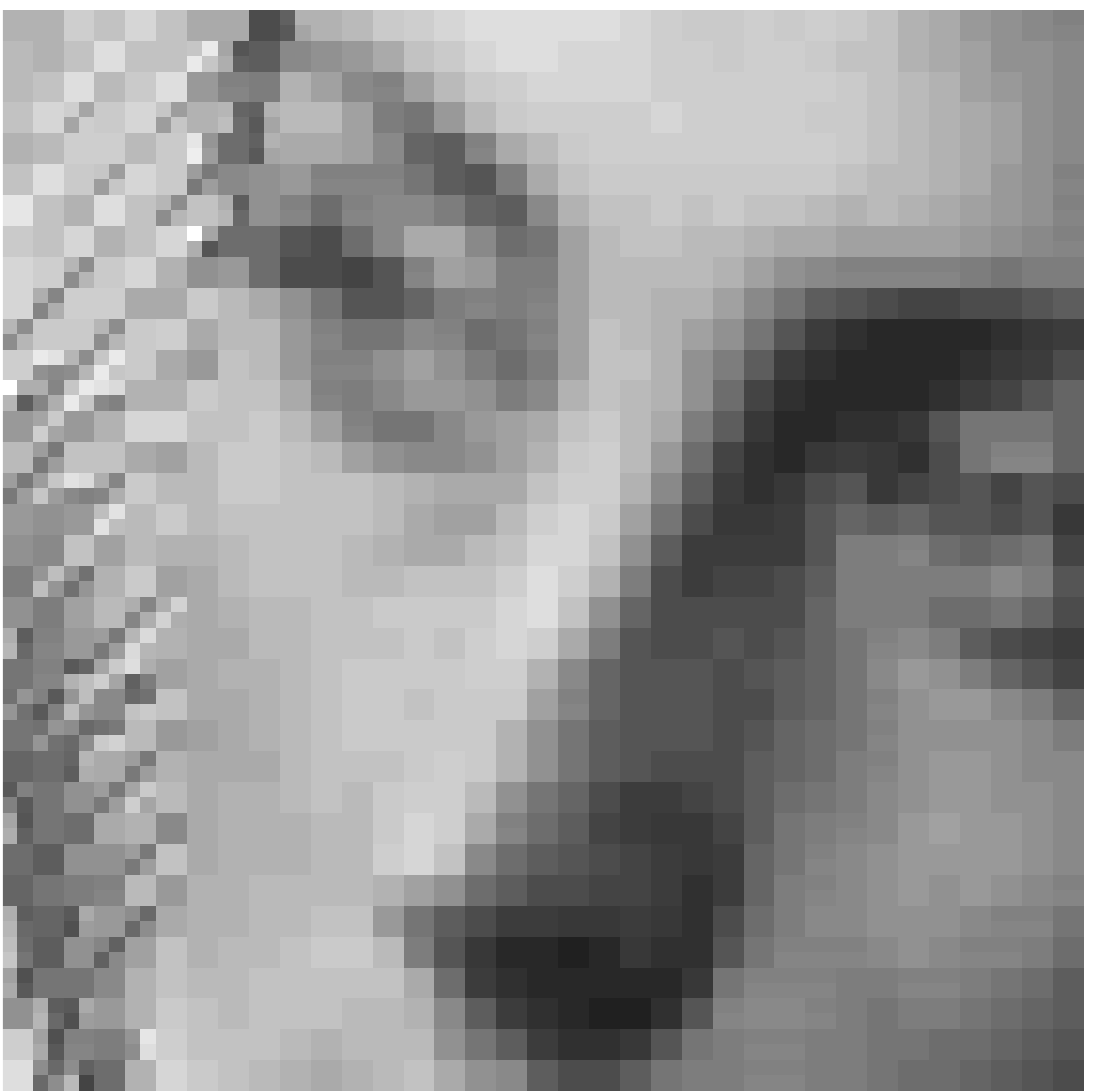} \\
   \small Original & \small PCA \\
\includegraphics[width=4.4cm,height=4.4cm]{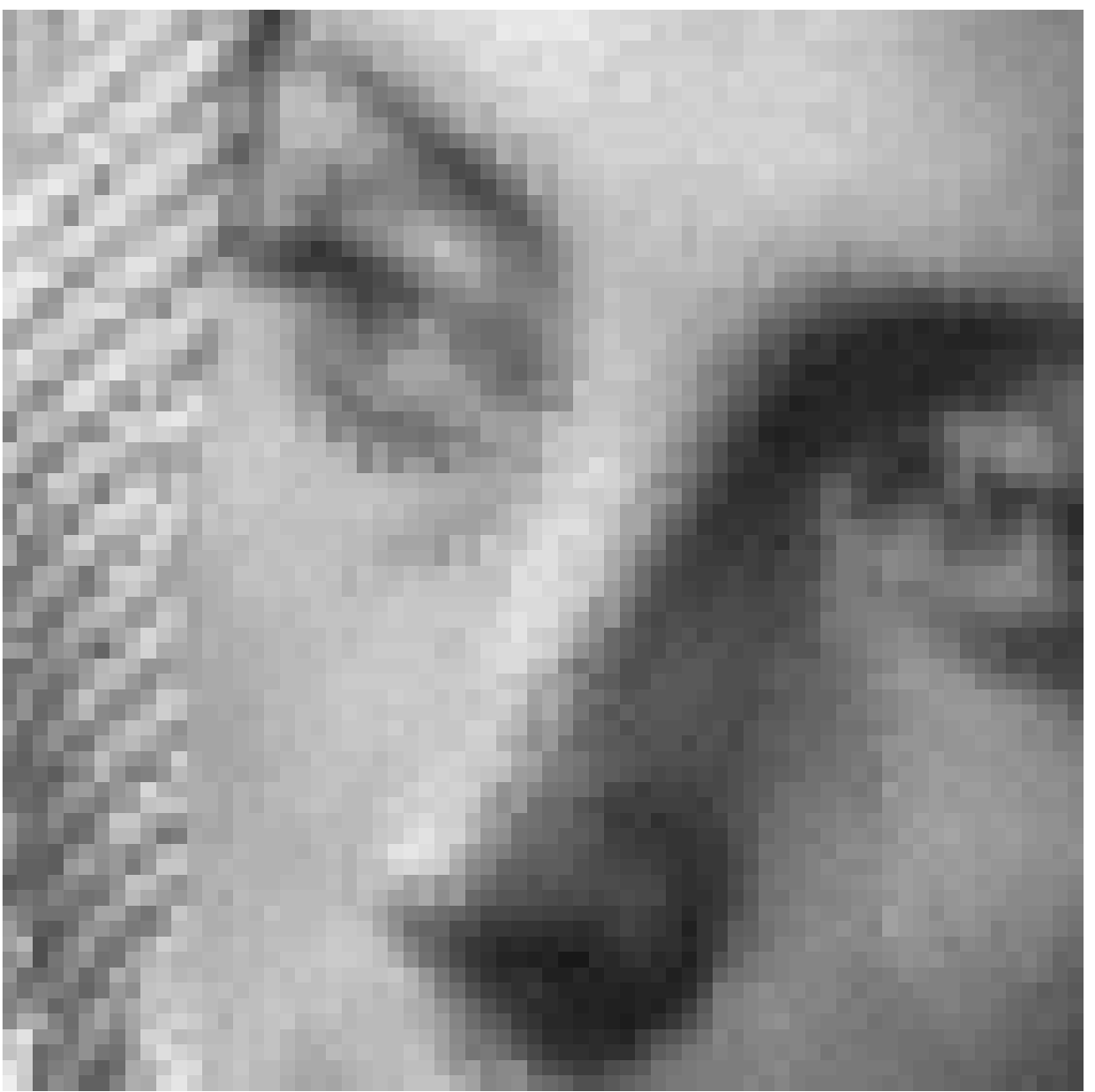} &
\includegraphics[width=4.4cm,height=4.4cm]{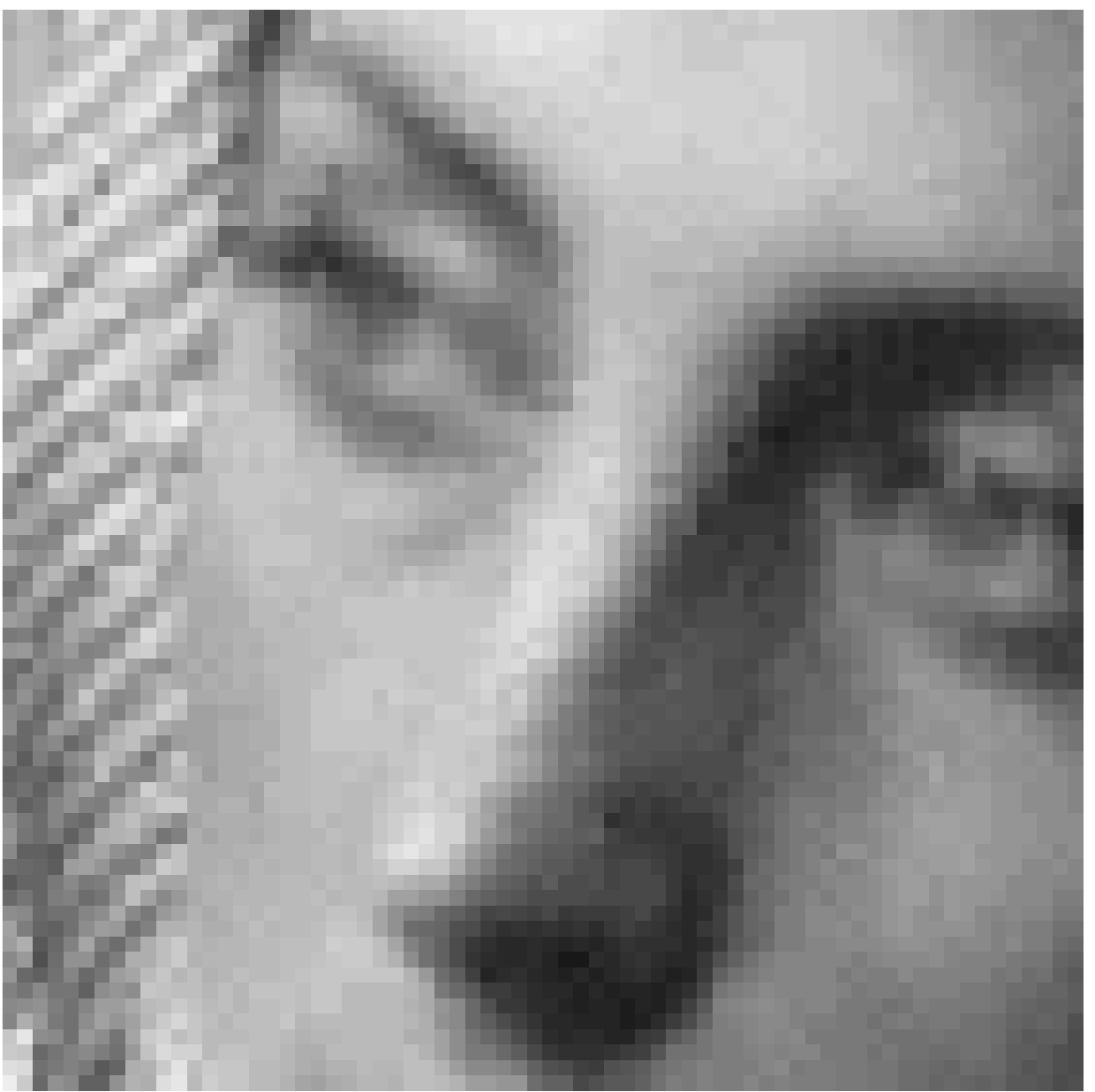} \\
   \small SPCA ($\gamma = 1$) & \small SPCA ($\gamma = 1/3$) \\
\end{tabular}
\end{center}
\vspace{-0.25cm}
\caption{Image coding results. Top: Resolution-Distortion plot.
Bottom: reconstructed images using the same resolution ($\sum_i n_i = 66$) in different representations.
}
\label{decoded}
\end{figure}

\begin{figure}[b!]
\begin{center}
\hspace{-1.3cm}
\setlength{\tabcolsep}{1pt}
\begin{tabular}{cc}
\includegraphics[width=4.1cm]{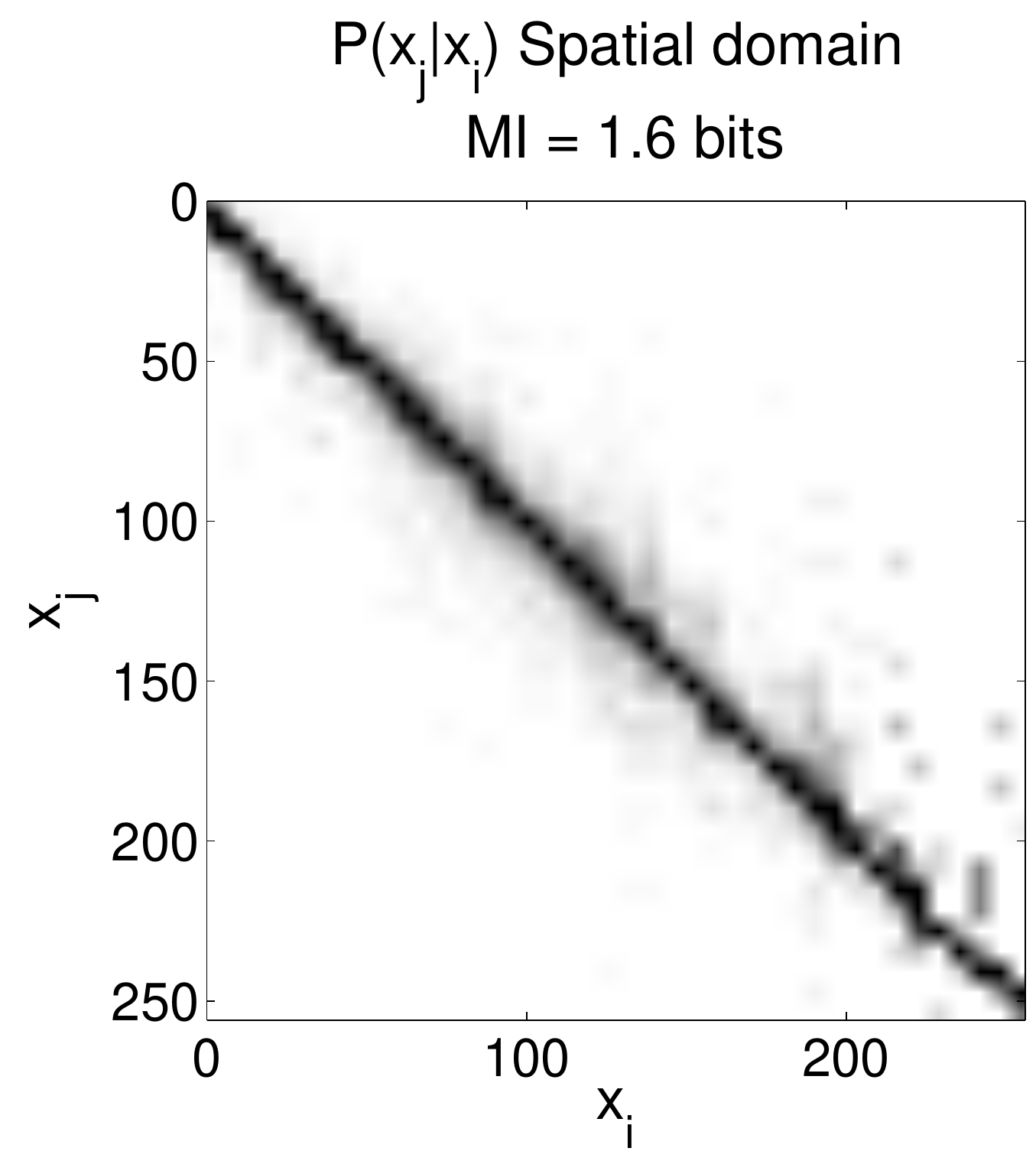} &
\includegraphics[width=4.1cm]{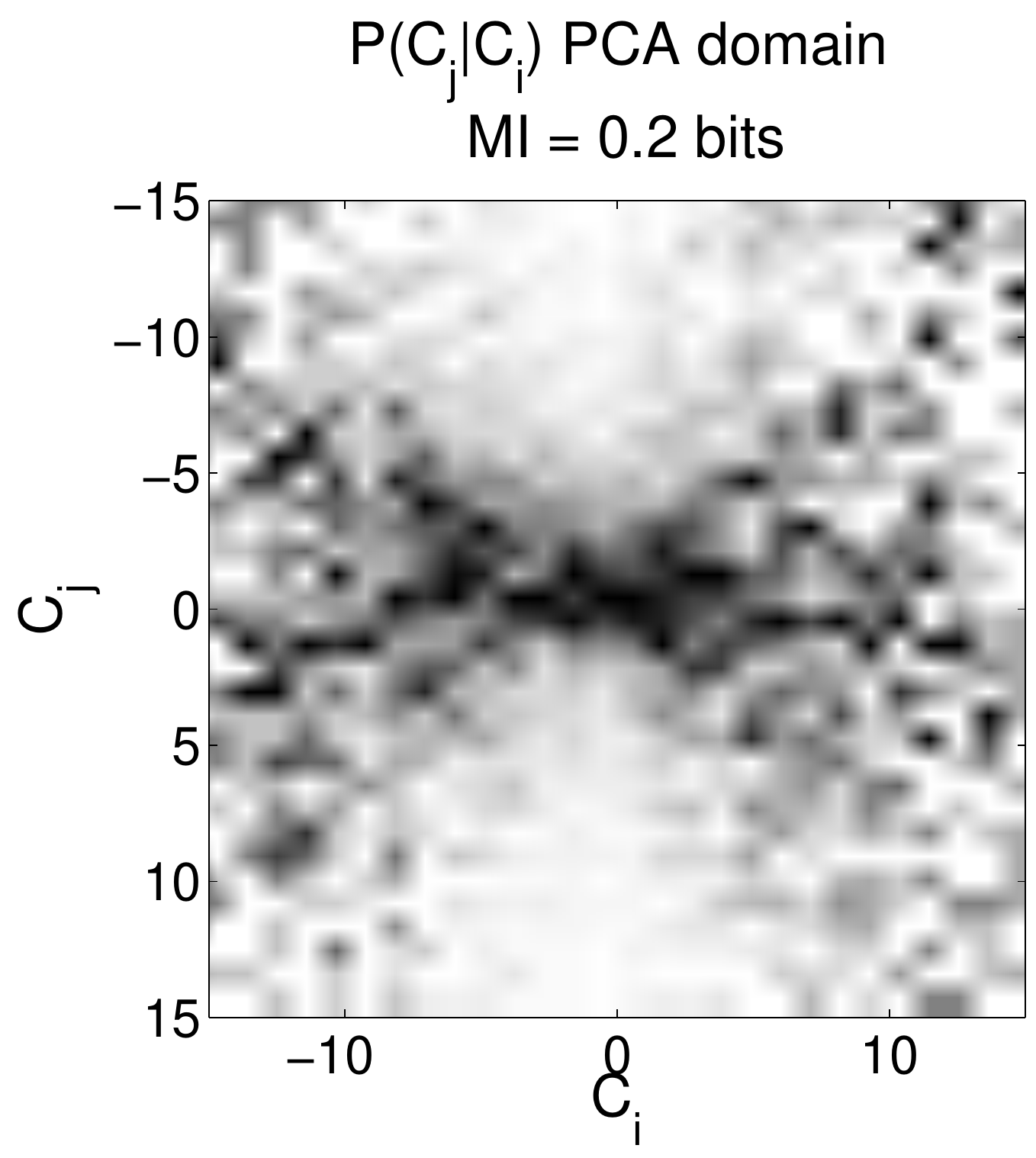} \\
\includegraphics[width=4.1cm]{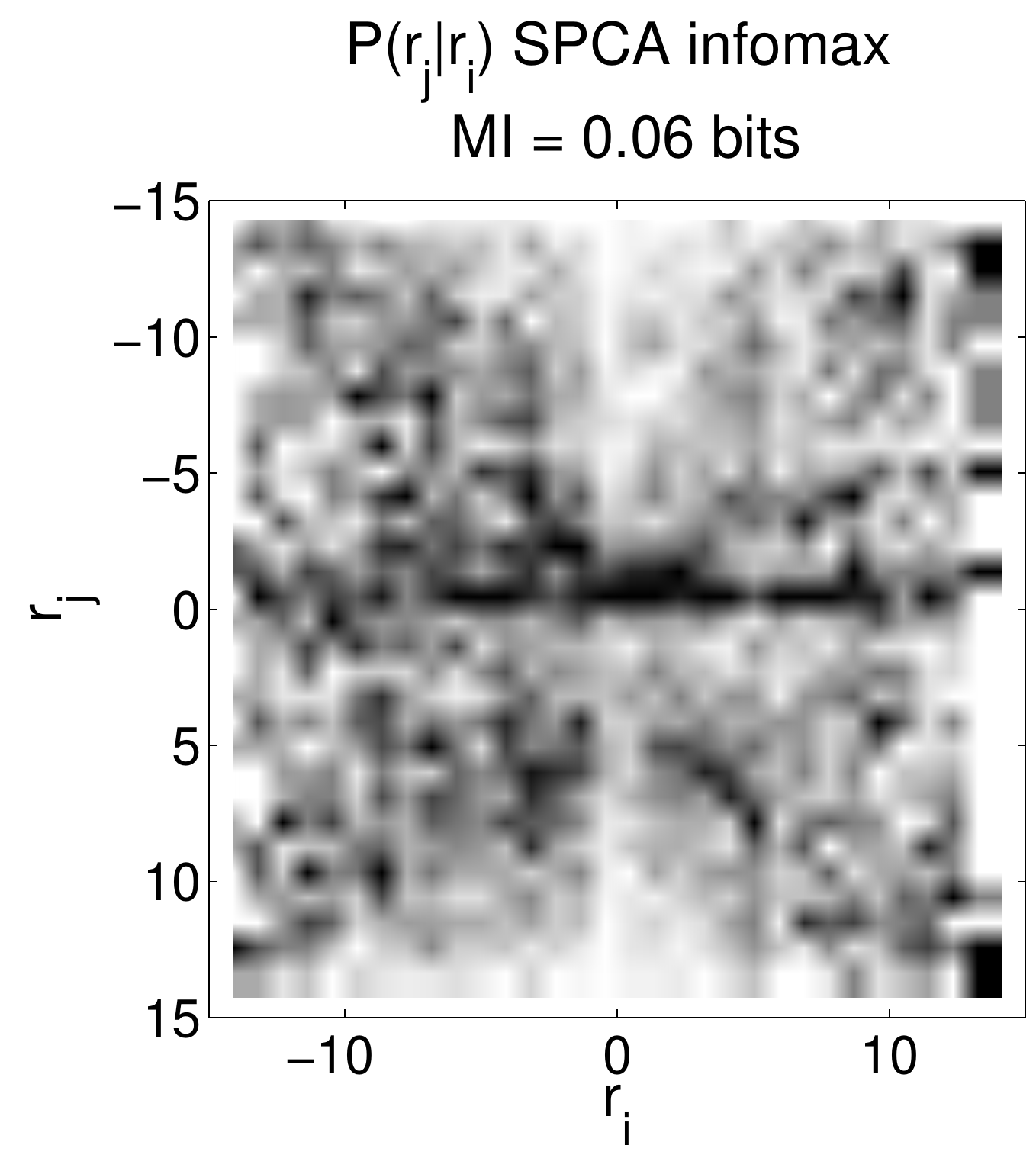} &
\includegraphics[width=4.1cm]{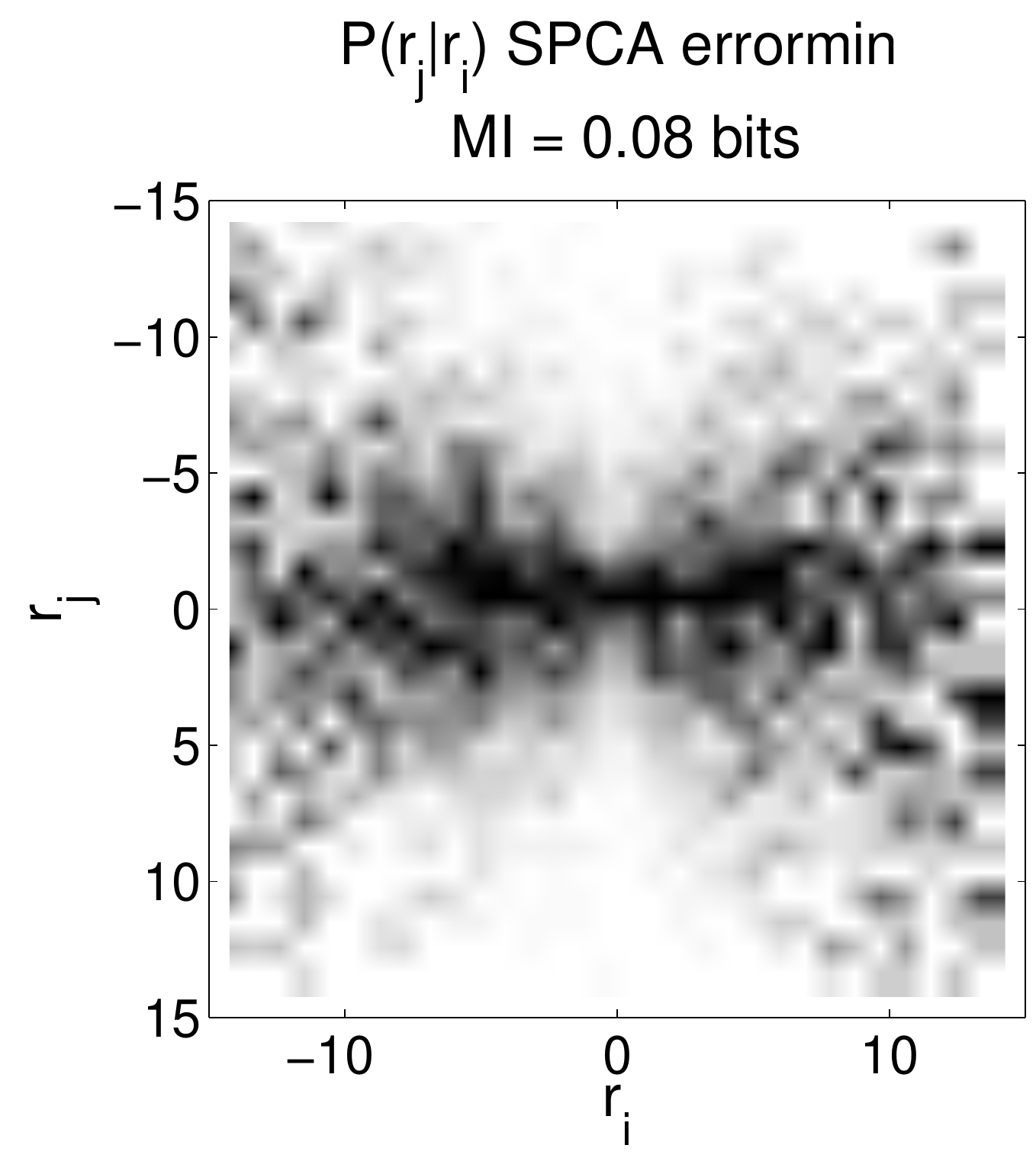} \\
\end{tabular}
\end{center}
\vspace{-0.25cm}
\caption{Conditional PDFs and quantitative independence measures in the considered domains. Top: spatial domain (left) and PCA domain (right). Bottom:
SPCA $\gamma=1$ (left) and SPCA $\gamma=1/3$ (right).
Note that bigger Mutual Information values are consistent with visible (correlation or bow-tie) structures in the conditional probabilities.
}
\label{pajaritas}
\end{figure}

\textbf{Encoding spatial texture.} Efficient representation of spatial information is a challenging problem for
manifold learning techniques and a suitable scenario to check the effect of different optimality
criteria. In this section the training set consists of $2\cdot10^5$ 4-dimensional vectors from $2\times2$ luminance patches of natural
images from the calibrated McGill database \cite{Olmos04}.
SPCA was trained on these samples in the PCA domain using $\gamma=1$ and $\gamma=1/3$ metrics.
Note that given the symmetry of image data\footnote{Image texture manifolds are not curved as shown in Fig. \ref{convergencia1}}, plain unfolding, i.e. $\gamma=0$, is very similar to PCA.
SPCA was then applied to 1500 test samples from the McGill database and to 2500 samples from the standard image Barbara
(not included in the training set).

The ability of SPCA for transform coding is illustrated by using sensors with limited resolution (uniform quantization in each dimension of the transformed domain) in the considered representations. In every case, resolution in each dimension was set according to standard
bit allocation \cite{Gersho92}. Figure \ref{decoded} shows the reconstruction error as a function of the resolution of the sensors
for 60 randomly chosen samples from the Barbara image. Figure \ref{decoded} also shows reconstructed images from the quantized representations
using the same sensor resolution (number of quantization bins). The resolution-distortion plot shows that SPCA with the \emph{error minimization} metric substantially reduces the RMSE in image coding with regard to Euclidean metric ($\gamma=0$, or uniform quantization of PCA) and to the \emph{infomax} SPCA. The decoded images show the practical relevance of the numerical gain achieved by the non-Euclidean $\gamma=1/3$ approach.

The ability of SPCA for nonlinear ICA is qualitatively and quantitatively assessed by inspecting the conditional PDFs
between AC coefficients (as in \cite{Bucigrossi99,Hyvarinen03,Malo10}) and by the corresponding mutual information (MI)
measures (see Fig. \ref{pajaritas}).
Bow-tie structures in the conditional PDFs and MI measures in the spatial domain and in the PCA domain are consistent with previously
reported results for natural images \cite{Malo10}. Uniform conditional PDF and small MI show that SPCA with the \emph{infomax} metric
strongly reduces the redundancy between the coefficients of the representation. On the contrary, in the case of SPCA with the \emph{error minimization} metric 
the bow-tie shape is still visible in the conditional PDF.

These results confirm the theoretical prediction that SPCA can be tuned either for \emph{infomax} (using $\gamma=1$) or for \emph{error minimization} (using $\gamma=1/3$).

\subsection{Dimensionality reduction}

\begin{figure*}[t!]
\begin{center}
\begin{tabular}{cccccc}
\small SPCA $\gamma$ = 0   & \small SPCA $\gamma$ = 1/3  & \small SPCA $\gamma = 1$  & \small LLE  & \small Isomap  & \small Charting\\
\includegraphics[height=3.7cm,width=2.7cm]{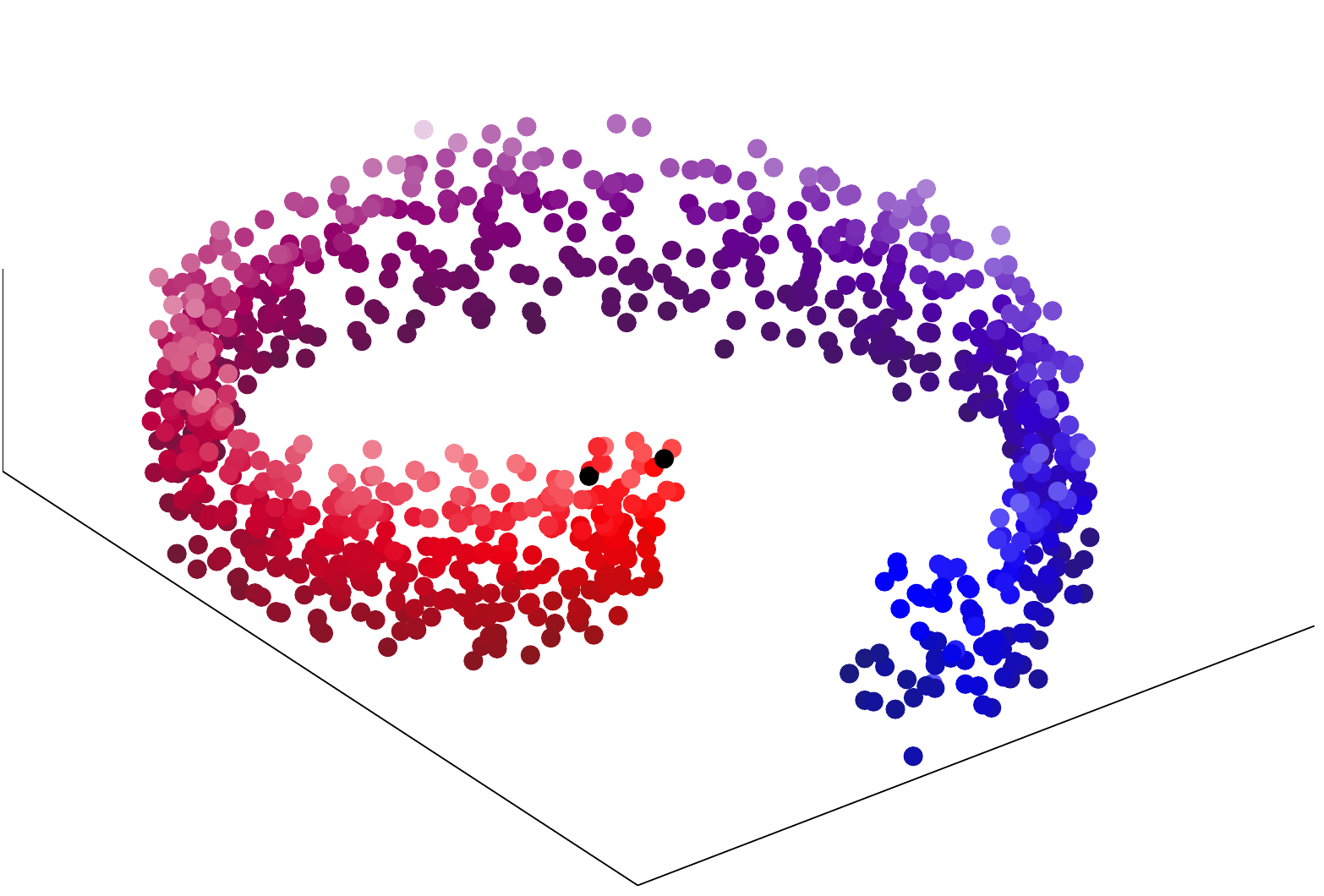} &
\includegraphics[height=3.7cm,width=2.7cm]{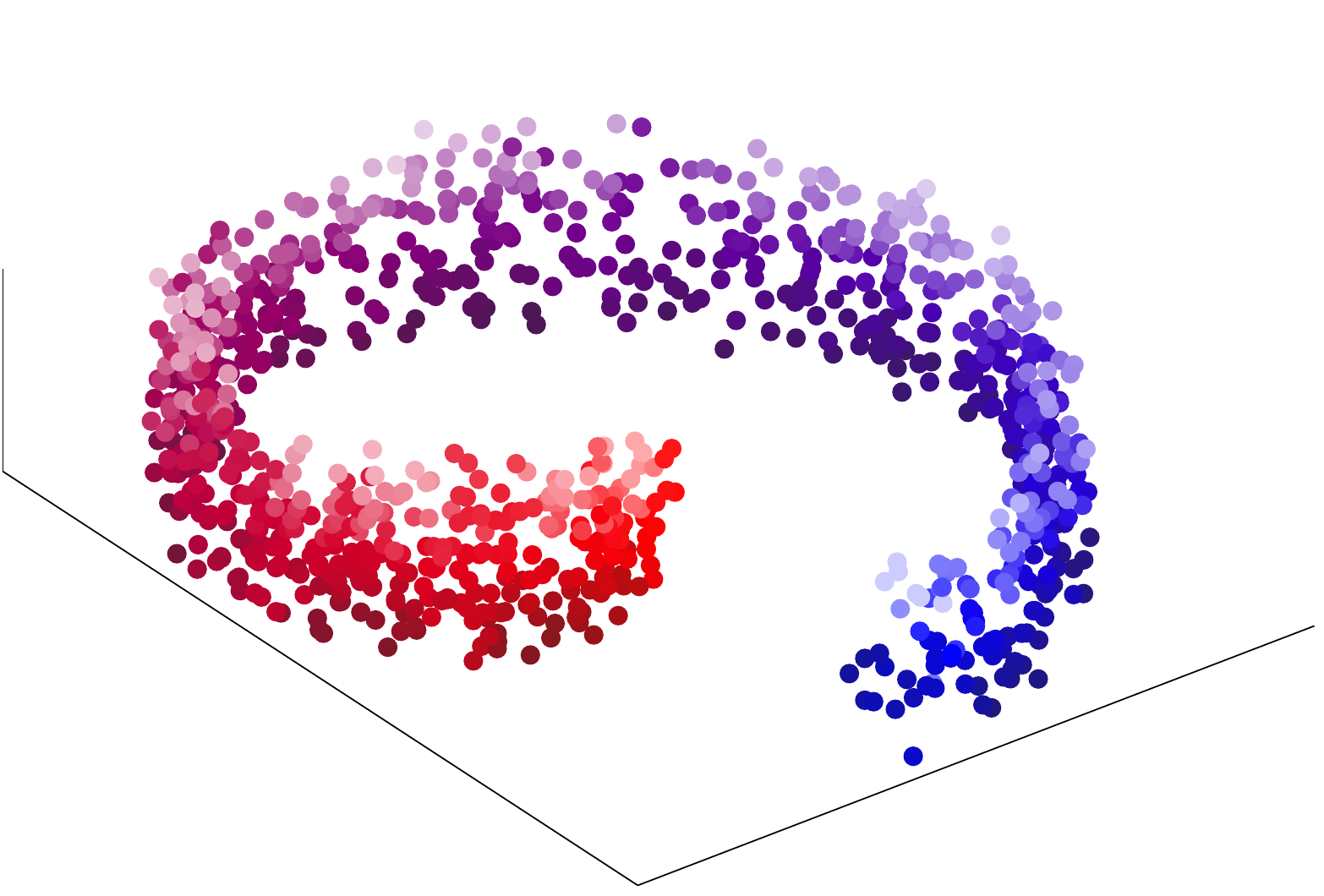} &
\includegraphics[height=3.7cm,width=2.7cm]{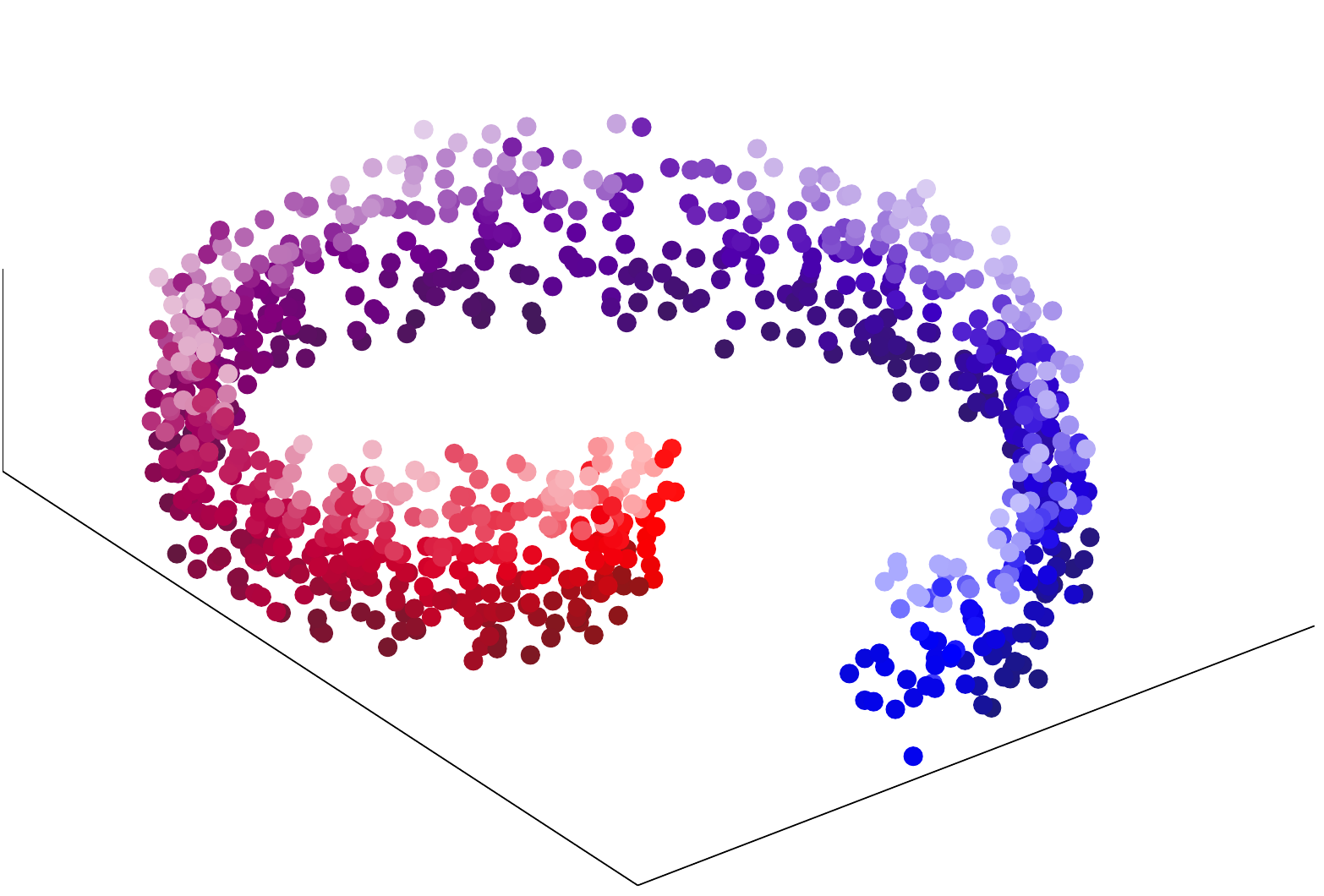} &
\includegraphics[height=3.7cm,width=2.7cm]{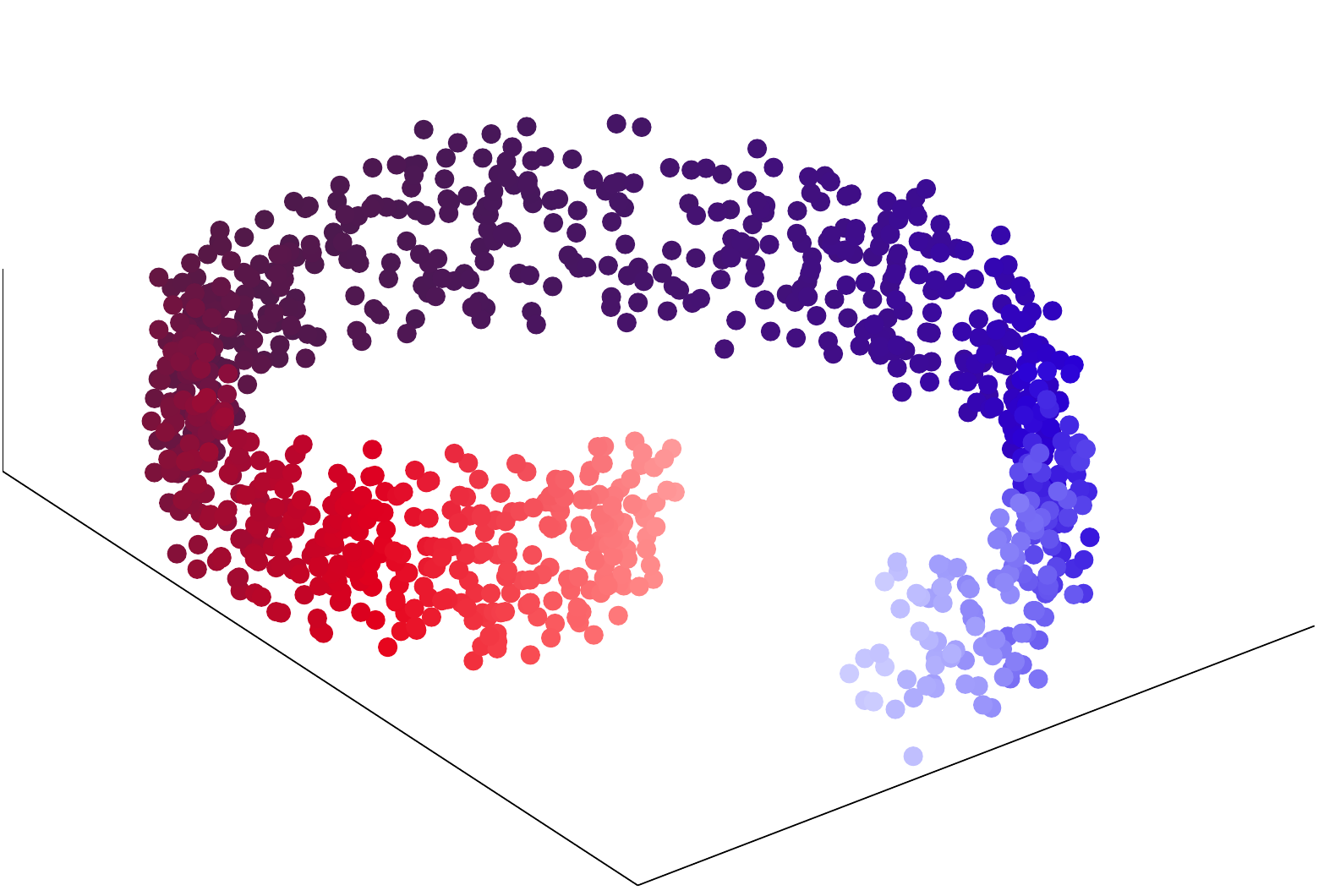} &
\includegraphics[height=3.7cm,width=2.7cm]{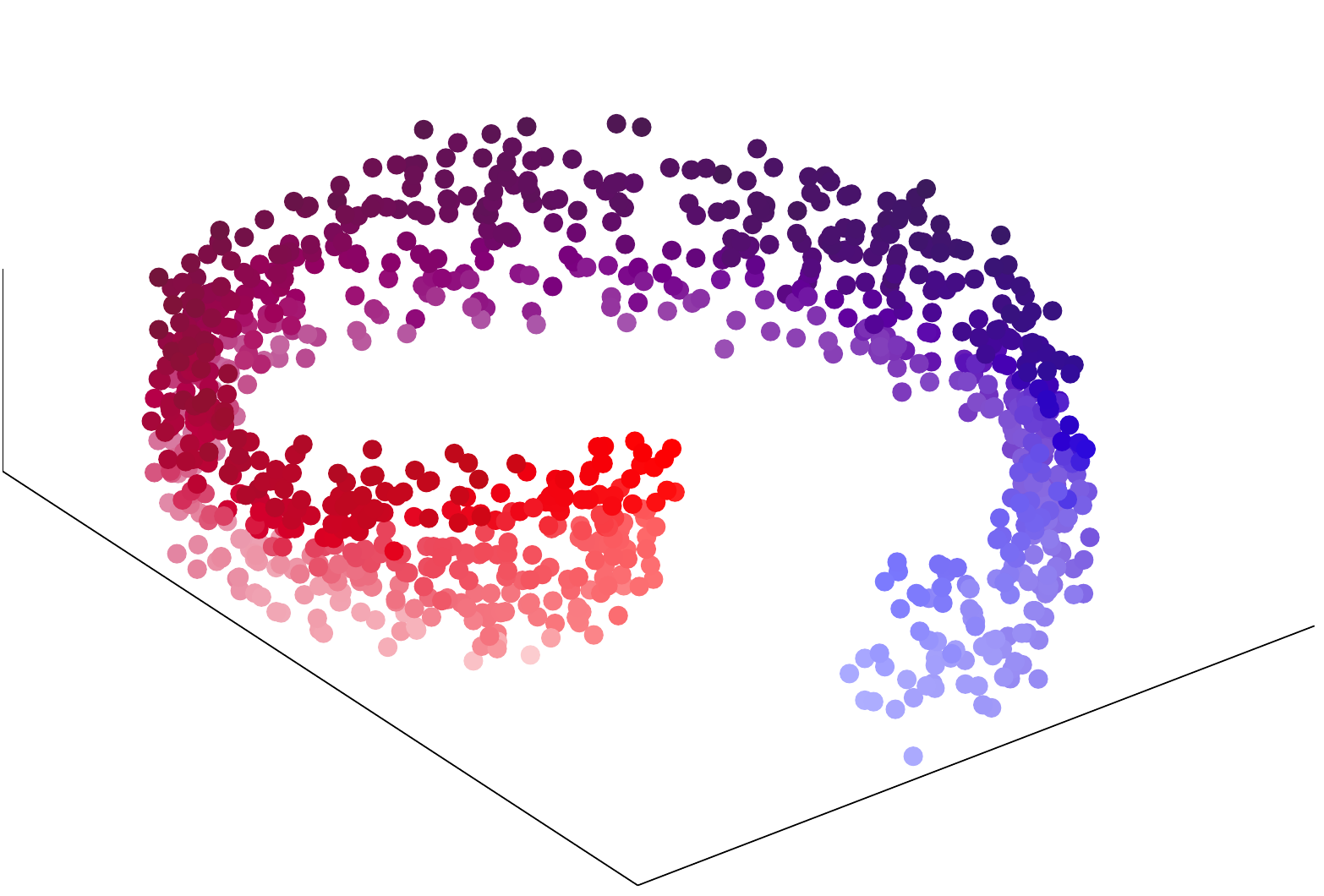} &
\includegraphics[height=3.7cm,width=2.7cm]{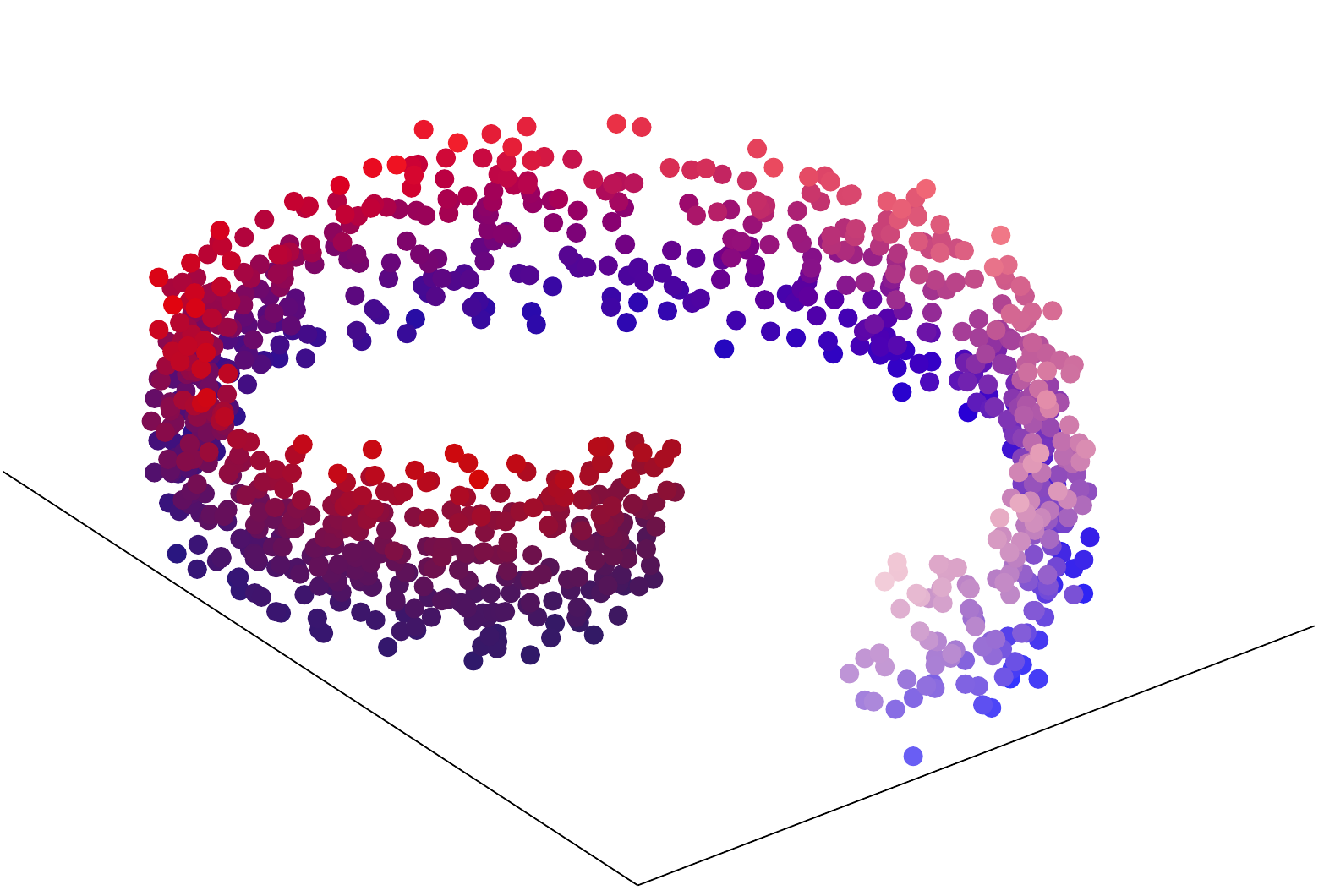} \\
\includegraphics[width=2.7cm]{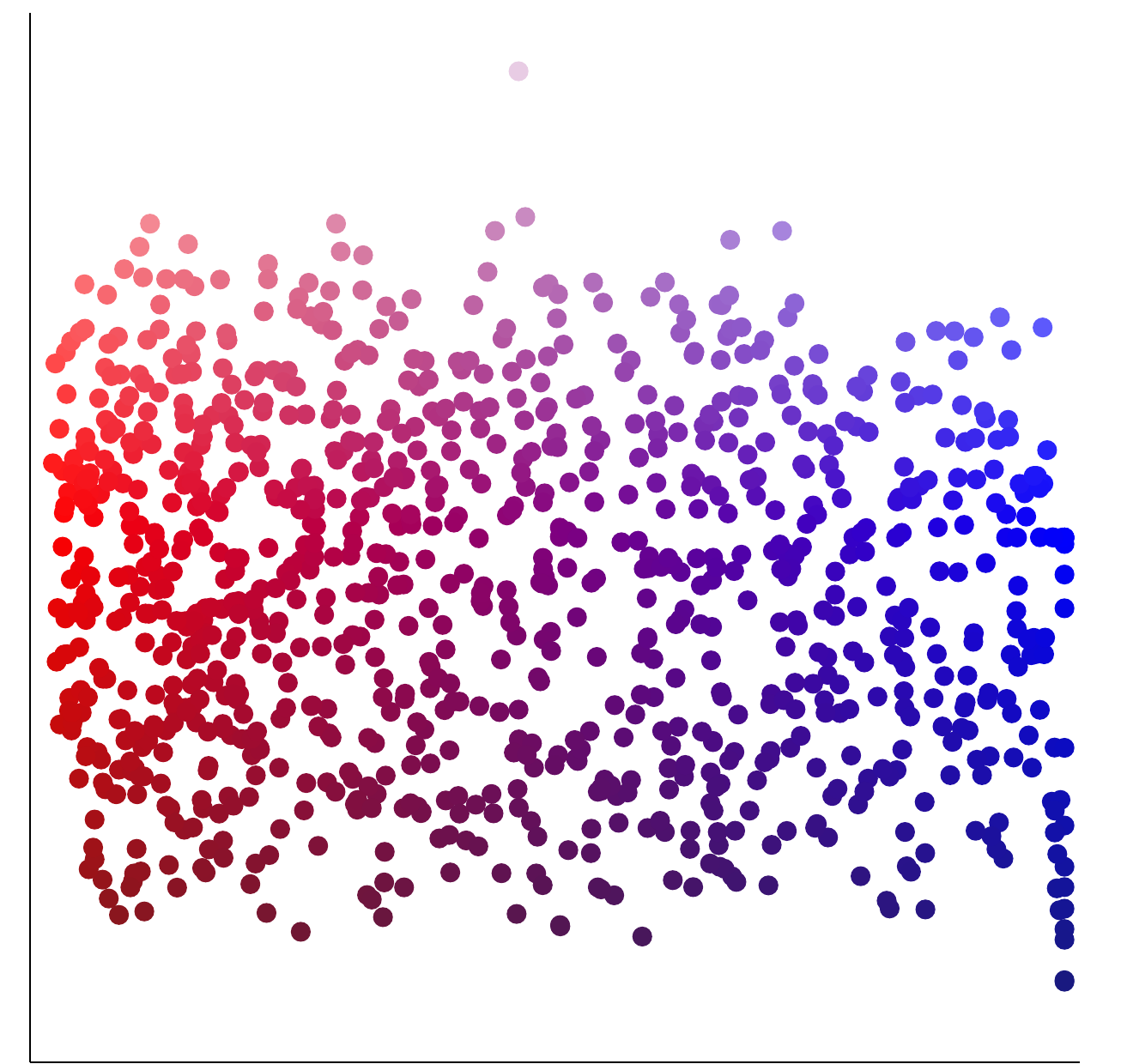} &
\includegraphics[width=2.7cm]{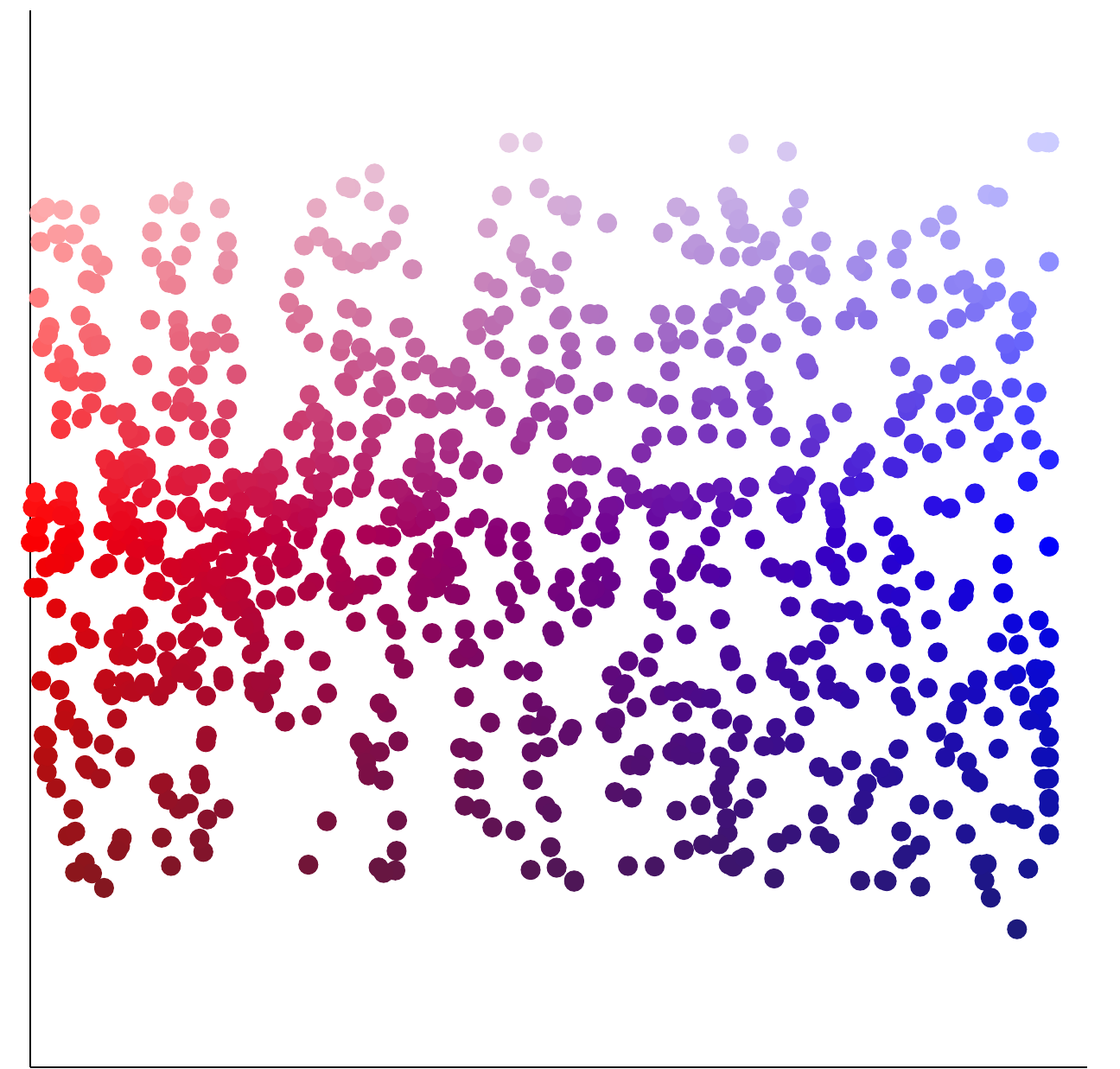} &
\includegraphics[width=2.7cm]{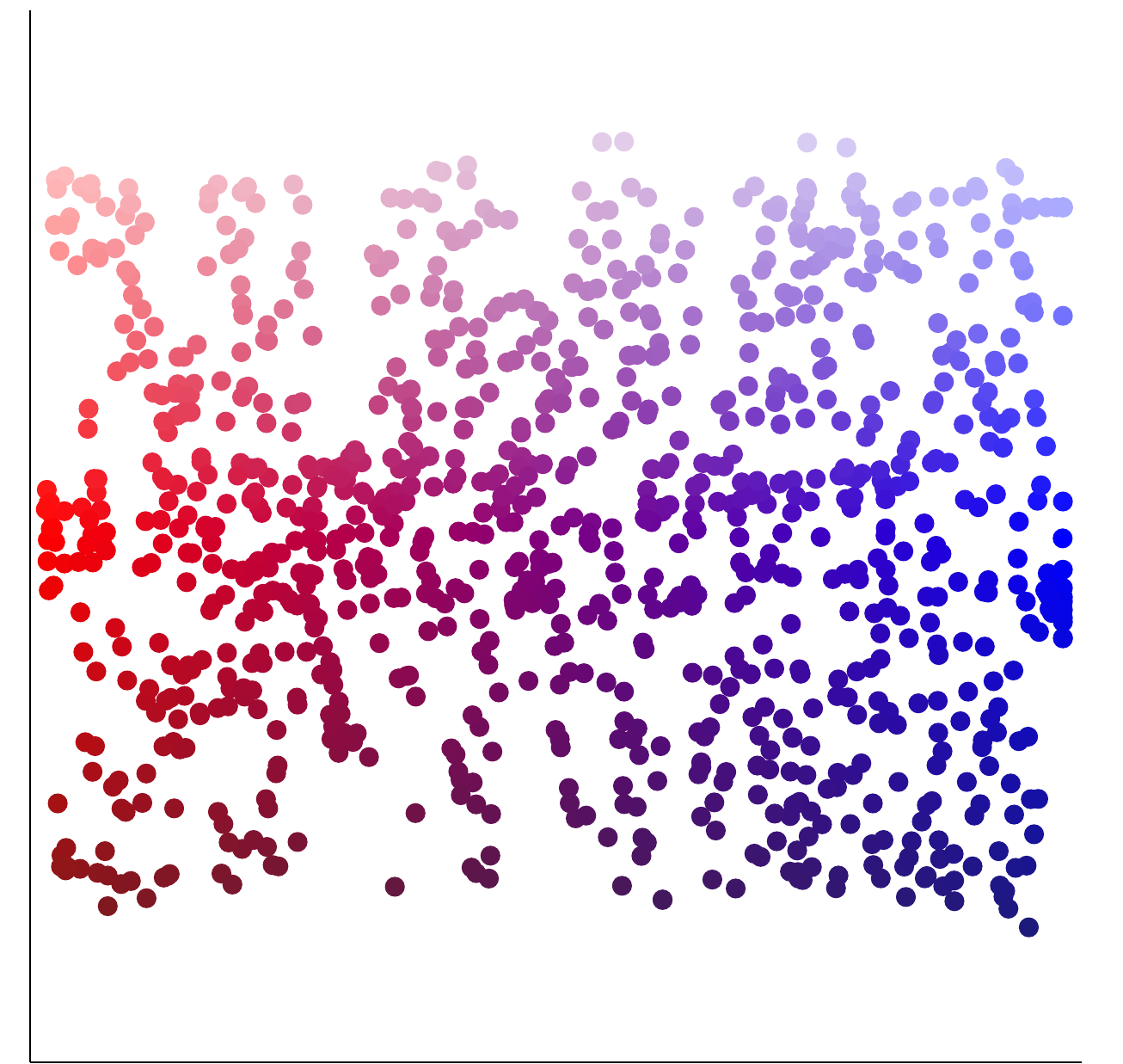} &
\includegraphics[width=2.7cm]{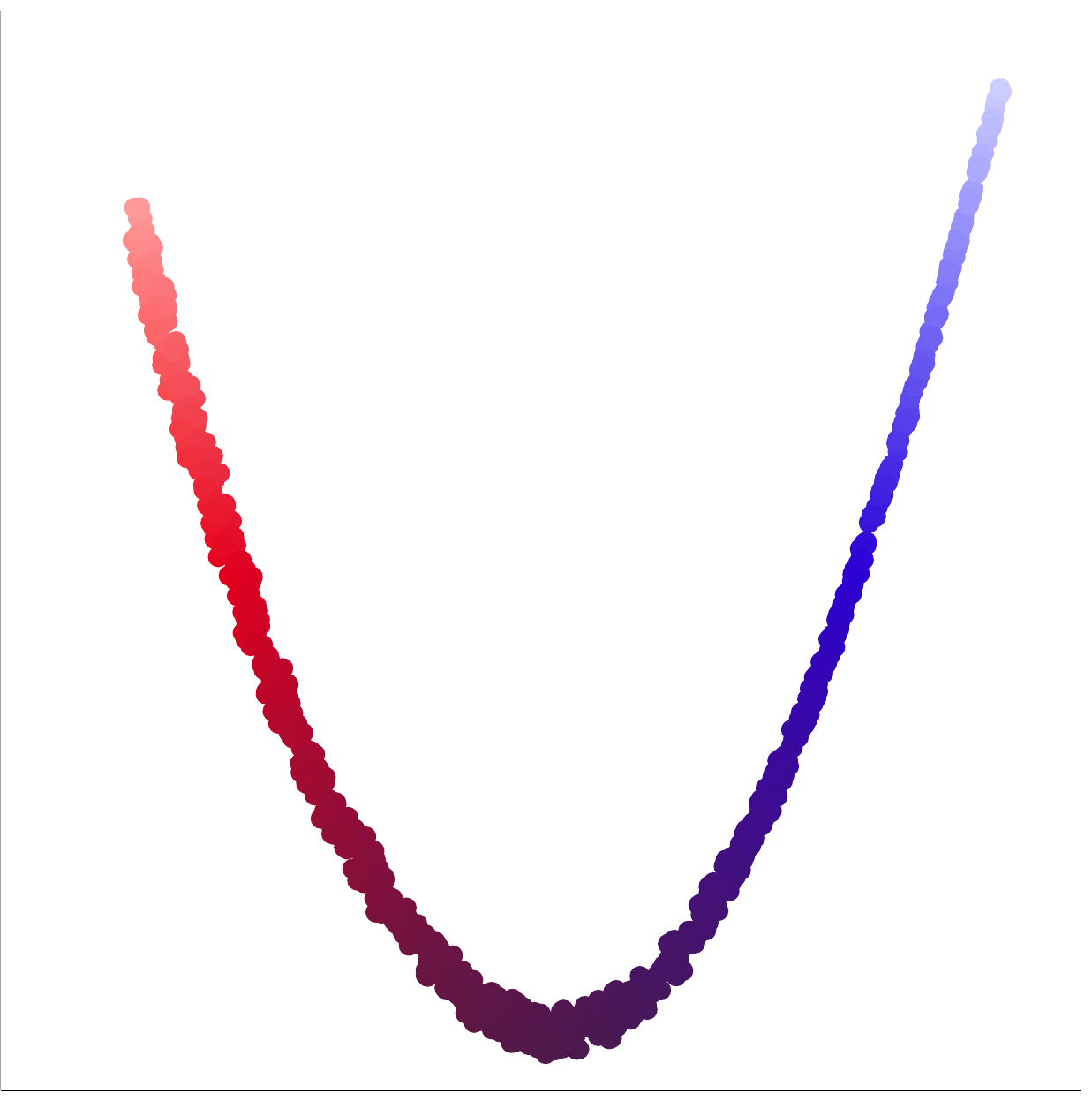} &
\includegraphics[width=2.7cm]{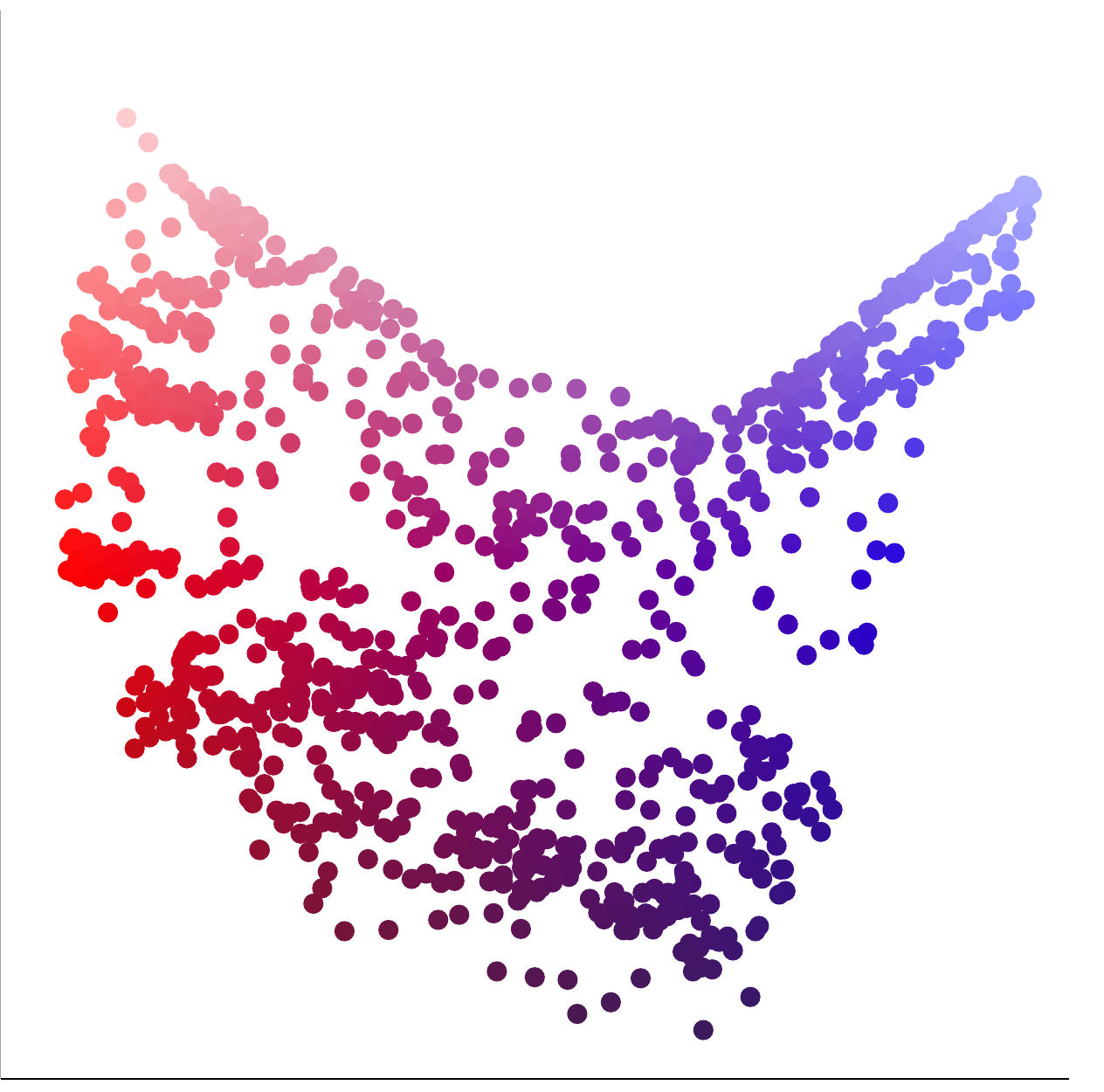} &
\includegraphics[width=2.7cm]{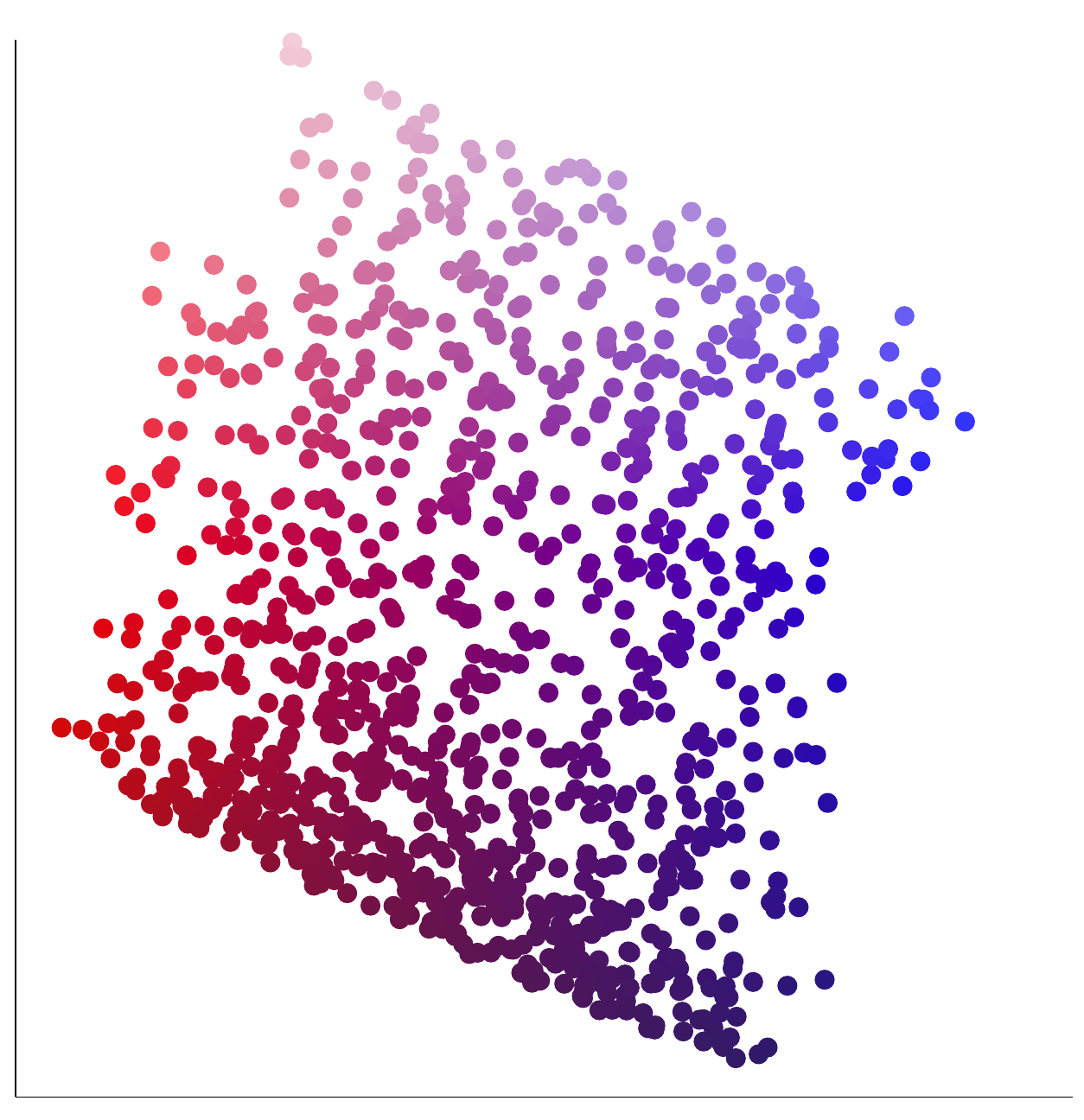}
\end{tabular}
\end{center}
\vspace{-0.25cm}
\caption{Dimensionality reduction (from 3d to 2d): unfolding a noisy swiss roll with SPCA using different metrics, LLE, Isomap and Charting.
In every case, $2000$ samples were used to learn the reduced dimensionality responses.
The number of nearest neighbors in LLE and Isomap was varied in the range [$2, 50$],
and the number of local models in charting was varied in the range [$2, 50$].
We show the solutions with best qualitative performance.
As stated above, in the SPCA cases, the parameters of the
algorithm to draw individual PCs were chosen to minimize the reconstruction
error (see Appendix).
}
\label{swiss}
\end{figure*}

Dimensionality reduction with SPCA consists of considering $d'<d$ components of the response vector.
Figure~\ref{swiss} shows the qualitative performance of SPCA in dimensionality reduction
by unfolding a {\em noisy} version of the popular 2d swiss roll embedded in $\Real^3$.
We compare SPCA to the solutions given by other popular dimensionality reduction
methods such as those based on geodesic computation (Isomap~\cite{Tenenbaum2000}),
spectral methods (LLE~\cite{Roweis00}), and local model coordination (charting~\cite{Brand03}).

Bottom row of Fig.~\ref{swiss} shows the scatter plot of the samples according to the
coordinates identified by each algorithm. In this transformed domain the variation in
the first dimension is depicted with a hue change from red to blue, while the variation
in the second dimension is depicted with a luminance change from dark to bright.
Top row of Fig.~\ref{swiss} shows the samples in the original domain colored according to
the color code found in the transformed domain.
Therefore, the quality of the curvilinear coordinates can be evaluated
in two ways: (1) by assessing the shape of the unfolded sets of samples, and (2) by assessing
the meaningfulness of the color distribution in the original domain.

\begin{figure}[b!]
\begin{center}
\hspace{-0.9cm}
\begin{tabular}{ccc}
\small SPCA $\gamma$ = 0  & \small SPCA $\gamma$ = 1/3 & \small SPCA $\gamma = 1$\\
\includegraphics[height=3.4cm,width=2.60cm]{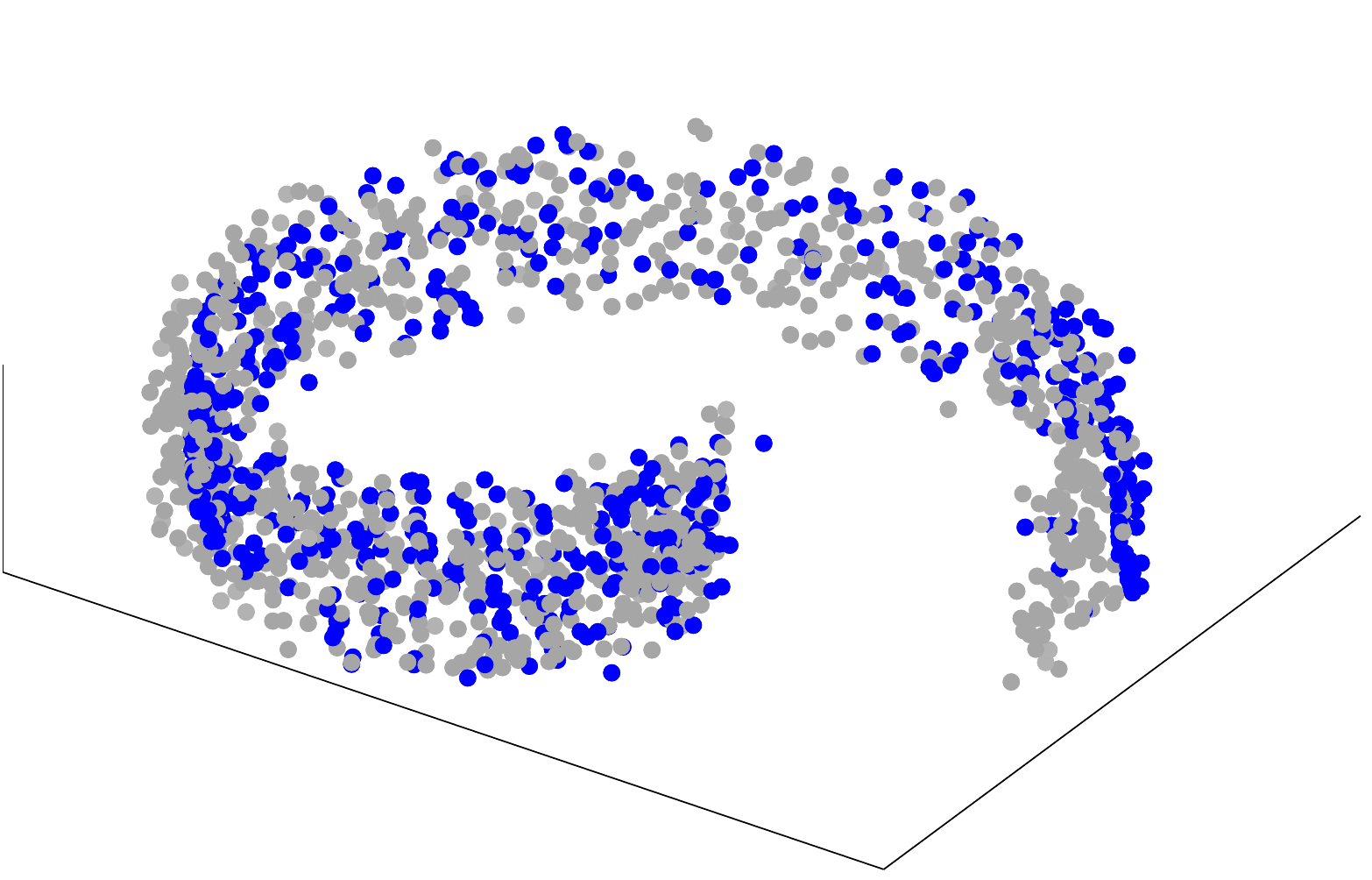} & \includegraphics[height=3.4cm,width=2.60cm]{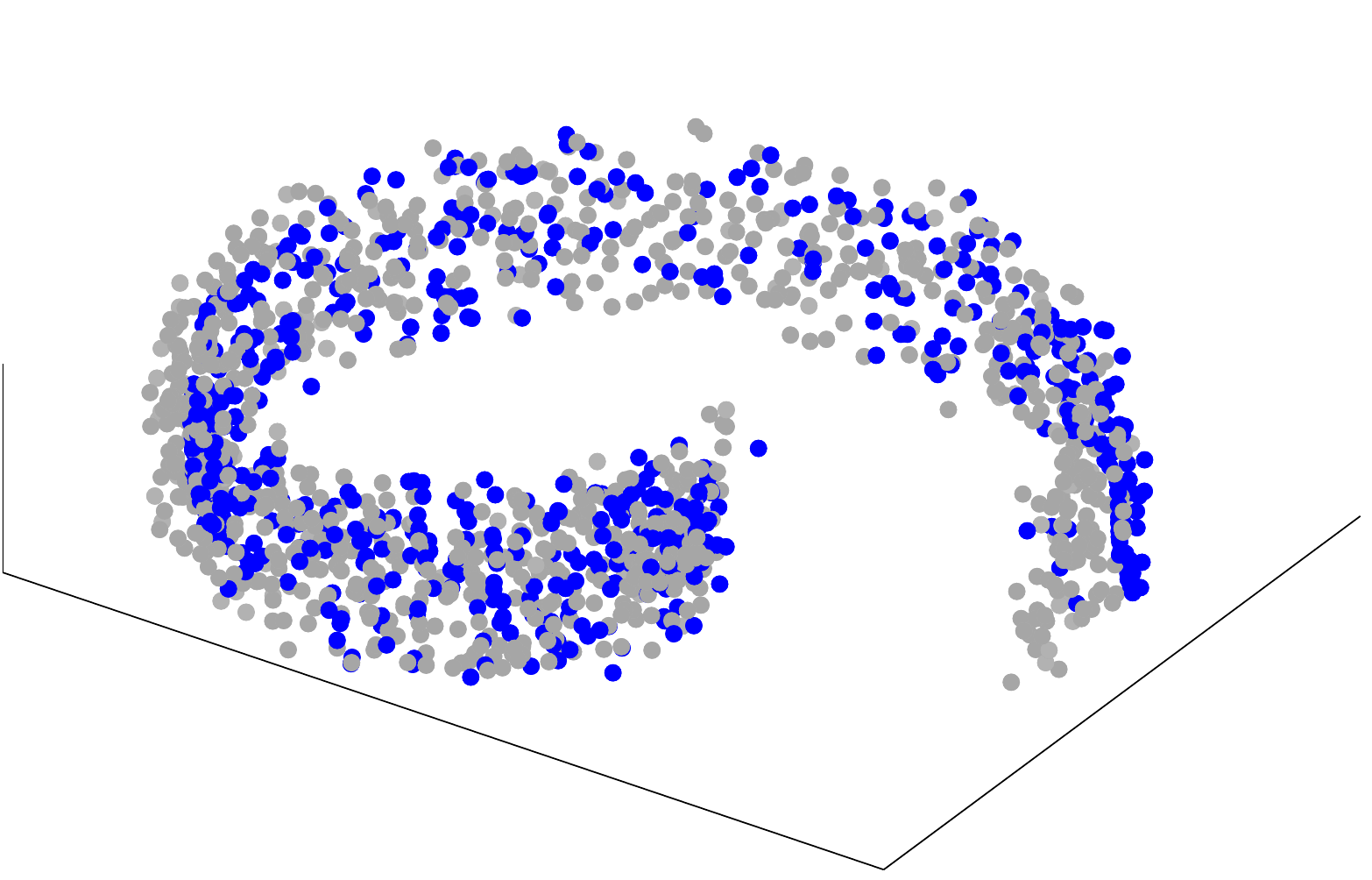} & \includegraphics[height=3.4cm,width=2.60cm]{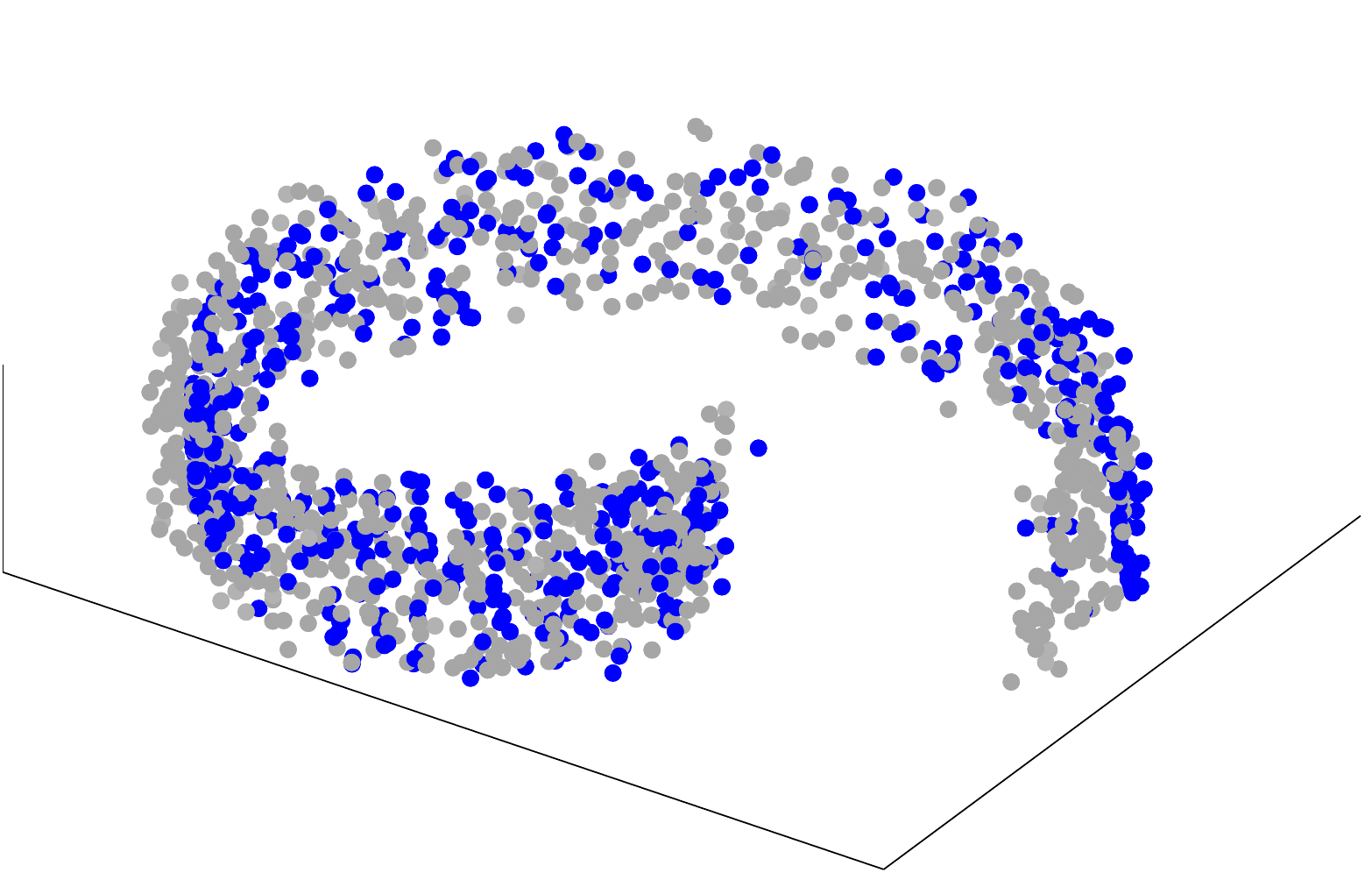}\vspace{-0.8cm} \\
\includegraphics[height=3.2cm,width=2.60cm]{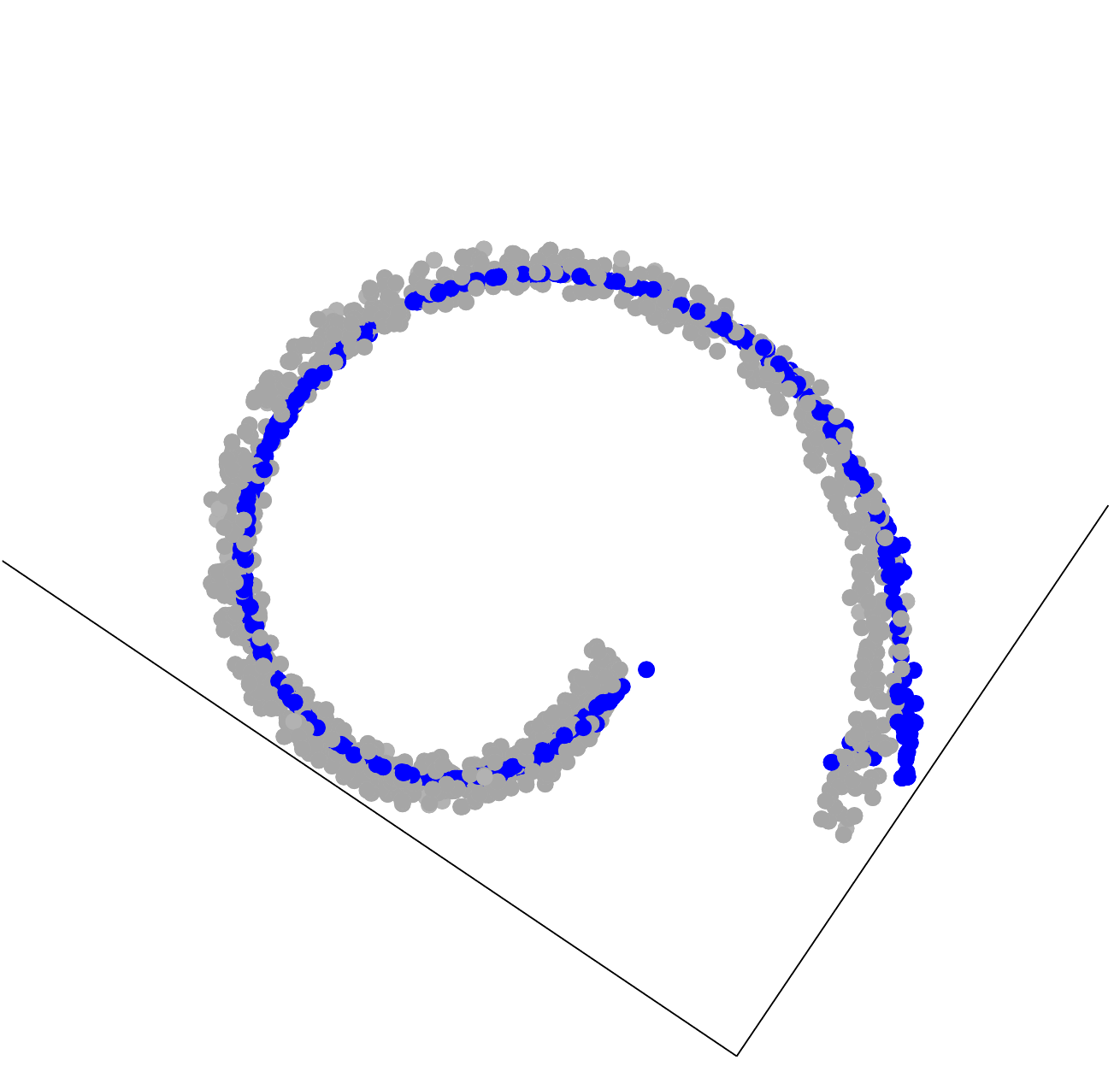} & \includegraphics[height=3.2cm,width=2.60cm]{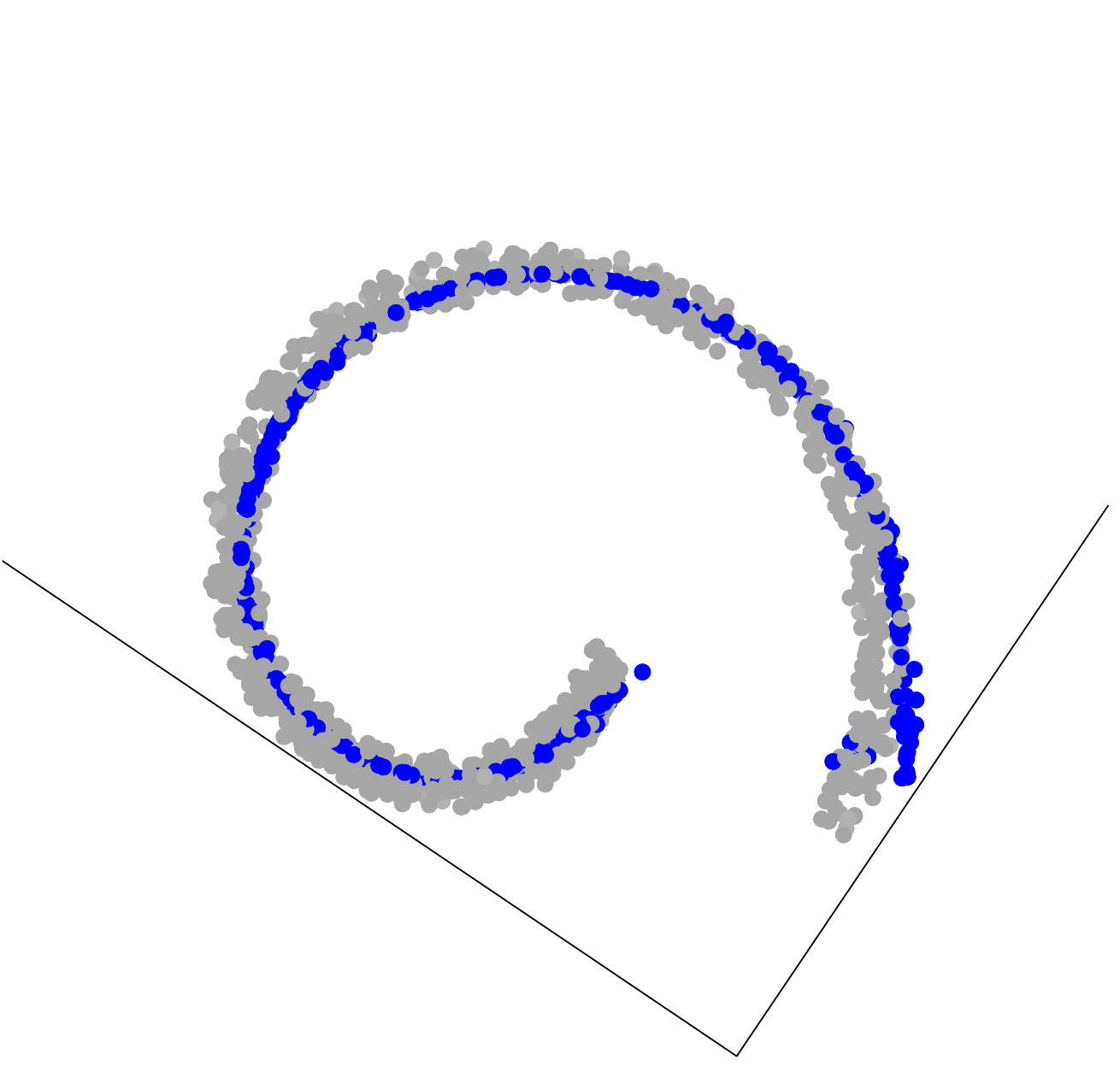} & \includegraphics[height=3.2cm,width=2.60cm]{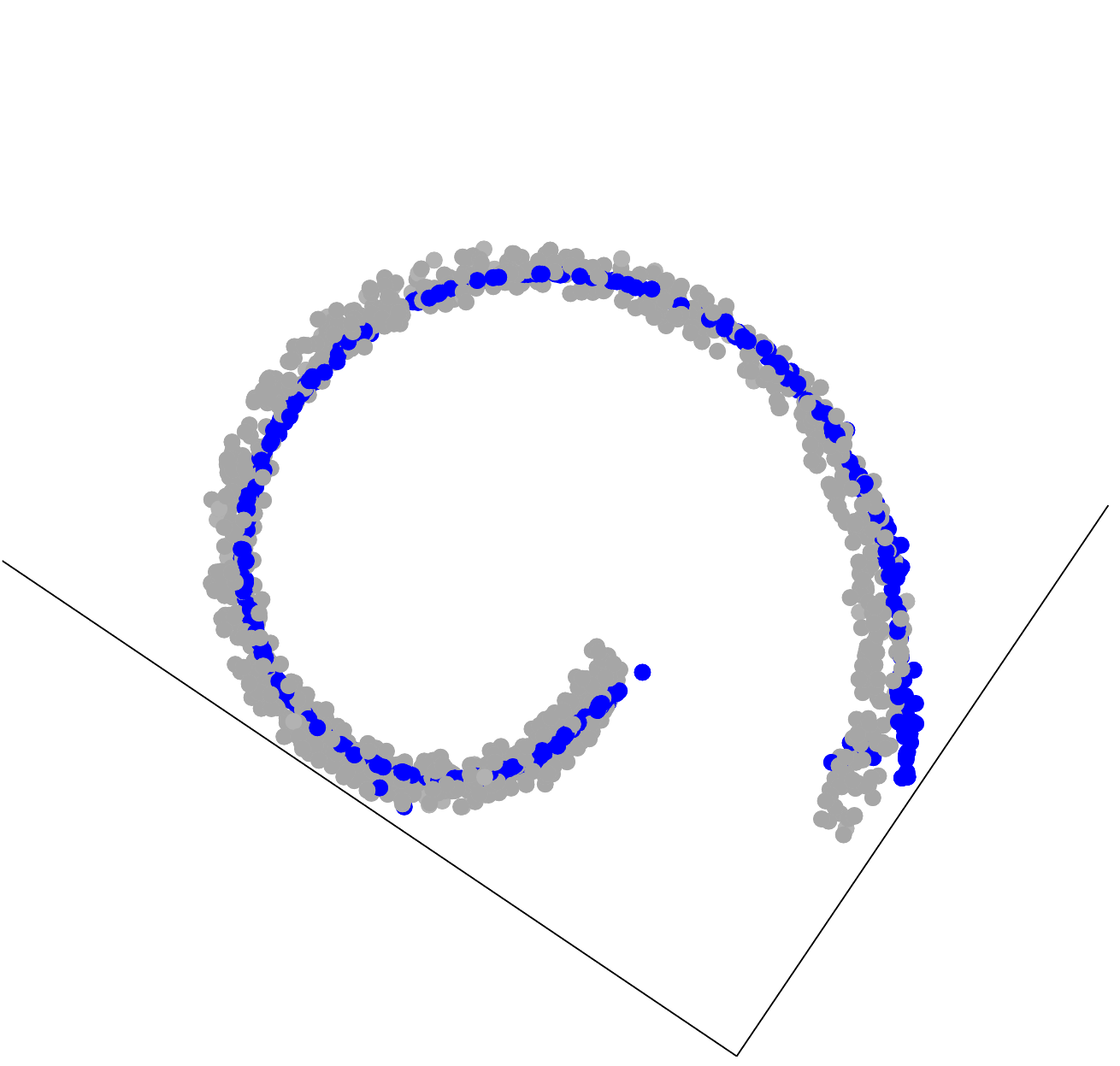}\\
\end{tabular}
\end{center}
\vspace{-0.25cm}
\caption{ Original samples (in gray) and SPCA samples with reduced dimensionality projected back
into the original domain (in blue). Top and bottom row show the same data from different view points to
illustrate the fact that, no matter the metric strategy, the inverse of SPCA properly identifies the
underlying 2d manifold.}
\label{swiss_inverse}
\end{figure}

It is observed a heavy warping of the tails of the embedding for LLE. As a result, the second
(dark/bright) dimension has no meaning in the input domain.
Color distribution (meaning of dimensions) is consistent in Isomap and the SPCA results no matter
the selected metric: the first (hue) dimension means angular position and the second (luminance)
dimension means height for a fixed angular position. It is reasonable that SPCA (which sorts the
dimensions according to entropy or variance) gives rise to the same dimension sorting as Isomap, which
considers the length of the geodesics. In this example the length (and the variance) along the
angular dimension is bigger than in the other dimension. On the contrary, the sorting in the case
of charting is the opposite.
Finally, it is worth to stress that SPCA (no matter the metric) gives rise to more regular and
uniform scatter plots in the transform domain.

On top of improved data visualization, the invertibility of SPCA allows to
project the low dimensional samples back into the original domain
revealing the latent manifold, Fig.~\ref{swiss_inverse}.

\subsection{Domain Adaptation}

\begin{figure}[t!]
\begin{center}
\small
\setlength{\tabcolsep}{0pt}
\begin{tabular}{cc}
\hspace{0.1cm} \textbf{Environment A:} &  \textbf{Environment B:} \\
\hspace{0.1cm} CIE D65 illumination &  CIE A illumination \\
\hspace{0.1cm} \includegraphics[width=3.6cm,height=3.6cm]{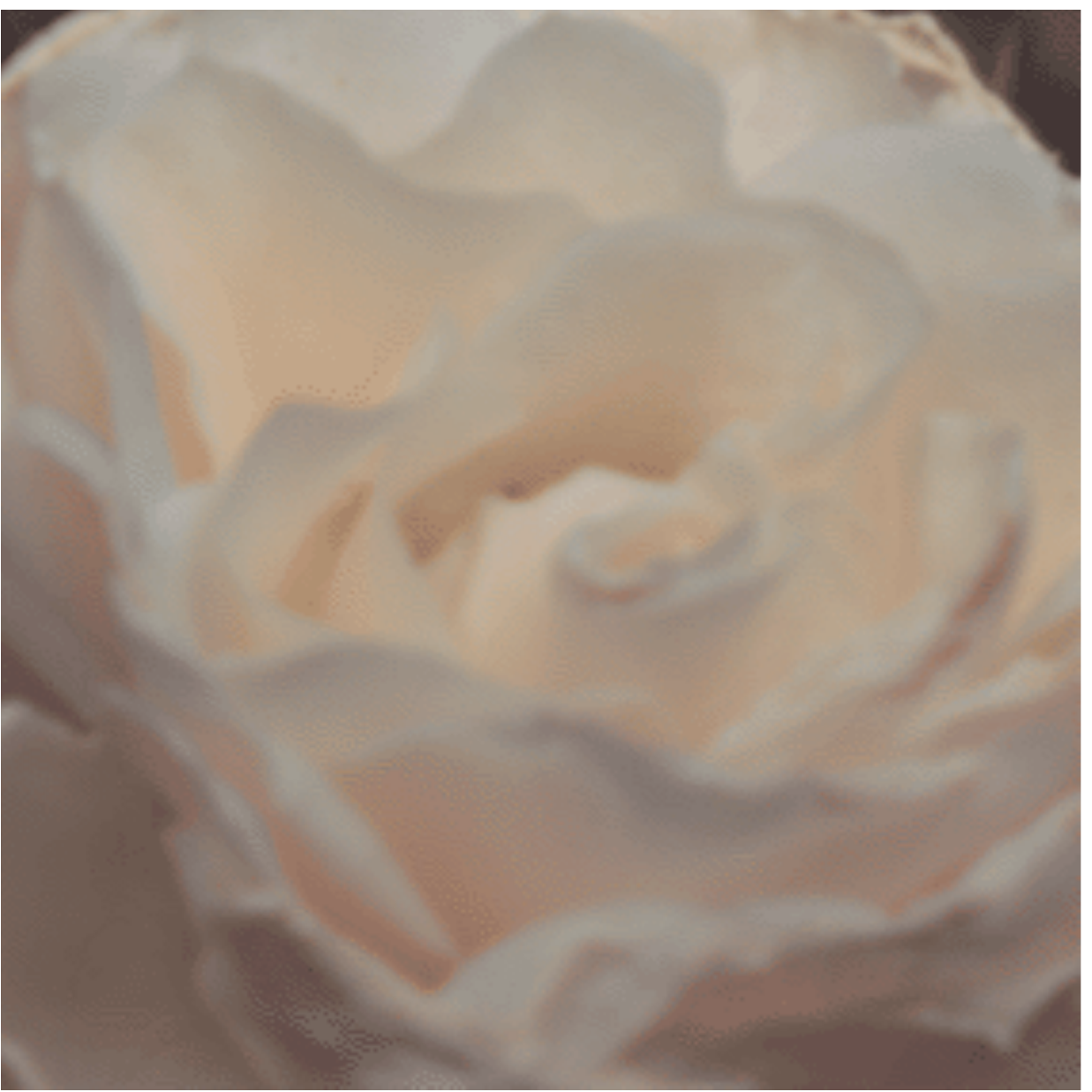} &
\includegraphics[width=3.6cm,height=3.6cm]{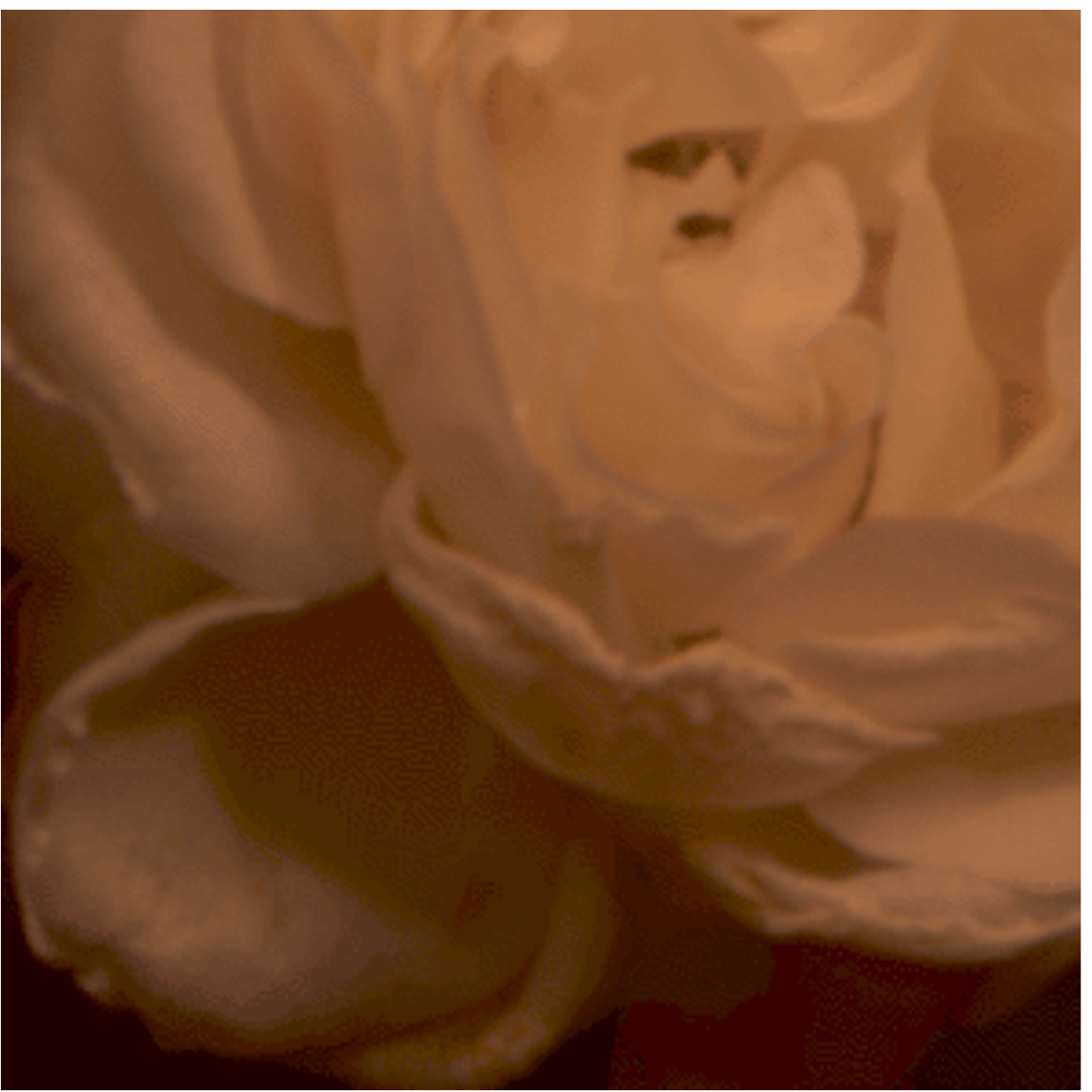} \\[0.4cm]
\multicolumn{2}{c}{
Linear Measurements and PCA+whitening Manifold Matching
} \\
\multicolumn{2}{c}{
\includegraphics[width=4.1cm]{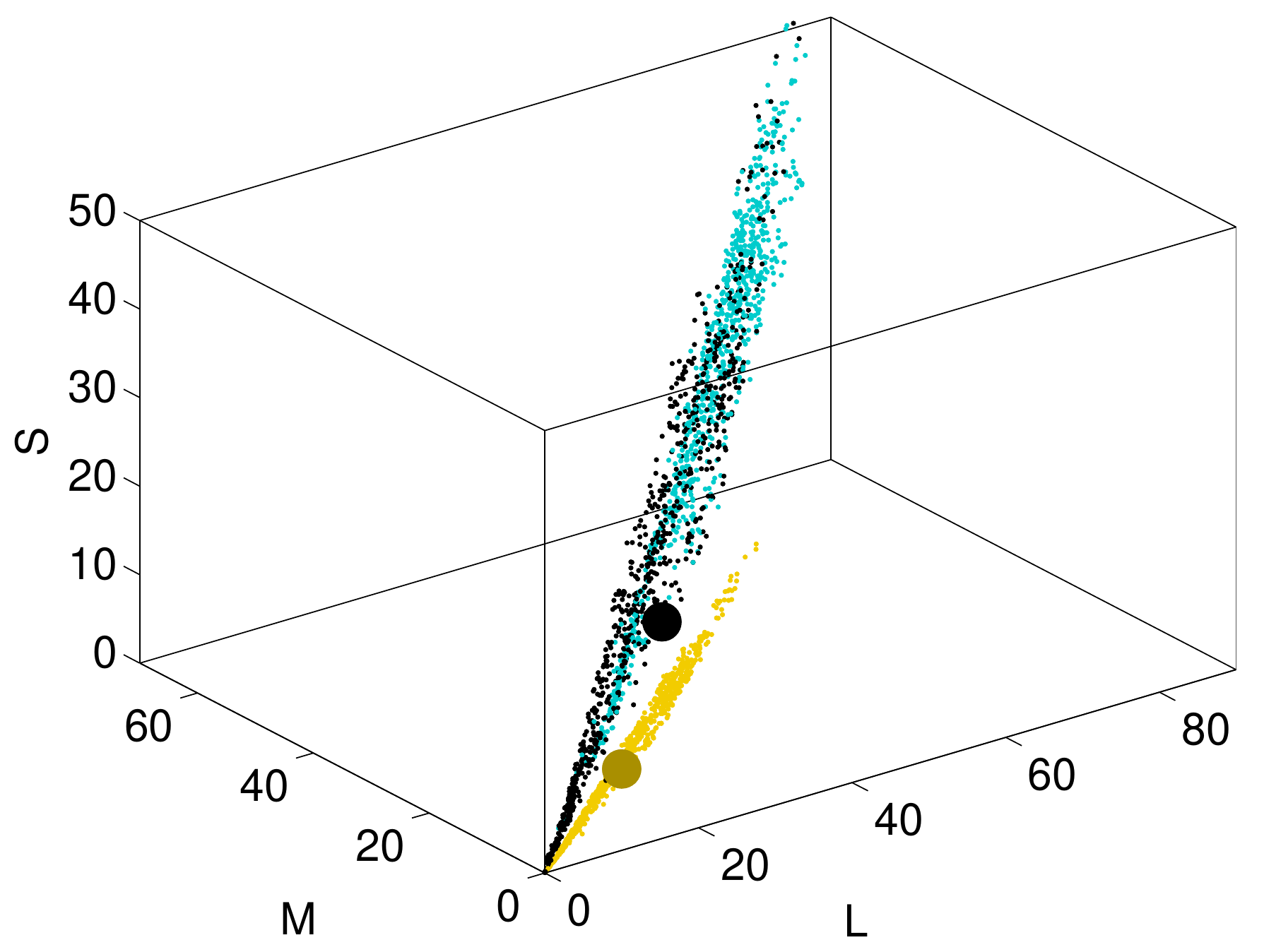}
} \\[0.2cm]
\multicolumn{2}{c}{
\includegraphics[width=8.1cm]{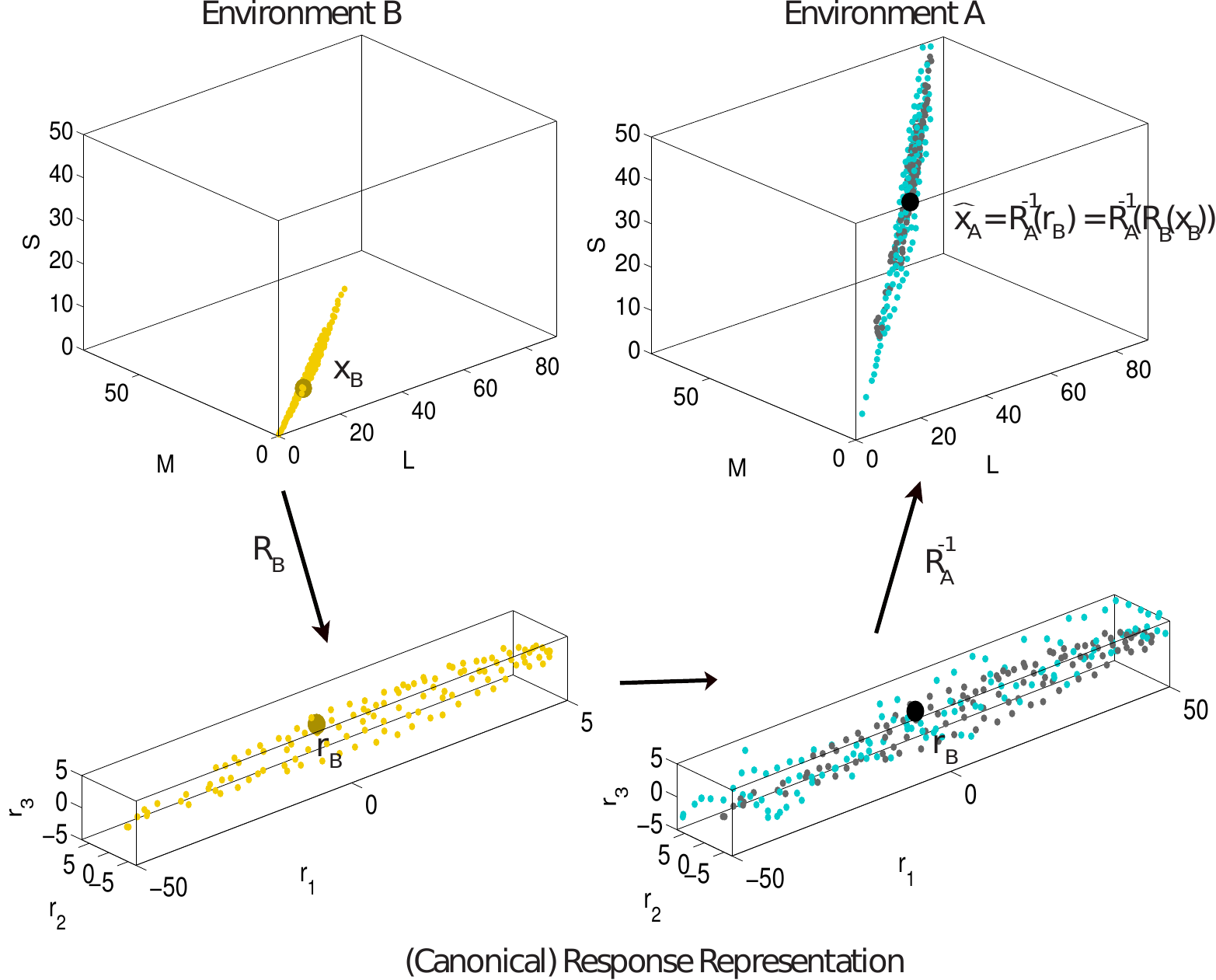}
} \\[0.2cm]
\hspace{0.1cm} PCA+whitening &  SPCA \\
\hspace{0.1cm} \includegraphics[width=3.6cm,height=3.6cm]{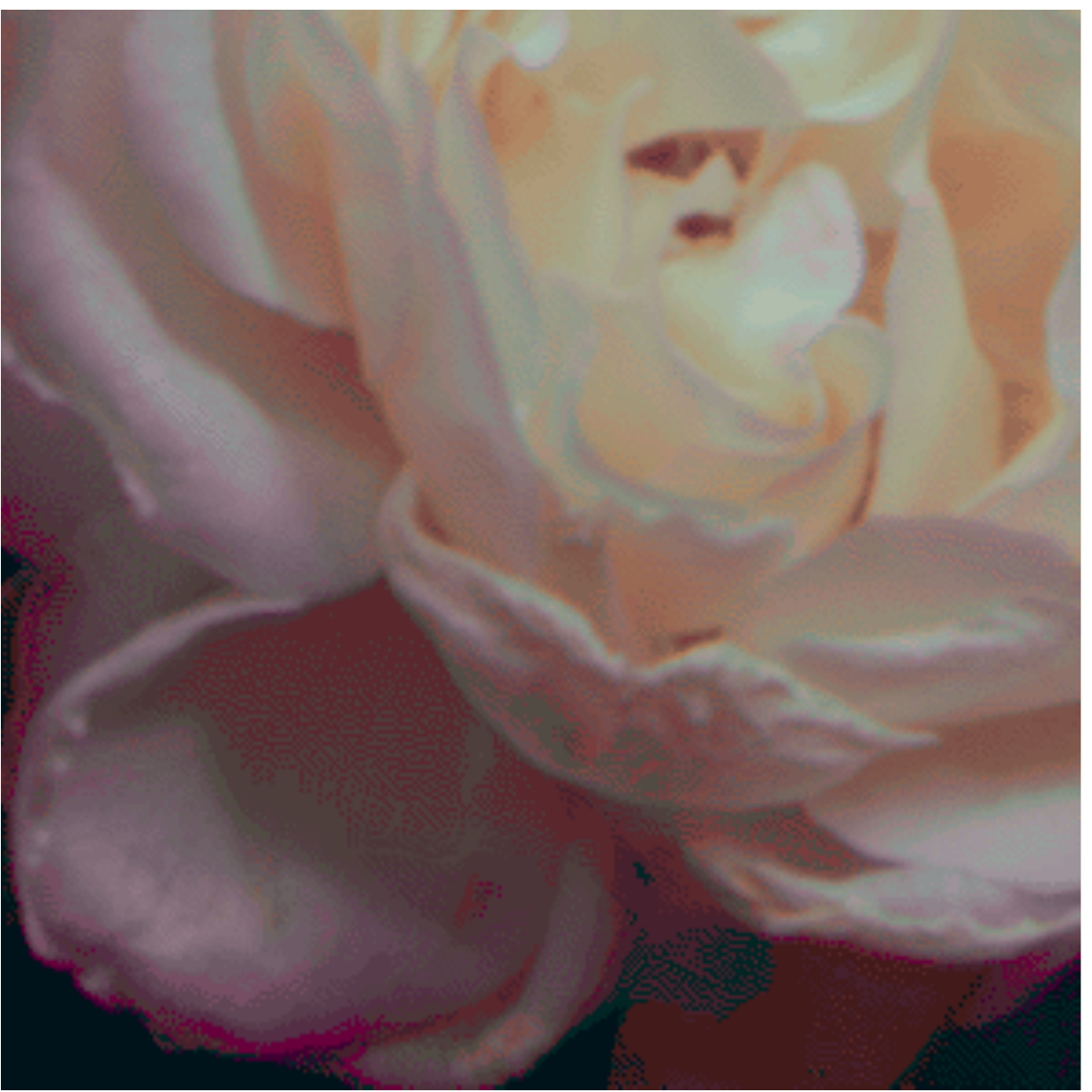} &
\includegraphics[width=3.6cm,height=3.6cm]{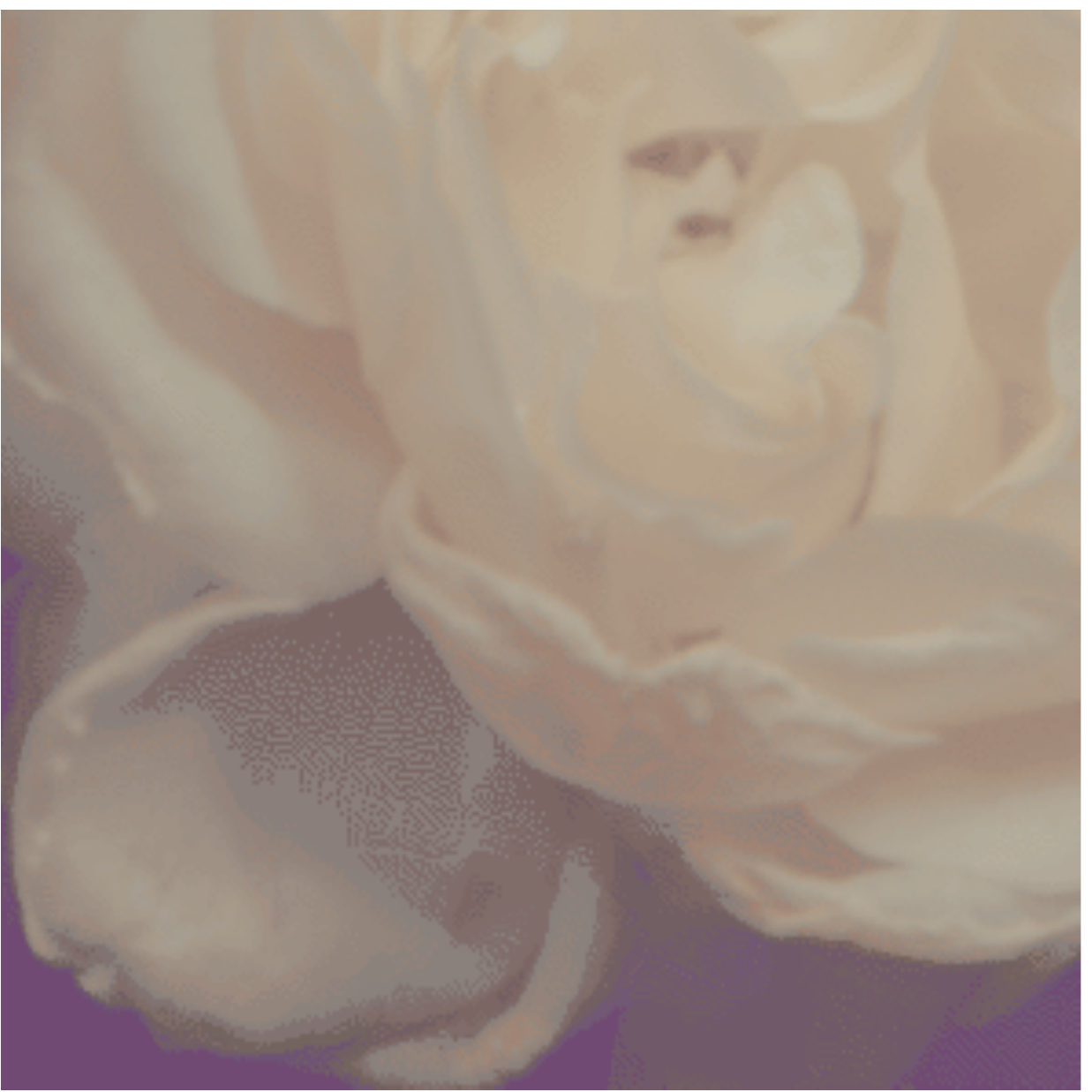} \\
\end{tabular}
\end{center}
\caption{Dataset shift compensation by using the corresponding pair concept and SPCA. First row: similar objects acquired in different spectral and geometry illumination conditions. Second row: manifolds of the colors in the two images are shown in blue (CIE D65) and in yellow (CIE A). In black we plot the colors of CIE A after color compensation using PCA+whitening. One particular color in the CIE A case has been highlighted (large yellow dot). The corresponding corrected color has been highlighted in black.
Diagram in the third row: procedure for color compensation using SPCA. The key for color compensation is inverting the
transform of points acquired in the B environment (gray points) using the inverse of the transform learned for the samples
acquired in the A environment. Bottom row: reconstructed images after color compensation by PCA+whitening and the proposed SPCA.
}
\label{adaptation}
\end{figure}

Color constancy under change of observation conditions is a challenge for machine learning techniques because it involves
strong changes in the samples coming from the same physical objects. Changes in linear measurements (tristimulus values) may arise from changes in the spectral illumination and from changes in the geometry of surfaces and light sources. In general such changes are non-linear, specially those coming from geometry changes, so they cannot be compensated through linear techniques. The examples of the top row of Fig.~\ref{adaptation} show two pictures of similar objects illuminated with different spectral radiance (CIE D65, left and CIE A, right) under different illumination angles. The differences in spectral radiance induce a rotation of the manifold in the color space, and the change of surface and illumination geometry induce the presence of shadows. This latter effect results in a different distribution of samples within the PDF support. In the CIE A case the low luminance region is relatively more populated. According to this, color constancy is an appropriate problem to assess the ability of a manifold learning technique in (non-linear) domain adaptation. In this section, we apply the proposed method for color compensation, or more specifically for manifold matching, as an example of domain adaptation. We compare the results with a classical linear technique for chromatic adaptation based on PCA and whitening~\cite{Webster97}.

Our strategy for domain adaptation is related to the corresponding pair procedure, which has been proposed in the computational color vision literature using psychophysically-based models~\cite{Capilla04}. Given a color vision model, ${\mathcal M}$, that describes the perception, $P$, of a test, $K$, in some adaptation state (e.g. under observation conditions, $S$), $P={\mathcal M}(K,S)$, the corresponding pair, $K'$ in a different adaptation state, $S'$, is given by~\cite{Capilla04}:
\begin{equation}
      K' = {\mathcal M}^{-1}({\mathcal M}(K,S),S').
      \label{corr_procedure}
\end{equation}
In our manifold learning context, this reduces to considering the color statistics of natural scenes under illuminations $S$ and $S'$. Then, the $K$ test points are transformed using SPCA with the set of colors under illumination $S$, and inverted back using the set of colors under $S'$. The match between invariant (adapted) representations obtained using SPCA from different linear (variant) representations, as suggested by Eq.~\eqref{corr_procedure}, is related to the nonlinear canonical correlation analysis~\cite{Verbeek02}.

As shown in Fig.~\ref{adaptation}, linear transforms can compensate the rotation due to the spectral radiance change but the resulting set is still biased towards the low luminance region. The diagram in Fig.~\ref{adaptation} illustrates the procedure for color compensation using SPCA. Transforms leading to the corresponding latent coordinates of the manifold in the environments $A$ and $B$ may be used to estimate the position in environment $A$ of new points measured in environment $B$. Unlike in the linear adaptation case above, the proposed nonlinear transform not only removes the yellowish appearance, but additionally reduces the shadows, as expected from a better PDF matching. In particular, note how the highlighted point $\mathbf{x}_B$ (the same one as above) results in a white, {\em higher luminance} corresponding point $\mathbf{\hat{x}}_A$.

\subsection{SPCA in classification}
The manifold dependent metric associated to SPCA (Eq. \ref{generalSPCA_jacobian} \ref{metrica}) can be used in classification problems (e.g. k-nearest neighbors) as alternative to Mahalanobis distance.
Euclidean distance in the whitened SPCA domain is a generalization of Mahalanobis metric (which is the Euclidean distance in the whitened PCA domain). Similar generalizations can be done with the related methods Principal Polynomial Analysis (PPA) \cite{Laparra14a} and Dimensionality Reduction via Regression (DRR) \cite{Laparra15a}.

An example of how this kind of metrics can benefit the classification can be found in \cite{Laparra14a} (Figs. 3 and 4), where we used PPA. The reason is that given a sample, the loci of equidistant neighbors is an ellipsoid which is oriented according to the local structure of the manifold (and not a sphere). This minimizes the number of wrong-class samples in the selected neighbors. On top of this advantage (shared by PPA-based and DRR-based metrics), SPCA-based may be even better since discrimination will be higher in high density regions thus reducing the average misclassification error.

\section{Summary}\label{conclusions}

Here we introduced the full technical details of Sequential Principal Curves Analysis \cite{Laparra12,Laparra15b}.
It is a nonlinear feature extraction method appropriate to design artificial sensory systems and to analyze the rationale of biological sensory systems. See also \cite{Malo06b,Laparra14a,Laparra15a} for prequels and sequels.

Here we introduced the general technical motivations of SPCA: data unfolding along PCs, followed by local metric change leads to
to easily interpretable nonlinear independent components.
Moreover, splitting the \emph{infomax} problem into two stages (unfolding and equalization) allows us to propose data representations guided by alternative optimization criteria such as \emph{error minimization}.
SPCA can be seen as a particular solution to the design of optimal feature extraction for
either independence or transform coding when imposing the preservation of local geometry
in addition to the \emph{infomax} or the \emph{error minimization} goals.
Following the first and secondary PCs of the manifold ensures the interpretability
of the identified features, as opposed to spectral, kernel, projection-pursuit, and
neural networks methods for manifold learning.

We explicitly checked the underlying assumptions of SPCA in synthetic and
real examples: (1) secondary PCs can be
used to define locally orthogonal subspaces, (2) the proposed transform
converges so that SPCA is invertible, and (3) nonlinear ICA and
optimal transform coding are available using SPCA by choosing the appropriate
metric.
We showed examples of the use of SPCA in dimensionality reduction, domain adaptation,
and classification (by generalizing the Mahalanobis distance).

\section{Acknowledgements}\label{aks}

This work has been supported by the Spanish Ministry for Science and Innovation under grants BFU2014-58776 and TIN2013-50520.
The authors thank Gustau Camps-Valls and Gema Denia for their support before a conference presentation of SPCA \cite{Laparra08b}.

\section[Appendix: Drawing one Principal Curve]{Appendix: \\ A bottom-up approach to draw \\ one Principal Curve}

In this appendix we present a local-to-global algorithm to draw \emph{first} \cite{Hastie89} or \emph{secondary} \cite{Delicado01} Principal Curves. This is just an \emph{instrument} to build the individual PCs required in the sequence followed in SPCA.
This element of the general SPCA framework could be implemented with already reported bottom-up algorithms such
as those in~\cite{Delicado01,Einbeck05,Einbeck10}. However, given the unavailability of Matlab working code,
we developed our own algorithm where the local structure is analyzed in terms of local PCA.

The proposed PC algorithm follows a local-to-global or bottom-up approach since it identifies the local structure around
the selected Principal Curve origin, and progressively builds the curve from that origin.
In this appendix we show the geometric meaning of the parameters of the technique (instrumental parameters)
and provide a particular procedure to tune them for the dataset at hand:
the appropriate set of parameters is the one that minimizes the projection error onto the
Principal Curve. This procedure is consistent with the original definition of Principal Curves
and with the definition of Principal Components in linear PCA.
\vspace{0.15cm}

\textbf{Algorithm.}\label{onePC}
This procedure draws one PC from a specific point, ${\bf x}^o$, in a particular direction, $B^i$.
The algorithm uses PCA to find linear directions that {\em go through the middle} of the dataset.
Hence, the proposed PC consists of segments obtained using local PCAs.

The procedure starts from ${\bf x}^o$, where a local PCA is computed.
Then, the most similar eigenvector to the desired direction is taken.
The eigenvector is modified so that it points to the mean of the data in
the direction ahead. The new point of the PC, ${\bf x}^1$, is obtained
by making a step from ${\bf x}^o$ in the direction of the modified eigenvector.
The obtained segment goes from ${\bf x}^o$ to ${\bf x}^1$. The rest of the curve
points are obtained applying the same procedure departing from the new points
until one of these stopping criteria is met:
(1) if the distance between the new point, ${\bf x}^{n+1}$, and some of the previously
drawn PCs is less than a threshold $\tau$, the algorithm stops since the current
PC is almost crossing a previous PC, and (2) if the distance between the new point,
${\bf x}^{n+1}$, and all the points in the manifold is bigger than a given
distance, $d_{out}$, the algorithm stops since ${\bf x}^{n+1}$ is assumed to be out
of the manifold.

Specifically, each point ${\bf x}^{n+1}\in$ PC is computed from the previous point,
${\bf x}^n$, and the reference matrix, ${\bf B}_n$, following the procedure in
Algorithm~\ref{alg:pc}.

The parameters of the algorithm to draw one PC at a given point basically
control the {\em rigidity} of the curve. These parameters include the size of the
local neighborhood (set using the $k$-neighbors rule), the stiffness $q$, and
the step size $\tau$ (see the table Algorithm~\ref{alg:pc}).

Intuitively, the bigger the locality, the stiffness and the step size, the bigger
the rigidity, so the obtained Principal Curves approach the global linear principal
components, thus reducing the flexibility of the technique to describe curved manifolds.

\begin{algorithm}[t!]
\small
 \caption{Draw {\em one} Principal Curve at a given point.}
 \begin{algorithmic}
 \STATE \textbf{Input}: dataset $\{{\bf x}^i| i=1,\ldots,N\}$
 \STATE \textbf{Set parameters}: Line increment $\tau$, stiffness $q$, and $k$ nearest neighbors
 \STATE \textbf{Output:} PC formed by a set of basis ${\bf B}_n$ and connected points
 \STATE 1: Set origin at $n=0$, ${\bf x}^o$, and center the data in ${\bf x}^o$
 \REPEAT
 \STATE 2: Define a neighborhood around ${\bf x}^n$ according to the $k$-neighbors rule
 \STATE 3: Compute PCA with the neighbors, obtaining eigenvectors ${\bf V}$
 \STATE 4: Align and reorder ${\bf V}$ with the corresponding vectors in ${\bf B}_n$
 \STATE 5: Compute the local mean ahead $\boldsymbol{\mu}$
 \STATE 6: Modify ${\bf V}_i$ to point the local mean ahead: ${\bf V}_i' = {\bf V}_i + \frac{1}{q} \cdot \boldsymbol{\mu}$
 \STATE 7: Compute the new point: ${\bf x}^{n+1} = {\bf x}^n + \tau {\bf V}_i'$
 \STATE 8: Compute the new reference basis: ${\bf B}_{n+1} = {\bf V}$
 \UNTIL{New point is outside the manifold or the PC crosses a previous PC}
 \end{algorithmic}
 \label{alg:pc}
\end{algorithm}

\begin{figure*}[t!]
\begin{center}

\setlength{\tabcolsep}{3pt}
\begin{tabular}{ccc}
\multicolumn{3}{c}{\textbf{Effect of $k$-neighborhood:} $\tau=1$, $q=10$} \\
$k=1 \%$ & $k=10 \%$ & $k=30 \%$ \\
\includegraphics[width=5.4cm]{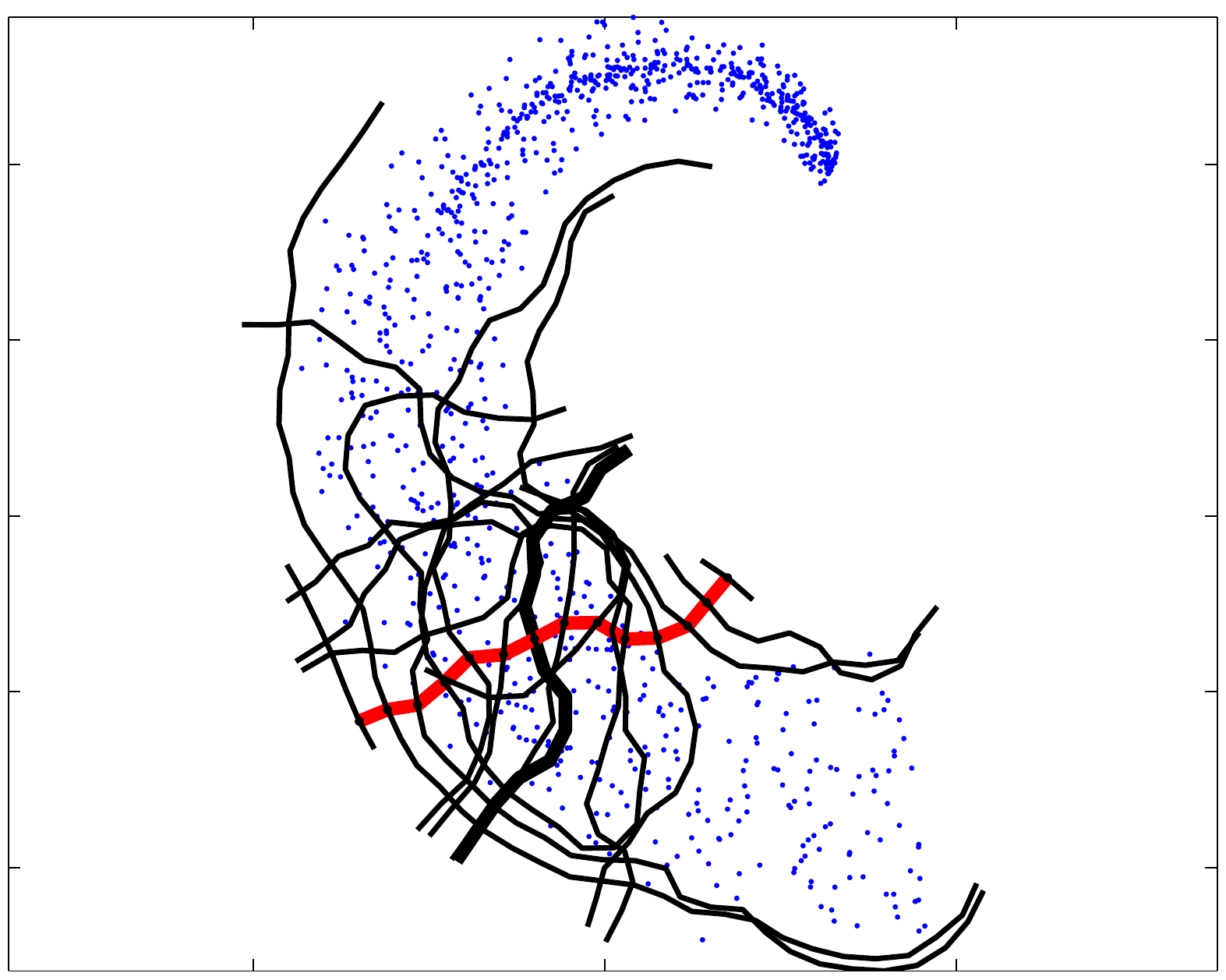} &
\includegraphics[width=5.4cm]{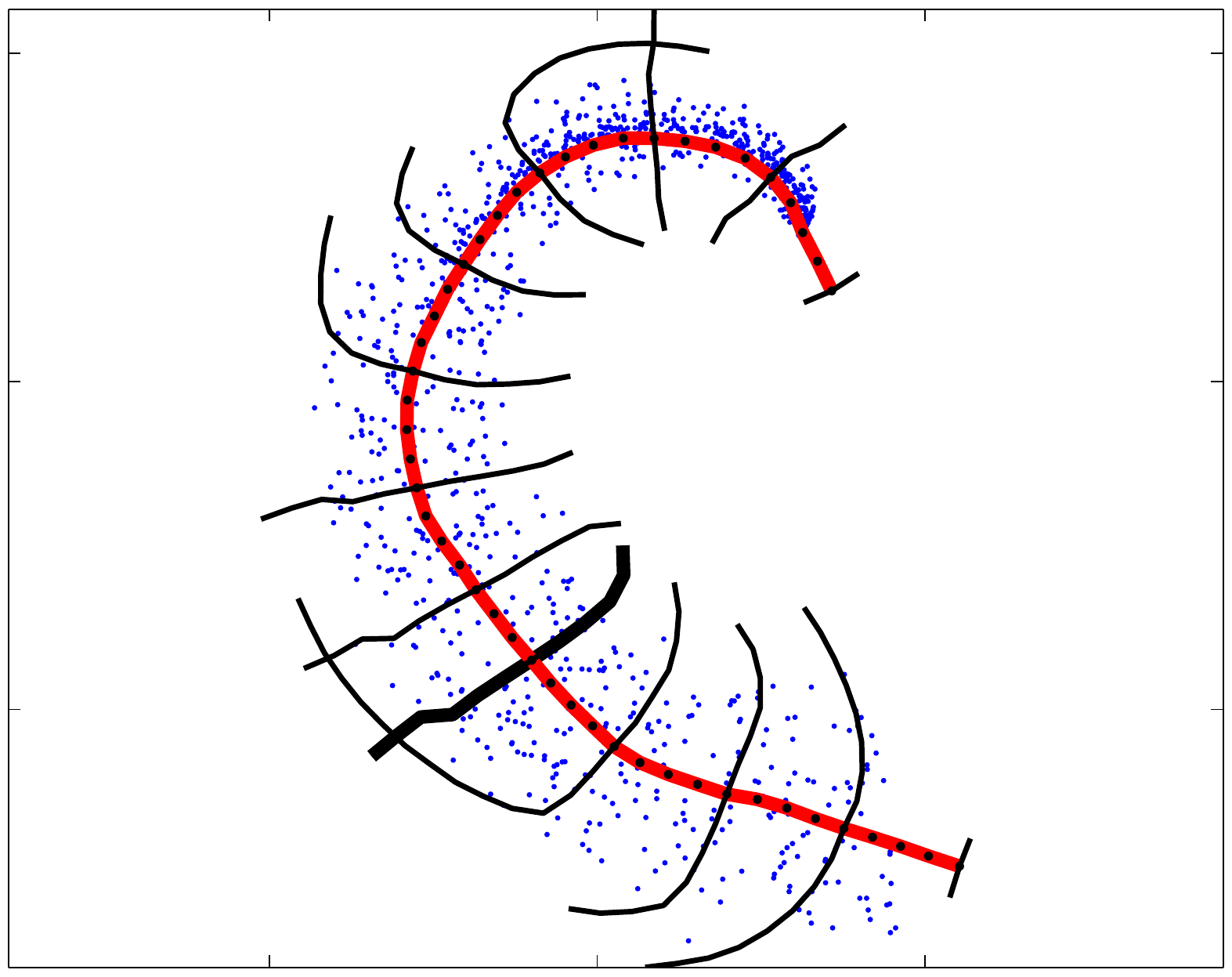} &
\includegraphics[width=5.4cm]{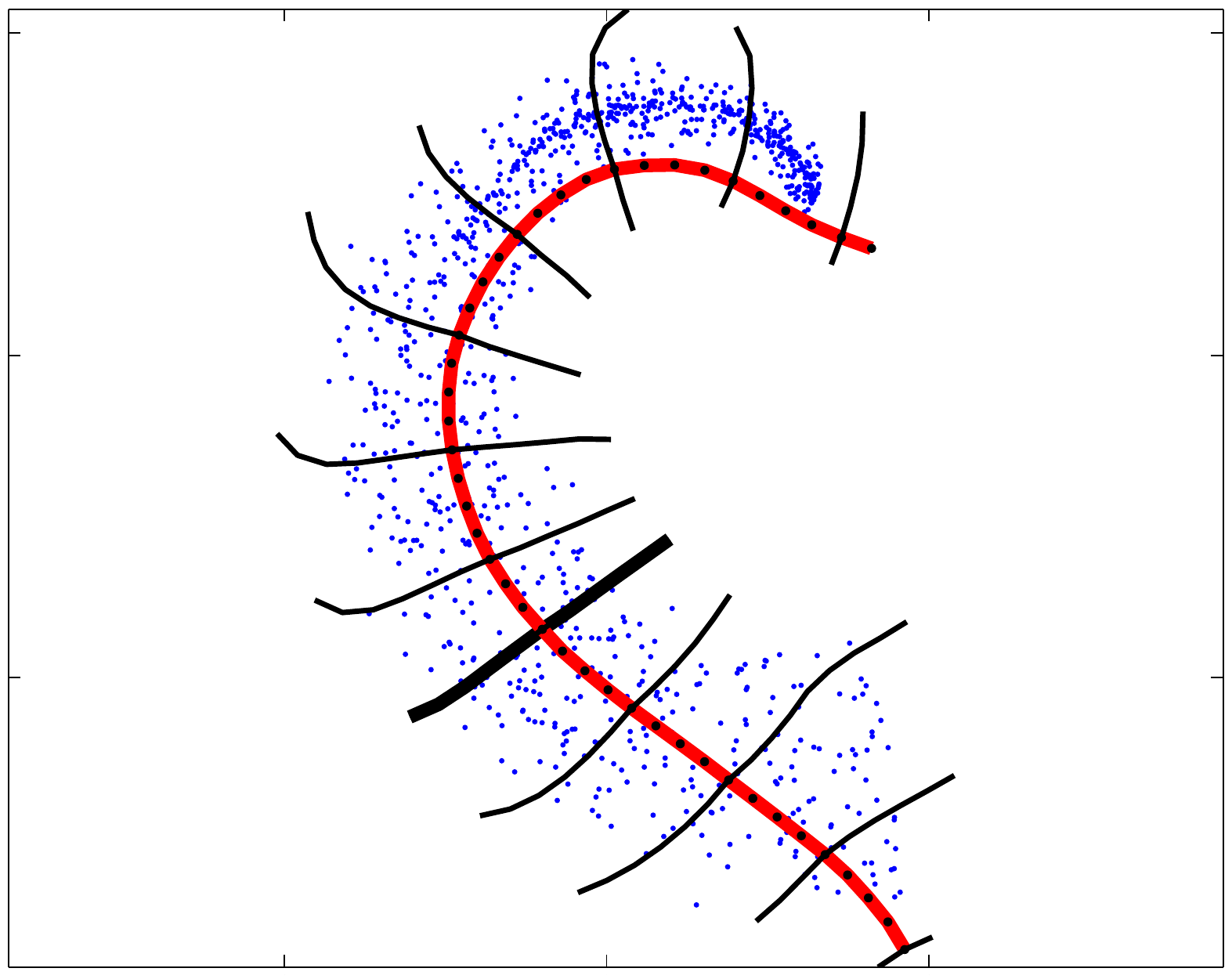} \\
\multicolumn{3}{c}{\textbf{Effect of step size $\tau$:} $k=10 \%$, $q=10$} \\
$\tau=0.02$ & $\tau=1$ & $\tau=3$ \\
\includegraphics[width=5.4cm]{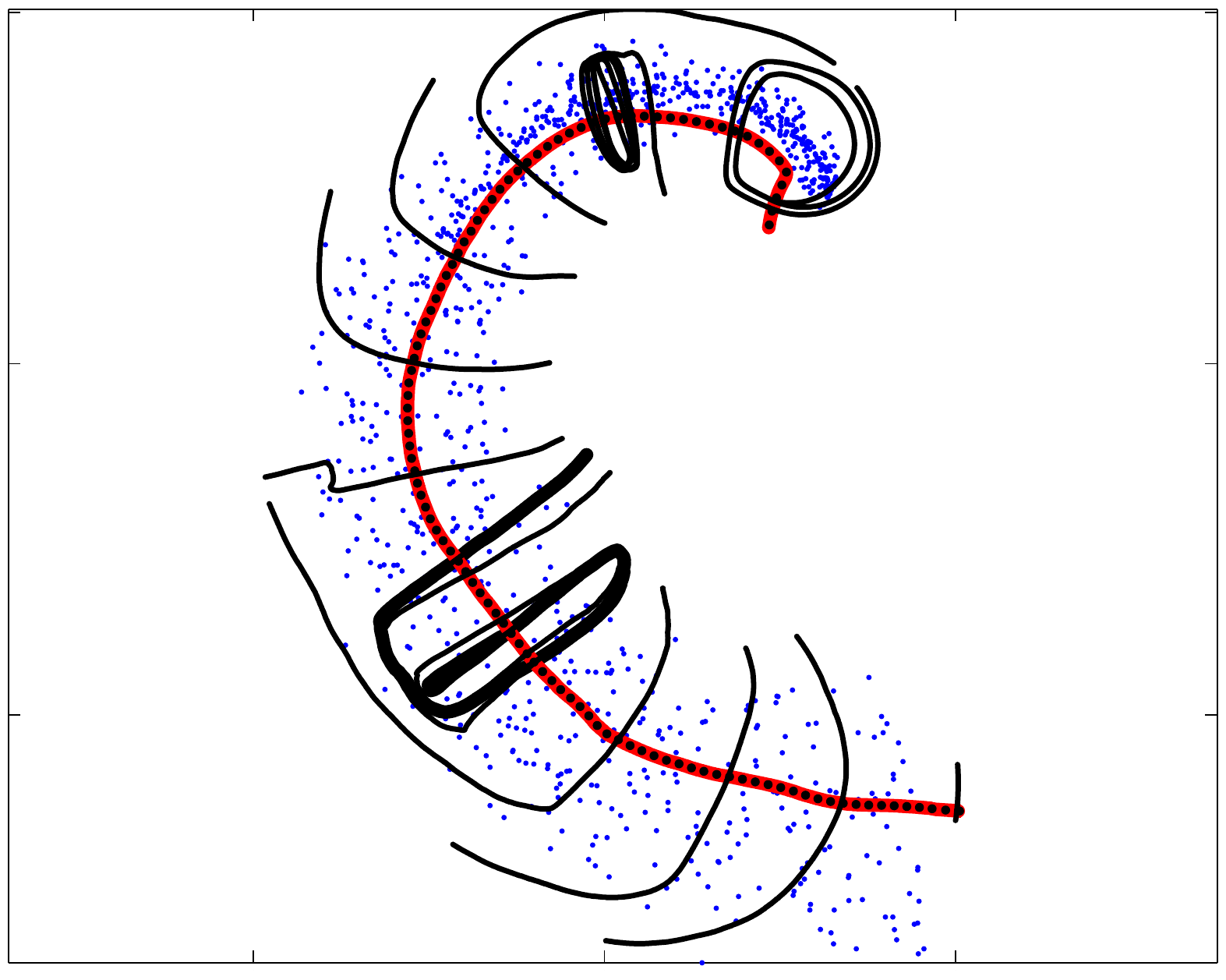} &
\includegraphics[width=5.4cm]{technical_repork/distintos_parametros/Cuerno_varias_curvas_par_1.pdf} &
\includegraphics[width=5.4cm]{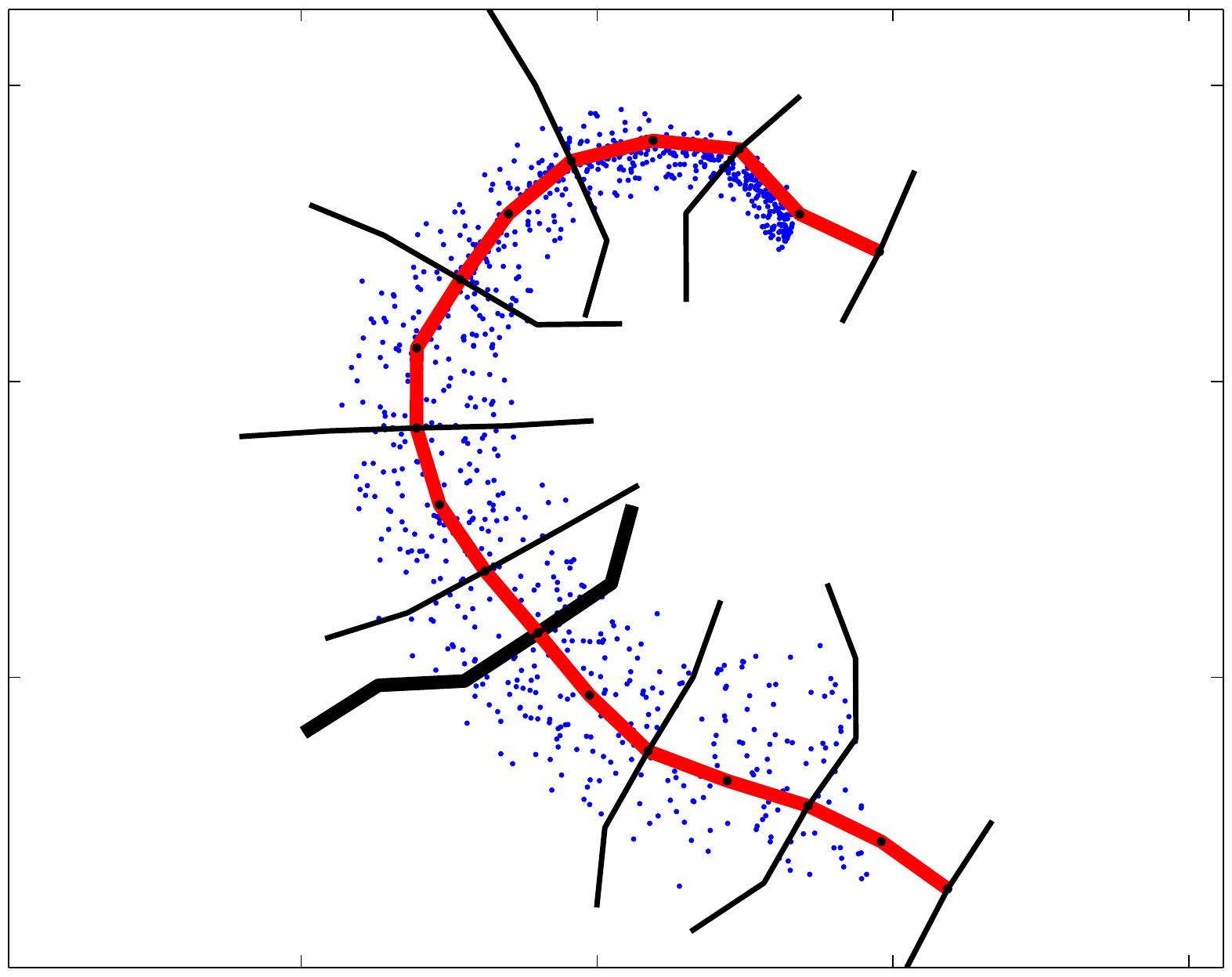} \\
\multicolumn{3}{c}{\textbf{Effect of stiffness $q$:} $k=10 \%$, $\tau=1$} \\
$q=2$ & $q=10$ & $q=\infty$ \\
\includegraphics[width=5.4cm]{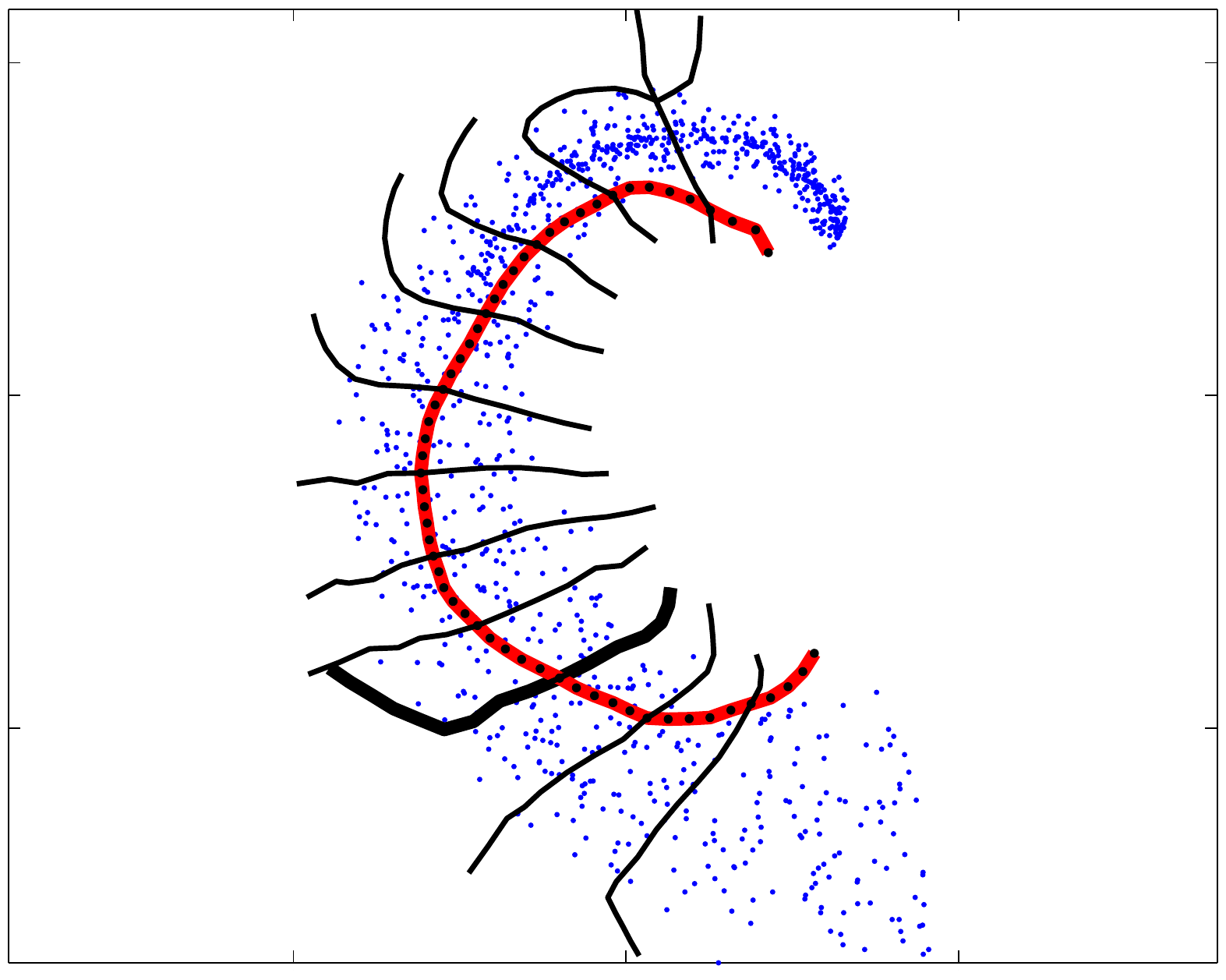} &
\includegraphics[width=5.4cm]{technical_repork/distintos_parametros/Cuerno_varias_curvas_par_1.pdf} &
\includegraphics[width=5.4cm]{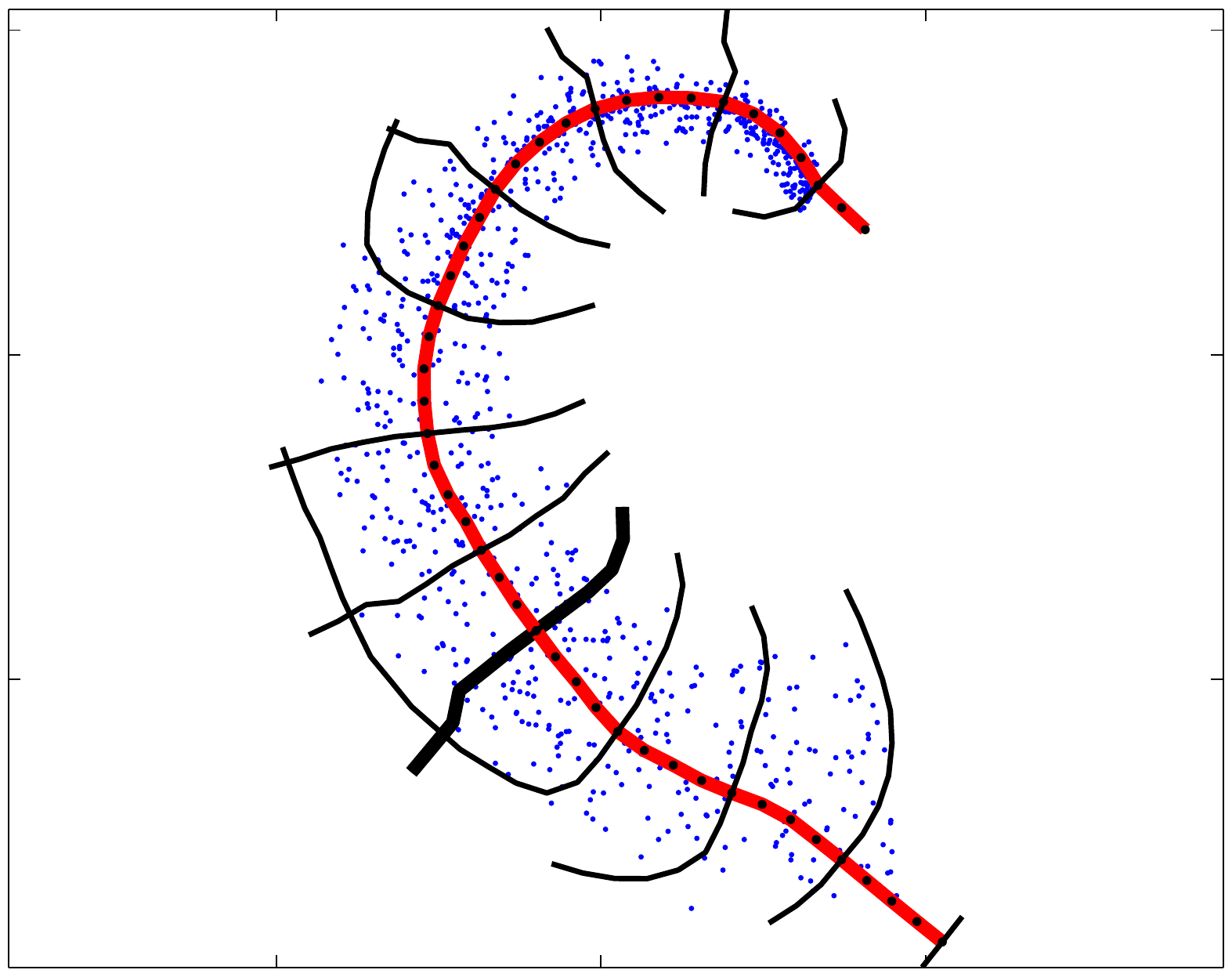} \\
\end{tabular}
\end{center}
\caption{Effect of the $k$-rule (top), the step size $\tau$ (middle), and the stiffness $q$ (bottom) on the first
and secondary PCs at different points. The first PC is shown in red and the secondary are plotted in black.
The secondary PC at the selected origin of coordinates is in bold style.
Note that in this example the crossing criterion is applied only between the first PC and each of the secondary PCs.
The second row shows the data projected onto the first PC, which helps in assessing the committed unfolding errors.
}
\label{effects}
\end{figure*}

\begin{figure*}[t!]
\begin{center}
\begin{tabular}{cc}
  \includegraphics[width=7cm]{images/SPCA_1_ori_swiss_3D.pdf} & \includegraphics[width=7cm]{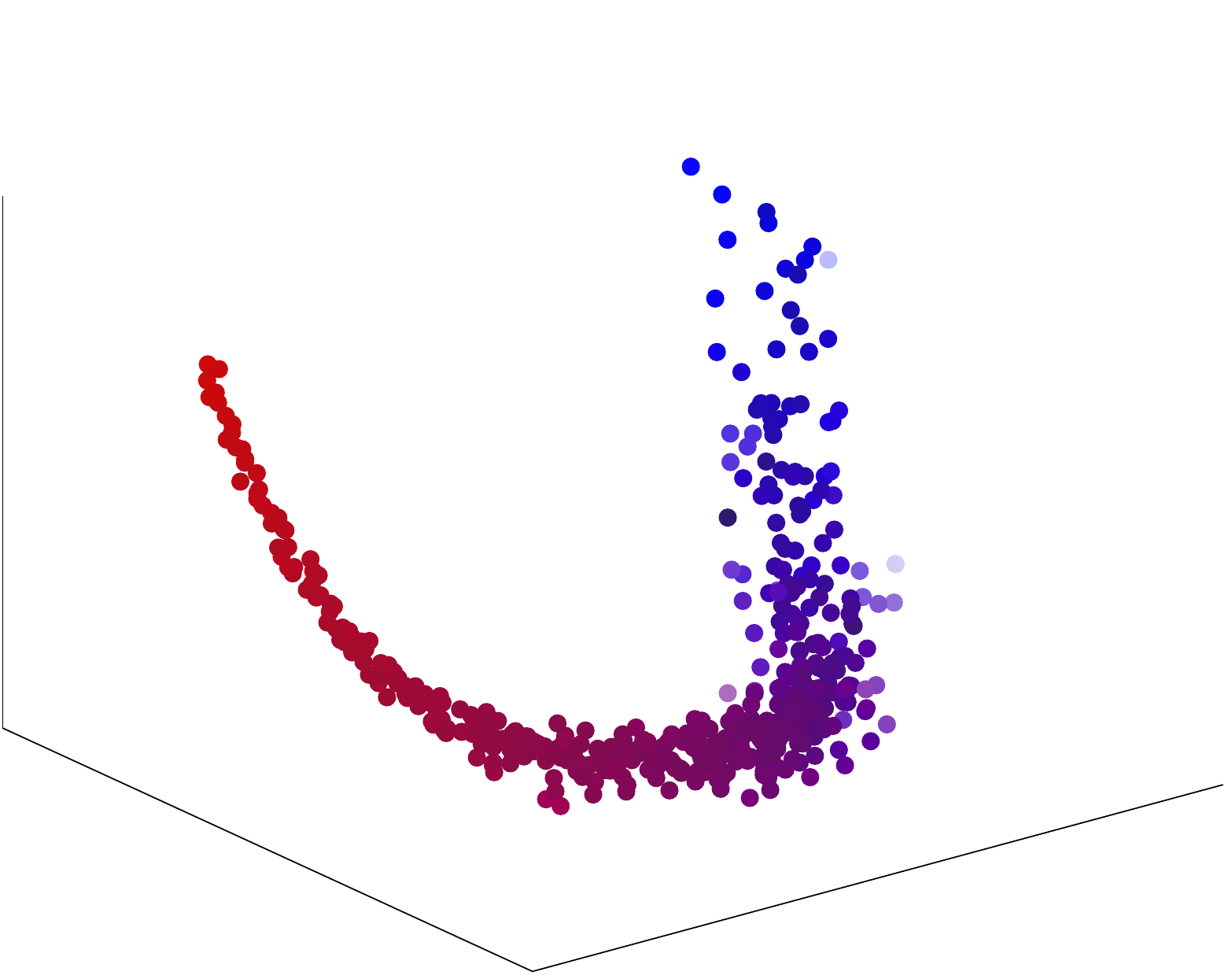}\\
\end{tabular}
\end{center}
\vspace{-0.25cm}
\caption{Manifolds for the parameter estimation experiment: noisy swiss roll (left), and noisy 1-d curve in a 3d-space (right)}.
\label{datos_procedure}
\end{figure*}


\vspace{0.15cm}
\textbf{Effect of instrumental Principal Curve parameters}
In this experiment we analyze the effect of the rigidity parameters.
The experiment is done on a synthetic example of a challenging 2D manifold: \emph{noisy spiral data}.
We generated $10000$ samples from a curved manifold with changing PDF along the underlying PC:
half of the manifold follows an increasing variance Laplacian distribution while the other
half follows an increasing variance uniform distribution (Fig.~\ref{effects}).




As anticipated above, the {\em rigidity} of the PCs depends on different parameters:
(2) the size of the $k$-rule neighborhood, (2) the step size, $\tau$, and (3) the stiffness parameter $q$.
Some prescriptions to tune them are given below:
\begin{itemize}
\item The parameter $k$ determines the number of samples of the considered neighborhood for local PCA computation.
      Too small $k$ values give rise to noisy PCA estimates and hence the obtained PC becomes unstable,
      while too large $k$ values may give rise to neighborhoods which are not local enough and thus too rigid PCs are obtained.
      See Fig.~\ref{effects}[top] for the illustration of these effects.
\item The step size, $\tau$, is the distance between the points where local basis are computed when drawing the PCs
      before setting the metric. Too large $\tau$ values give rise to too rigid PCs while too small values not only
      increase the computational time but also gives rise to unstable results: when the curve approaches one extreme
      of the manifold too small step size may cause unexpected turns, see Fig.~\ref{effects}[middle]. This parameter
      is given in Euclidean units in the input domain: if prior  information on the maximum possible curvature for
      the manifold is available, this can be used to set the $\eta$ value conveniently.
\item Finally (Fig.~\ref{effects}[bottom]), the stiffness controls the impact of the local mean ahead in the
      modification of the local PCA direction. Too small stiffness results on a big impact of the local mean,
      increased flexibility and unstable results. Too large stiffness implies neglecting the local mean which for big
      $\tau$ leads to too rigid solutions.
\end{itemize}
\vspace{0.15cm}


\textbf{Fitting the instrumental Principal Curve parameters}
Different datasets may present different curvatures and inhomogeneities so different parameters of the algorithm may be needed.
Here we present a criterion in order to tune the parameters of the algorithm for a particular dataset.
This criterion is based on the definition of PC given in \cite{Hastie89}: it is a self-consistent smooth curve
which passes through the middle of a d-dimensional data cloud. In particular the parameters can be selected in order
to minimize the distance between the data points and its projections on the drawn PC. This criterion is equivalent to
minimize the reconstruction error obtained when using the PC for dimensionality reduction (reduction from dimension $d$ to dimension $1$),
and it is also equivalent to minimize the variance in the subspace orthogonal to the PC which is the optimization criterion in classical
(linear) Principal Component Analysis \cite{Jolliffe02}.

\begin{figure}[t!]
\begin{center}
\includegraphics[width=9cm]{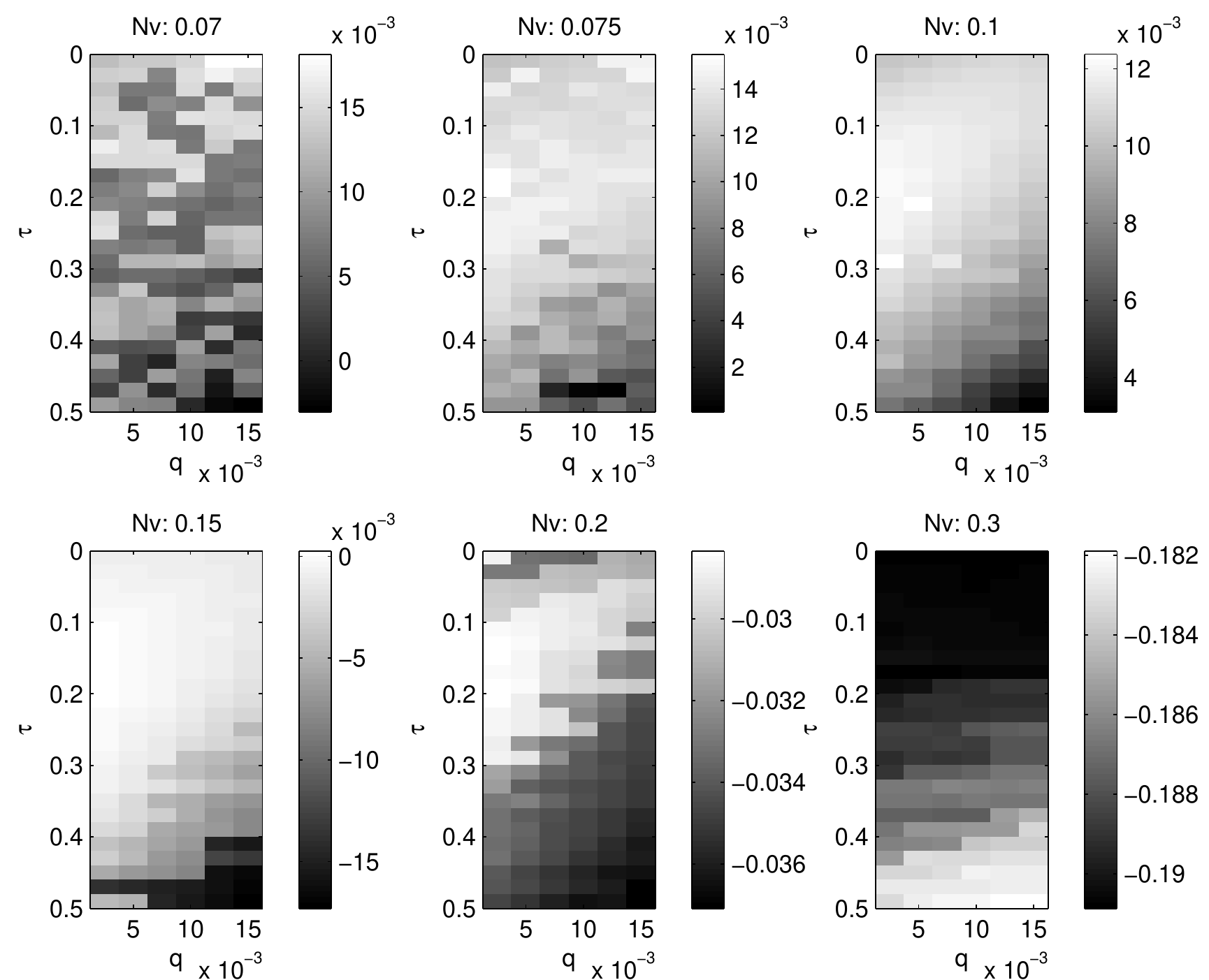}
\end{center}
\vspace{-0.25cm}
\caption{Projection error in the swiss roll example. Darker means more error. Each image is from different
k-neighbors, inside each image the parameters $\tau$ and $q$ are modified.
}
\label{ERR_swiss}
\end{figure}

\begin{figure}[t!]
\begin{center}
\includegraphics[width=9cm]{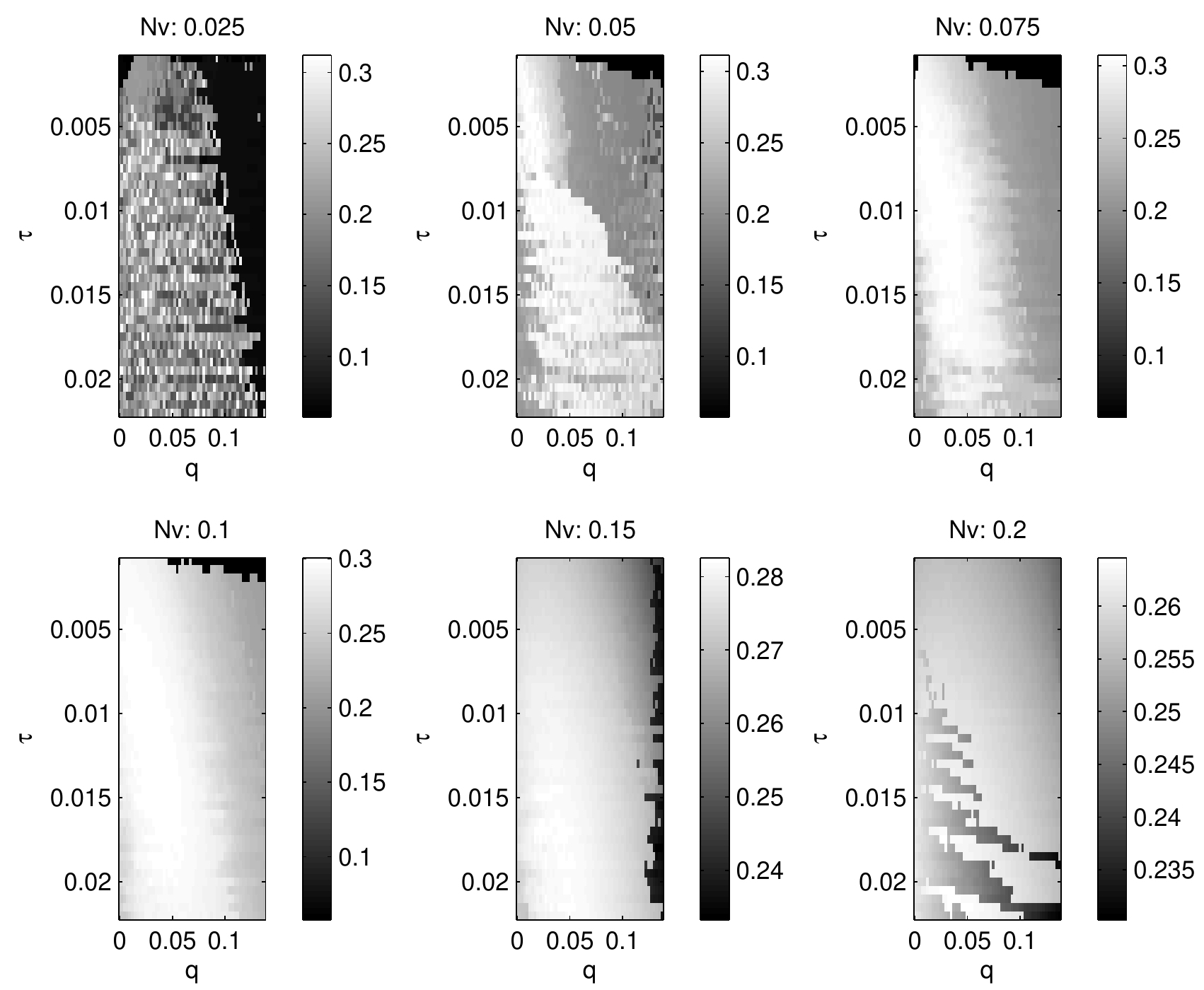}
\end{center}
\vspace{-0.25cm}
\caption{Projection error 1-d curve embedded in the 3-d space. Darker means more error. Each image is fro different
k-neighbors, inside each image the parameters $\tau$ and $q$ are modified.
}
\label{ERR_cuerno}
\end{figure}

In order to show that different manifolds may require different parameters of the algorithm, in the following experiment
we consider data from the manifolds shown in Fig. \ref{datos_procedure}: swiss roll, and 1-d curve embedded in a 3-d space.
Figures \ref{ERR_swiss} and \ref{ERR_cuerno} show how the error varies depending on the parameters for the considered manifolds.
The optimal parameters can be obtained from the minima (lighter points) in these surfaces.
As apparent in the figures, these manifolds require different parameters to minimize the projection error.

\bibliographystyle{IEEEbib}
\bibliography{sandra_adicionales,NeCoColor2,SPC_short}

\begin{thebibliography}{10}

\bibitem{Laparra12}
V.~Laparra, S.~Jim\'enez, G.~Camps-Valls, and J.~Malo,
\newblock ``Nonlinearities and adaptation of color vision from {S}equential
  {P}rincipal {C}urves {A}nalysis,''
\newblock {\em Neural Computation}, vol. 24, no. 10, pp. 2751–--2788, 2012.

\bibitem{Laparra15b}
V.~Laparra and J.~Malo,
\newblock ``Visual aftereffects and sensory nonlinearities from a single
  statistical framework,''
\newblock {\em Front. Human Neurosci.}, vol. 9, 2015.

\bibitem{Malo06b}
J.~Malo and J.~Guti\'errez,
\newblock ``{V1} non-linear properties emerge from local-to-global non-linear
  {ICA},''
\newblock {\em {Network: Comp. Neur. Syst.}}, vol. 17, no. 1, pp. 85--102,
  2006.

\bibitem{Laparra14a}
V.~Laparra, S.~Jim\'enez, D.~Tuia, G.~Camps, and J.~Malo,
\newblock ``Principal polynomial analysis,''
\newblock {\em Int. J. Neur. Syst.}, vol. 24, no. 6, pp. 1--21, 2014.

\bibitem{Laparra15a}
V.~Laparra, J.~Malo, and G.~Camps,
\newblock ``Dimensionality reduction via regression in hyperspectral imagery,''
\newblock {\em IEEE J. Sel. Topics Sig. Proc.}, vol. 9, no. 9, 2015.

\bibitem{Clarke85}
R.J. Clarke,
\newblock {\em Transform Coding of Images},
\newblock Acad. Press, NY, 1985.

\bibitem{Gersho92}
A.~Gersho and R.~M. Gray,
\newblock {\em Vector Quantization and Signal Compression},
\newblock Kluwer, Boston, 1992.

\bibitem{Hancock92}
P.J. Hancock, R.J. Baddeley, and L.S. Smith,
\newblock ``The principal components of natural images,''
\newblock {\em Network}, vol. 3, no. 1, pp. 61--70, 1992.

\bibitem{Taubman01}
D.S. Taubman and M.W. Marcellin,
\newblock {\em {JPEG}2000: Image Compression Fundamentals, Standards and
  Practice},
\newblock Kluwer Academic Publishers, Boston, 2001.

\bibitem{Lowe04}
D.~G. Lowe,
\newblock ``Distinctive image features from scale-inveriant keypoints,''
\newblock {\em Intl. J. Comp. Vis.}, vol. 60, pp. 91--110, 2004.

\bibitem{Wright10}
J.~Wright, Y.~Ma, J.~Mairal, G.~Sapiro, T.S. Huang, and S.~Yan,
\newblock ``Sparse representation for computer vision and pattern
  recognition,''
\newblock {\em Proc. IEEE}, vol. 98, pp. 1031--1044, 2010.

\bibitem{Field96}
B.~A. Olshausen and D.~J. Field,
\newblock ``Emergence of simple-cell receptive field properties by learning a
  sparse code for natural images,''
\newblock {\em Nature}, vol. 381, pp. 607--609, 1996.

\bibitem{Bell97}
A.~J. Bell and T.~J. Sejnowski,
\newblock ``The `independent components' of natural scenes are edge filters,''
\newblock {\em Vision Research}, vol. 37, no. 23, pp. 3327--3338, 1997.

\bibitem{Hoyer00}
P.O. Hoyer and A.~Hyvarinen,
\newblock ``Independent component analysis applied to feature extraction from
  colour and stereo images,''
\newblock {\em Network: Computation in Neural Systems}, vol. 11, pp. 191--210,
  2000.

\bibitem{Simoncelli01}
E.~P. Simoncelli and B.~O. Olshausen,
\newblock ``Natural image statistics and neural representation.,''
\newblock {\em Annu. Rev. Neurosci.}, vol. 24, pp. 1193--1216, 2001.

\bibitem{Doi03}
E.~Doi, T.~Inui, T.~Lee, T.~Wachtler, and T.~Sejnowski,
\newblock ``Spatiochromatic receptive field properties derived from
  information-theoretic analysis of cone mosaic responses to natural scenes,''
\newblock {\em Neural Computation}, vol. 15, no. 2, pp. 397--417, 2003.

\bibitem{Laughlin83}
S.~B. Laughlin,
\newblock ``Matching coding to scenes to enhance efficiency,''
\newblock in {\em In Braddick, O.J. \& Sleigh, A.C. (Eds) Physical and
  Biological Processing of Images}. 1983, pp. 42--52, Springer.

\bibitem{Bell95}
A.~J. Bell and T.~J. Sejnowski,
\newblock ``An information-maximization approach to blind separation and blind
  deconvolution,''
\newblock {\em Neural. Comput}, vol. 7, no. 6, pp. 1129--1159, 1995.

\bibitem{Simoncelli97}
E~P Simoncelli,
\newblock ``Statistical models for images: {C}ompression, restoration and
  synthesis,''
\newblock in {\em Proc 31st Asilomar Conf on Signals, Systems and Computers},
  Pacific Grove, CA, Nov 2-5 1997, vol.~1, pp. 673--678, IEEE Computer Society.

\bibitem{Bucigrossi99}
R.~W. Buccigrossi and E.P. Simoncelli,
\newblock ``Image compression via joint statistical characterization in the
  wavelet domain,''
\newblock {\em IEEE Tr. Im. Proc.}, vol. 8, no. 12, pp. 1688--1701, 1999.

\bibitem{Hyvarinen03}
A.~Hyv\"arinen, J.~Hurri, and J.~V\"ayrynen,
\newblock ``Bubbles: a unifying framework for low-level statistical properties
  of natural image sequences,''
\newblock {\em JOSA A}, vol. 20, no. 7, pp. 1237--1252, 2003.

\bibitem{Malo06ieee}
J.~Malo, I.~Epifanio, R.~Navarro, and E.~Simoncelli,
\newblock ``Non-linear image representation for efficient perceptual coding,''
\newblock {\em IEEE Tr. Im. Proc.}, vol. 15, no. 1, pp. 68--80, 2006.

\bibitem{Gutierrez06}
J.~Guti\'errez, F.~Ferri, and J.~Malo,
\newblock ``Regularization operators for natural images based on nonlinear
  perception models,''
\newblock {\em IEEE Tr. Im. Proc.}, vol. 15, pp. 189--200, 2006.

\bibitem{Camps08}
G.~Camps, J.~Guti\'errez, G.~G\'omez, and J.~Malo,
\newblock ``On the suitable domain for {SVM} training in image coding.,''
\newblock {\em JMLR}, vol. 9, pp. 49--66, 2008.

\bibitem{Hyvarinen09}
A.~Hyvarinen, J.~Hurri, and P.~Hoyer,
\newblock {\em Natural Image Statistics: A Probabilistic Approach to Early
  Computational Vision.},
\newblock Springer, Berlin, 2009.

\bibitem{Malo10}
J.~Malo and V.~Laparra,
\newblock ``Psychophysically tuned divisive normalization approximately
  factorizes the {PDF} of natural images,''
\newblock {\em Neur. Comp.}, vol. 22, no. 12, pp. 3179--3206, 2010.

\bibitem{Foley94}
J.~M. Foley,
\newblock ``Human luminance pattern mechanisms: Masking experiments require a
  new model,''
\newblock {\em JOSA}, vol. 11, no. 6, pp. 1710--1719, 1994.

\bibitem{Watson97}
A.~B. Watson and J.~A. Solomon,
\newblock ``A model of visual contrast gain control and pattern masking,''
\newblock {\em JOSA A}, vol. 14, pp. 2379--2391, 1997.

\bibitem{Carandini94}
M.~Carandini and D.~J. Heeger,
\newblock ``Summation and division by neurons in visual cortex,''
\newblock {\em Science}, vol. 264, pp. 1333--1336, 1994.

\bibitem{Cavaugnagh02}
J.~R. Cavanaugh, W.~Bair, and J.~A. Movshon,
\newblock ``Selectivity and spatial distribution of signals from the receptive
  field surround in macaque v1 neurons,''
\newblock {\em J. Neurophysiol}, vol. 88, no. 5, pp. 2547--2556, 2002.

\bibitem{Carandini12}
M.~Carandini and D.~Heeger,
\newblock ``{Normalization as a canonical neural computation.},''
\newblock {\em Nature Reviews. Neurosci.}, vol. 13, no. 1, pp. 51--62, 2012.

\bibitem{Epifanio03}
I.~Epifanio, J.~Guti\'errez, and J.Malo,
\newblock ``Linear transform for simultaneous diagonalization of covariance and
  perceptual metric matrix in image coding,''
\newblock {\em Pattern Recognition}, vol. 36, pp. 1799--1811, 2003.

\bibitem{Laparra10}
V.~Laparra, J.~Mu\~{n}oz Mar\'i, and J.~Malo,
\newblock ``Divisive normalization image quality metric revisited,''
\newblock {\em JOSA A}, vol. 27, no. 4, pp. 852--864, 2010.

\bibitem{Roweis00}
S.~T. Roweis and L.~K. Saul,
\newblock ``Nonlinear dimensionality reduction by locally linear embedding,''
\newblock {\em Science}, vol. 290, no. 5500, pp. 2323--2326, December 2000.

\bibitem{Belkin02}
M.~Belkin and P.~Niyogi,
\newblock ``Laplacian eigenmaps for dimensionality reduction and data
  representation,''
\newblock {\em Neural Computation}, vol. 15, pp. 1373--1396, 2002.

\bibitem{Weinberger04}
K.~Q. Weinberger and L.~K. Saul,
\newblock ``Unsupervised learning of image manifolds by semidefinite
  programming,''
\newblock in {\em Proc. IEEE CVPR}, 2004, pp. 988--995.

\bibitem{Scholkopf98}
B~Sch{\"o}lkopf, A~J. Smola, and K-R. M{\"u}ller,
\newblock ``Nonlinear component analysis as a kernel eigenvalue problem,''
\newblock {\em Neural Computation}, vol. 10, no. 5, pp. 1299--1319, 1998.

\bibitem{Burges99}
C.~J.~C. Burges,
\newblock ``{Geometry and Invariance in Kernel Based Methods},''
\newblock in {\em Advances in Kernel Methods: Support Vector Learning},
  B.~Sch{\"o}lkopf, C.~J.~C. Burges, and A.~J. Smola, Eds. MIT Press, 1999.

\bibitem{Kambhatla97}
N.~Kambhatla and T.~Leen,
\newblock ``Dimension reduction by local {PCA},''
\newblock {\em Neural Computation}, vol. 9, pp. 1493--1500, 1997.

\bibitem{Roweis02}
S.~T. Roweis, L.~K. Saul, and G.~E. Hinton,
\newblock ``Global coordination of local linear models,''
\newblock in {\em Advances in Neural Information Processing Systems 14}. 2002,
  pp. 889--896, MIT Press.

\bibitem{Verbeek02}
J.~J. Verbeek, N.~Vlassis, and B.~Krose,
\newblock ``Coordinating principal component analyzers,''
\newblock in {\em In Proc. International Conference on Artificial Neural
  Networks}. 2002, pp. 914--919, Springer.

\bibitem{Teh03}
Y.~W. Teh and S.~Roweis,
\newblock ``Automatic alignment of local representations,''
\newblock in {\em NIPS 15}. 2003, pp. 841--848, MIT Press.

\bibitem{Brand03}
Matthew Brand,
\newblock ``Charting a manifold,''
\newblock in {\em NIPS 15}. 2003, pp. 961--968, MIT Press.

\bibitem{Lin99}
J.~Lin,
\newblock ``Factorizing probability density functions: Generalizing {ICA},''
\newblock in {\em Proc. 1st Intl. Workshop on {ICA} and Signal Separation},
  1999, pp. 313--318.

\bibitem{Kohonen82}
T.~Kohonen,
\newblock ``Self-organized formation of topologically correct feature maps,''
\newblock {\em Biological Cybernetics}, vol. 43, no. 1, pp. 59--69, January
  1982.

\bibitem{Bishop98}
C.~M. Bishop, M.~Svens\'en, and C.~K.~I. Williams,
\newblock ``{GTM}: The generative topographic mapping,''
\newblock {\em Neural Computation}, vol. 10, pp. 215--234, 1998.

\bibitem{Tenenbaum2000}
Joshua~B. Tenenbaum, Vin Silva, and John~C. Langford,
\newblock ``A global geometric framework for nonlinear dimensionality
  reduction,''
\newblock {\em Science}, vol. 290, no. 5500, pp. 2319--2323, December 2000.

\bibitem{Delicado01}
P.~Delicado,
\newblock ``Another look at principal curves and surfaces,''
\newblock {\em J. Multivar. Anal.}, vol. 77, pp. 84--116, 2001.

\bibitem{Hastie89}
T.~Hastie and W.~Stuetzle,
\newblock ``Principal curves,''
\newblock {\em J. Am. Stats. Assoc.}, vol. 84, no. 406, pp. 502--516, 1989.

\bibitem{Einbeck05}
J.~Einbeck, G.~Tutz, and L.~Evers,
\newblock ``Local principal curves,''
\newblock {\em Statistics and Computing}, vol. 15, pp. 301--313, 2005.

\bibitem{Einbeck10}
J.~Einbeck, L.~Evers, and K.~Hinchliff,
\newblock {\em Data compression and regression based on local principal
  curves}, pp. 701--712,
\newblock Springer, 2010.

\bibitem{OzertemTesis}
U.~Ozertem,
\newblock {\em Locally Defined Principal Curves and Surfaces},
\newblock Ph.D. thesis, Dept. of Science and Engineering, Oregon, USA, Sept.
  2008.

\bibitem{Ozertem11}
U.~Ozertem and D.~Erdogmus,
\newblock ``Locally defined principal curves and surfaces,''
\newblock {\em JMLR}, vol. 12, pp. 1249--1286, 2011.

\bibitem{Jolliffe02}
I.~T. Jolliffe,
\newblock {\em Principal {C}omponent {A}nalysis},
\newblock Springer Verlag, Berlin, Germany, 2002.

\bibitem{Donnell94}
D.~Donnell, A.~Buja, and W.~Stuetzle,
\newblock ``Analysis of additive dependencies and concurvities using smallest
  additive principal components,''
\newblock {\em The Annals of Statistics}, vol. 22, no. 4, pp. 1635--1668, 1994.

\bibitem{Besse95}
P.~C. Besse and F.~Ferraty,
\newblock ``Curvilinear fixed effect model,''
\newblock {\em Computational Statistics}, vol. 10, pp. 339--351, 1995.

\bibitem{Laparra12ppa}
V.~Laparra, D.~Tuia, S.~Jimenez, G.~Camps, and J.~Malo,
\newblock ``Nonlinear data description with {P}rincipal {P}olynomial
  {A}nalysis,''
\newblock in {\em IEEE Mach. Learn. Sig. Proc.}, 2012.

\bibitem{Kramer91}
M.~A. Kramer,
\newblock ``Nonlinear principal component analysis using autoassociative neural
  networks,''
\newblock {\em AIChE Journal}, vol. 37, no. 2, pp. 233--243, 1991.

\bibitem{Hinton06}
G.~E. Hinton and R.~R. Salakhutdinov,
\newblock ``Reducing the dimensionality of data with neural networks,''
\newblock {\em Science}, vol. 313, no. 5786, pp. 504--507, July 2006.

\bibitem{Scholz07}
M.~Scholz, M.~Fraunholz, and J.~Selbig,
\newblock {\em Nonlinear principal component analysis: neural networks models
  and applications}, chapter~2, pp. 44--67,
\newblock Springer, 2007.

\bibitem{MacLeod03}
D.~A. MacLeod,
\newblock ``Color discrimination, color constancy, and natural scene
  statistics,''
\newblock in {\em Normal and defective color vision}, J.~Mollon, J.~Pokorny,
  and K.~Knoblauch, Eds., pp. 189--218. Oxford Univ. Press, 2003.

\bibitem{Hyvarinen99}
A.~Hyv\"arinen and P.~Pajunen,
\newblock ``Nonlinear independent component analysis: existence and uniqueness
  results,''
\newblock {\em Neural Networks}, vol. 12, no. 3, pp. 429--439, 1999.

\bibitem{Laparra10rbig}
V.~Laparra, G.~Camps, and J.~Malo,
\newblock ``Iterative gaussianization: from {ICA} to random rotations,''
\newblock {\em IEEE Trans. Neur. Nets.}, vol. 22, no. 4, pp. 537--549, 2011.

\bibitem{CoverThomas06}
T.~M. Cover and J.~A. Thomas,
\newblock {\em {Elements of Information Theory 2nd Edition}},
\newblock Wiley-Interscience, 2 edition, July 2006.

\bibitem{Olmos04}
A.~Olmos and F.~Kingdom,
\newblock ``A biologically inspired algorithm for the recovery of shading and
  reflectance images,''
\newblock {\em Perception}, vol. 33, pp. 1465--1473, 2004.

\bibitem{Koenderink10}
Jan~J. Koenderink,
\newblock ``The prior statistics of object colors,''
\newblock {\em JOSA A}, vol. 27, no. 2, pp. 206--217, 2010.

\bibitem{Lyu09c}
S~Lyu and E~P Simoncelli,
\newblock ``Nonlinear extraction of `independent components' of natural images
  using radial {Gaussianization},''
\newblock {\em Neural Computation}, vol. 21, no. 6, pp. 1485--519, 2009.

\bibitem{Sinz10}
F.~Sinz and M.~Bethge,
\newblock ``Lp-nested symmetric distributions,''
\newblock {\em Journal of Machine Learning Research}, vol. 11, pp. 3409--3451,
  2010.

\bibitem{Numerical92}
W.H. Press, B.P. Flannery, S.A. Teulosky, and W.T. Vetterling,
\newblock {\em Numerical Recipes in {C}: The Art of Scientific Computing},
\newblock Cambridge University Press, Cambridge, 1992.

\bibitem{Webster97}
M.~Webster and J.~Mollon,
\newblock ``Adaptation and the color statistics of natural images,''
\newblock {\em Vision Res}, vol. 37, no. 23, pp. 3283--3298, 1997.

\bibitem{Capilla04}
P.~Capilla, M.~D\'{i}ez, M.~J. Luque, and J.~Malo,
\newblock ``Corresponding-pair procedure: a new approach to simulation of
  dichromatic color perception,''
\newblock {\em J. Opt. Soc. Am. A}, vol. 21, no. 2, pp. 176--186, 2004.

\bibitem{Laparra08b}
V.~Laparra and J.~Malo,
\newblock ``Masking-like non-linearities from non-linear {PCA},''
\newblock in {\em {GRC} Conference: Sensory Coding and The Natural
  Environment}, Lucca, Italy, 2008.

\end{thebibliography}

\end{document}